\documentclass[a4paper,fleqn]{cas-dc}

\usepackage[numbers]{natbib}
\usepackage{subcaption}
\newtheorem{theorem}{Theorem}[section]

\def\tsc#1{\csdef{#1}{\textsc{\lowercase{#1}}\xspace}}
\tsc{WGM}
\tsc{QE}
\tsc{EP}
\tsc{PMS}
\tsc{BEC}
\tsc{DE}
\begin{document}
\let\WriteBookmarks\relax
\def\floatpagepagefraction{1}
\def\textpagefraction{.001}
\shorttitle{Feasibility-Aware Conformal Coverage Guarantees}
\shortauthors{Ye Li et~al.}

\title [mode = title]{Set-Valued Prediction for Large Language Models with Feasibility-Aware Coverage Guarantees}                      
% \tnotemark[1,2]

% \tnotetext[1]{This document is the results of the research
   % project funded by the National Science Foundation.}

% \tnotetext[2]{The second title footnote which is a longer text matter
%    to fill through the whole text width and overflow into
%    another line in the footnotes area of the first page.}

\author[1]{Ye Li}
% [type=editor,
                        % auid=000,bioid=1,
                        % prefix=Sir,
                        % role=Researcher,
                        % orcid=0000-0001-0000-0000]
% \cormark[1]
\ead{li_ye@std.uestc.edu.cn}
% \ead[url]{www.jkkrishnan.in}

% \credit{Conceptualization of this study, Methodology, Software}

%\address[1]{, Street 129, 1043 NX Amsterdam, The Netherlands}

\author[1]{Anqi Hu}
\ead{hu_anqi@std.uestc.edu.cn}

\author[2]{Yuanchang Ye}
% [style=chinese]
\ead{yuanchang0213@zufe.edu.cn}

\author[3]{Shiyan Tong}
% [%
%    role=Co-ordinator,
%    suffix=Jr,
\ead{tongshiyan@seu.edu.cn}
% \ead[URL]{https://www.university.org}

% \credit{Data curation, Writing - Original draft preparation}

\author[1]{Zhiyuan Wang}[orcid=0009-0009-3835-4596]
\cormark[1]
\ead{yhzywang@gmail.com}

\author[1]{Bo Fu}
\cormark[1]
% \fnmark[1,3]
\ead{fubo@uestc.edu.cn}
% \ead[URL]{www.campus.in}

\affiliation[1]{organization={University of Electronic Science and Technology of China},
                % addressline={Jawahar Nagar}, 
                city={Chengdu},
%               citysep={}, % Uncomment if no comma needed between city and postcode
                % postcode={611731}, 
                state={Sichuan},
                country={China}}

\affiliation[2]{organization={Zhejiang University of Finance and Economics},
                % addressline={Street 29}, 
                % postcode={310018}, 
                city={Hangzhou},
                state={Zhejiang},
                country={China}}

\affiliation[3]{organization={Southeast University},
                % addressline={Street 15}, 
                city={Nanjing},
                % postcode={xxx}, 
                state={Jiangsu}, 
                country={China}}

\cortext[cor1]{Corresponding authors}
% \cortext[cor2]{Principal corresponding author}
% \fntext[fn1]{This is the first author footnote, but is common to third
  % author as well.}
% \fntext[fn2]{Another author footnote, this is a very long footnote and
%   it should be a really long footnote. But this footnote is not yet
%   sufficiently long enough to make two lines of footnote text.}

% \nonumnote{This note has no numbers. In this work we demonstrate $a_b$
%   the formation Y\_1 of a new type of polariton on the interface
%   between a cuprous oxide slab and a polystyrene micro-sphere placed

\begin{abstract}
Large language models (LLMs) inherently operate over a large generation space, yet conventional usage typically reports the most likely generation (MLG) as a point prediction, which underestimates the model's capability: although the top-ranked response can be incorrect, valid answers may still exist within the broader output space and can potentially be discovered through repeated sampling. 
This observation motivates moving from point prediction to set-valued prediction, where the model produces a set of candidate responses rather than a single MLG. 
In this paper, we propose a principled framework for set-valued prediction, which provides feasibility-aware coverage guarantees. 
We show that, given the finite-sampling nature of LLM generation, coverage is not always achievable: even with multiple samplings, LLMs may fail to yield an acceptable response for certain questions within the sampled candidate set. 
To address this, we establish a minimum achievable risk level (MRL), below which statistical coverage guarantees cannot be satisfied. 
Building on this insight, we then develop a data-driven calibration procedure that constructs prediction sets from sampled responses by estimating a rigorous threshold, ensuring that the resulting set contains a correct answer with a desired probability whenever the target risk level is feasible. 
Extensive experiments on six language generation tasks with five LLMs demonstrate both the statistical validity and the predictive efficiency of our framework. 
\end{abstract}

% \includegraphics[width=\linewidth]{figs/overview.pdf}

% \begin{highlights}
% \item Feasibility-aware set-valued prediction with finite-sample coverage guarantees for open-ended LLM generation;
% \item Minimum risk level characterizes the feasibility boundary induced by finite sampling;
% \item Data-driven threshold calibration yields efficient prediction sets with valid coverage when the target risk is feasible. 
% \end{highlights}

\begin{keywords}
large language models \sep set-valued prediction \sep feasibility-aware coverage guarantees \sep minimum achievable risk level \sep data-driven calibration
\end{keywords}

\maketitle

\section{Introduction}
Large language models (LLMs) have shown remarkable capabilities across a broad range of natural language generation (NLG) tasks, including question answering (QA)~\citep{ma2025large,lin2025explore,duan-etal-2024-shifting,tian2025reinforcementmidtraining,jiang2025koreenhancingknowledgeinjection,zhang2023spot}, reasoning~\citep{duan2024gtbench}, and dialogue~\citep{duan2025guidellm}. 
Their impressive generative ability has driven widespread deployment in real-world applications~\citep{bi-etal-2025-llava,bi2025prismselfpruningintrinsicselection,peng2025visualinputcompressedvisual}. 
However, LLMs remain prone to hallucinations and factual errors~\citep{farquhar2024detecting,wang2025ascd,yang2026alignsaeconceptalignedsparseautoencoders,rong2025backdoor}, raising serious concerns about their trustworthiness in high-stakes applications such as healthcare and finance~\citep{li2025investorbench,wang2025word,huang2025loongsynthesizelongchainofthoughts,lu2025mvdebatemultiviewagentdebate}. 
These issues have made reliability assessment and risk control a central challenge for the safe deployment of LLMs~\citep{huang2024position,wang-etal-2024-conu,bi2025cot,wan2025magicwordssharpnessawareprompt}.

Uncertainty quantification (UQ) can address these issues by estimating how likely a model prediction is to be untrustworthy. Existing UQ methods, including entropy-based~\citep{kuhn2023semantic,duan-etal-2024-shifting,wang2025word}, consistency-based~\citep{wang-etal-2024-conu}, and self-evaluation-based approaches~\citep{xiong2024can,yona-etal-2024-large}, often provide useful risk signals. 
However, these methods remain heuristic: their uncertainty estimates cannot perfectly separate incorrect from correct outputs~\citep{wang2025coin}, and they do not offer rigorous finite-sample guarantees. 
More importantly, previous UQ methods are mainly designed for a point prediction. 
Such formulation is fundamentally limited for LLMs, because an inadmissible most likely generation (MLG) does not necessarily indicate that the model is incapable of producing a valid response; correct responses may still exist in the output distribution and could be uncovered via sampling~\citep{wang2025word}. 
Therefore, rather than quantifying uncertainty solely for a single point prediction, it is more appropriate to investigate risk-controlled set-valued prediction, where trustworthiness is assessed over a prediction set~\citep{bates2021distribution,wang2025sample}. 
This perspective better reflects the generative nature of LLMs, while also calling for principled calibration methods that can translate model-derived uncertainty scores into statistically valid reliability guarantees.

Split conformal prediction (SCP) provides a promising framework for moving beyond heuristic UQ and establishing statistical reliability guarantees~\citep{bates2021distribution,angelopoulos2023conformal,angelopoulos2024theoretical}. 
Under a mild assumption of exchangeability, SCP offers distribution-free coverage guarantees by calibrating a nonconformity threshold over a held-out calibration set~\citep{wang-etal-2025-sconu,wang-etal-2024-conu}. 
This threshold can be utilized at test time to construct prediction sets that contain the ground-truth with at least a user-specified probability. 
Nonetheless, existing SCP-based methods for LLMs largely remain limited to closed-ended settings~\citep{kumar2023conformal,ye2024benchmark}, or implicitly assume that at least one valid answer appears in every finite sampling set~\citep{wang-etal-2024-conu,kaur2024addressing,wang2025sample}. 
Such assumptions are often violated in open-ended generation scenarios, where the output space is effectively unbounded and finite sampling may fail to produce any admissible response at all. 
As a consequence, standard conformal guarantees become difficult to apply directly in such a practical generation setting.

\begin{figure*}[!t]
    \centering
\includegraphics[width=0.9\linewidth]{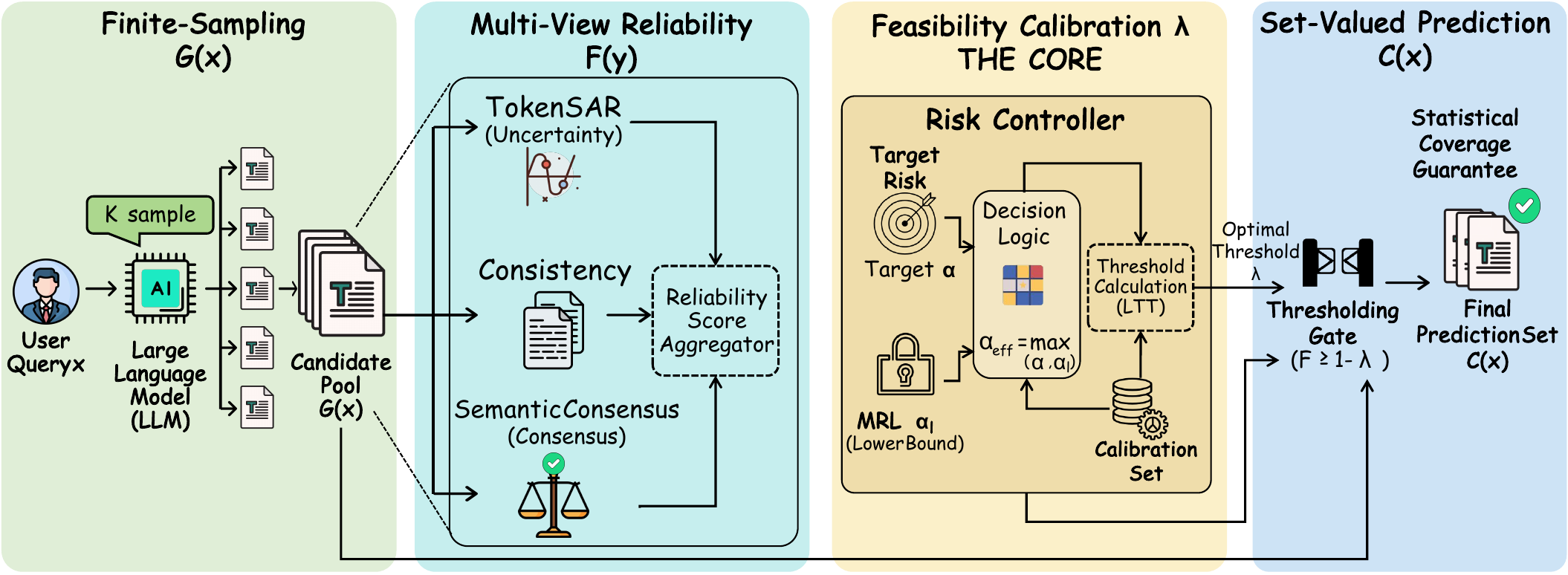}
    \caption{Overview of our feasibility-aware calibration framework with statistically rigorous coverage guarantees. }
    \label{fig: overview}
\end{figure*}

In this paper, we propose a feasibility-aware calibration framework for open-ended generation. Our method consists of two key components. 
Firstly, given a user-specified sampling budget, we employ a held-out calibration set to derive the minimum risk level (MRL) that can be achieved at test time under finite sampling. 
This step explicitly characterizes the feasibility limit imposed by LLM's capability, the fact that even an ideal selection rule cannot guarantee arbitrarily low risk when no correct response is present in the sampled candidate set. 
In this sense, MRL serves as the fundamental prerequisite for any valid risk-controlled prediction guarantee in open-ended generation tasks. 

Secondly, we develop a learn-then-test (LTT) calibration procedure to conformalize an uncertainty threshold. 
Specifically, we construct prediction sets by selecting candidate answers based on an initialized threshold, and define a set-level miscoverage loss for each calibration data point. 
We then calibrate this threshold so that the empirical loss satisfies a finite-sample validity condition on calibration samples under exchangeability. 
The data-driven threshold is then applied to construct prediction sets for new test data, without assuming that every finite sampling set contains an admissible answer. 
Furthermore, when the user-specified target risk level is no smaller than the derived MRL, the resulting prediction sets satisfy the desired coverage guarantee. 
The overview of our framework is illustrated in Figure~\ref{fig: overview}.

We validate our framework on six popular NLG datasets, including medical and financial QA, leveraging five representative LLMs. 
Experimental results show that set-valued prediction significantly outperforms MLG-based point prediction, confirming the advantage of set-valued inference in open-ended QA settings. 
Furthermore, whenever the user-specified target risk level is above the derived minimum risk level, our method consistently attains valid coverage guarantees on the test set. 
Beyond coverage, we show that under rigorous risk constraints, the average prediction set size (APSS) serves as an informative benchmark for characterizing LLM uncertainty. 
We also incorporate semantic deduplication to remove redundant responses, which improves the predictive efficiency of the resulting prediction sets.

Our contributions can be summarized as follows:
\begin{itemize}
    \item We propose a feasibility-aware conformal calibration framework for open-ended generation, which practically accounts for the finite-generation failure mode where no correct answer appears in the sampling set.

    \item We develop the minimum feasible risk level under a given sampling budget over a held-out calibration set. 

    \item We conduct data-driven conformalization to calibrate a rigorous threshold, which yields prediction sets with finite-sample coverage guarantees whenever the user-specified risk level is above the MRL. 
\end{itemize}

% More fundamentally, conventional LLM usage typically treats generation as a point prediction problem, returning only the single most likely response to the user. 
% Such a paradigm can underestimate the model's true capability: although the top-ranked answer may be incorrect, acceptable responses may still exist in the broader generation space and can often be uncovered through repeated sampling. 
% This observation motivates transitioning from point prediction to set-valued prediction, where the model outputs a set of candidate responses rather than a single answer. Ideally, such a prediction set should satisfy a rigorous coverage guarantee, namely, it should contain a correct answer with a user-specified probability.
\section{Related Work}

\paragraph{Split Conformal Prediction.}
SCP is a distribution-free and model-agnostic framework that provides user-specified coverage guarantees for ground-truth labels in \emph{closed-ended} tasks~\citep{angelopoulos2023conformal}, like classification~\citep{angelopoulos2021uncertainty} and image segmentation~\citep{tan2025conformal}. 
Under exchangeability, SCP defines a nonconformity score for each calibration data point (e.g., one minus the softmax probability assigned to the true class in classification) and computes a quantile-based threshold from the calibration set. At test time, this threshold is used to construct prediction sets by retaining candidate labels whose nonconformity scores are sufficiently small. 
The resulting prediction sets are guaranteed to contain the ground-truth with at least a target probability~\citep{angelopoulos2024theoretical}, and improve human decision making~\citep{cresswell2024conformal}. 

\paragraph{Split Conformal Prediction for Language Generation.} 
Recent studies have begun to extend SCP to language generation tasks~\citep{campos-etal-2024-conformal,zhou2025conformal,wang-etal-2025-sconu}. 
One line of work focuses on \emph{point prediction}~\citep{wang2025coin}, where conformal factuality frameworks are proposed to improve the trustworthiness of a single generated response by removing unsupported or unreliable sub-claims, thereby ensuring that at least a user-specified proportion of responses on the test set are factual~\citep{cherian2024large,mohri2024language,rubin-toles2025conformal}. 
However, such methods remain fundamentally constrained by the point prediction setting: a single MLG may not fully reflect the underlying capability of the model, and modifying the original output to satisfy factuality constraints may result in answers that are overly vague or less informative to users. 

Another line of work explores set-valued prediction for language generation~\citep{quach2024conformal}, aiming to construct prediction sets that provide coverage guarantees for admissible responses. 
While this direction is better aligned with the generative nature of LLMs, existing methods either focus on closed-ended settings, such as multiple-choice question answering (MCQA)~\citep{kostumov2024uncertainty,ye2024benchmark,kumar2023conformal}, or assume that both calibration and test examples can produce a correct or admissible answer through finite sampling~\citep{wang-etal-2024-conu,wang2025sample,kaur2024addressing,wang2025copu}. 
Such assumptions substantially limit their applicability in practical open-ended generation, where the output space is unbounded and finite sampling may fail to produce any admissible response. 

Since admissible-answer coverage is not guaranteed under finite sampling in practical open-ended generation, the standard nonconformity score cannot be directly defined for every calibration example. In particular, nonconformity in SCP measures the discrepancy between the model prediction and the ground-truth, but such a comparison becomes unavailable when no admissible response appears in the sampled candidate set. As a result, the standard SCP framework cannot be directly applied in this setting. 
Next, we introduce our feasibility-aware conformal calibration framework. 

\section{Methodology}

\subsection{Notations and Problem Formulation}
Denote by $\mathcal{G}: \mathcal{X} \rightarrow \mathcal{Y}$ a generative LLM, where $\mathcal{X}$ and $\mathcal{Y}$ denote the input and output spaces for language generation tasks, respectively. 
Following the SCP paradigm~\citep{angelopoulos2024theoretical}, we hold out a calibration set $\{(x_i, y_i^*)\}_{i=1}^{n}$, where $x_i \in \mathcal{X}$ is the $i$-th calibration query and $y_i^* \in \mathcal{Y}$ is its ground-truth answer. Let $x_{n+1}$ denote a test query with unknown ground-truth $y_{n+1}^*$.

Our goal is to construct a prediction set for $x_{n+1}$ such that it contains at least one admissible answer that is semantically aligned with $y_{n+1}^*$ with high probability. Formally, let $\alpha \in (0,1)$ denote the user-specified risk level. We define $A(\cdot,\cdot)$ as a task-specific alignment function, where $A(y, y^*) = 1$ if the response $y$ is semantically equivalent to the ground-truth answer $y^*$, and $A(y, y^*) = 0$ otherwise. Our objective is to calibrate a set-valued predictor $\mathcal{C}_{\alpha}$ such that
\begin{equation}
    \Pr \big( \exists y \in \mathcal{C}_{\alpha}(x_{n+1}) : A(y, y_{n+1}^*) = 1 \big) \geq 1 - \alpha,
\end{equation}
where $\mathcal{C}_{\alpha}(\cdot)$ is constructed from the calibration set. 

\subsection{Minimum Feasible Risk Level}

As discussed above, user-specified coverage guarantees can be fundamentally compromised by the finite-sampling failure mode: an admissible answer may not appear in the sampled candidate set. 
In practical deployment, the difficulty of both calibration and test queries is unknown, and the deployed LLM has limited capability, making it unrealistic to guarantee perfect coverage for every input. 
Consequently, under a fixed sampling budget, not every user-specified risk level (i.e., $\alpha$) is achievable. 
This motivates the introduction of the MRL, capturing the irreducible risk that cannot be eliminated by any downstream set construction procedure. 

Let the sampling budget be $K$. 
For each calibration data point, we obtain the candidate set $\{\hat{y}_{k}^{(i)}\}_{k=1}^{K}$ from the output space $\mathcal{G}(x_i)$. 
Given the test query $x_{n+1}$, we also construct the sampling set $\{\hat{y}_{k}^{(n+1)}\}_{k=1}^{K}$. 
Since any prediction set $\mathcal{C}_{\alpha}(x_{n+1})$ is a subset of the sampling set, i.e., $\mathcal{C}_{\alpha}(x_{n+1}) \subseteq \{\hat{y}_{k}^{(n+1)}\}_{k=1}^{K}$, the MRL can be defined as the irreducible risk when the prediction set is maximized to include all candidates. 
Formally, let $a_l$ denote this lower bound; then
\begin{equation}
    a_l := \Pr \Big( \forall \hat{y} \in \{\hat{y}_{k}^{(n+1)}\}_{k=1}^{K}, \; A(\hat{y}, y_{n+1}^*) = 0 \Big),
\end{equation}
which corresponds to the probability that none of the sampled candidates aligns with the ground-truth answer.

To relate this probability to the calibration set, we define for each calibration example an indicator of finite-sampling failure:
\begin{equation}
    l_i := \mathbf{1}\Big\{ \forall k \in [K],\; A(\hat{y}_k^{(i)}, y_i^*) = 0 \Big\}, \quad i = 1, \dots, n.
\end{equation}

\paragraph{Exchangeability.} 
We assume that the calibration examples and the test example are {exchangeable}, meaning that for any permutation $\pi$ of $\{1,\dots,n+1\}$, the joint distribution satisfies $(l_{\pi(1)},\dots,l_{\pi(n+1)}) \overset{d}{=} (l_1,\dots,l_{n+1})$. 
Intuitively, this suggests that the test miscoverage indicator $l_{n+1}$ is statistically indistinguishable from any calibration miscoverage loss $l_i$~\citep{angelopoulos2024conformal,farinhas2024nonexchangeable}.

\paragraph{Expectation under exchangeability.} 
Under this mild assumption~\citep{campos-etal-2024-conformal}, the marginal expectation of the test miscoverage loss can be expressed as
\begin{equation}
\begin{split}
    \mathbb{E}[l_{n+1}]
    &= \mathbb{E}\Big[\mathbf{1}\Big\{ \forall k \in [K],\; A(\hat{y}_k^{(n+1)}, y_{n+1}^*) = 0 \Big\} \Big] \\
    &= \frac{1}{n+1} \sum_{i=1}^{n+1} l_i \\
    &= \frac{1}{n+1} \sum_{i=1}^{n} l_i + \underbrace{\frac{1}{n+1} l_{n+1}}_{l_{n+1}=\text{0 or 1}}.
\end{split}
\end{equation}

By upper bounding the unknown contribution of the test loss by $1$, we obtain the following finite-sample lower bound on any achievable target risk under finite sampling:
\begin{equation}
    \alpha_l := \frac{1}{n+1} \sum_{i=1}^{n} l_i,
\end{equation}
which forms the foundation for calibrating a feasible uncertainty threshold in the subsequent step. 

\subsection{Set-Valued Predictor}

After deriving the MRL, we next construct a set-valued predictor in a data-driven manner. 
Our approach follows a learn-then-test principle~\citep{angelopoulos2025learn,wang2025lec}: we first calibrate a reliability threshold on the held-out calibration set, and then apply the calibrated threshold at test time to obtain finite-sample control of the expected miscoverage loss. 

To this end, we introduce a confidence evaluation function
$F: \mathcal{Y} \rightarrow [0,1]$, which assigns each sampled candidate response a reliability score, where a larger value represents higher trustworthiness. 
Equivalently, $F$ can be viewed as the inverse of an uncertainty estimator. 
For notational simplicity, we write $F(\hat{y}_k^{(i)})$, although in general the score may depend on the full sampled candidate set associated with $x_i$.

For a given query $x_i$ with sampled candidate set $\{\hat{y}_k^{(i)}\}_{k=1}^{K}$, and for any threshold parameter $\lambda \in [0,1]$, we formulate the corresponding prediction set as
\begin{equation}
\mathcal{C}_{\lambda}(x_i)
:=
\Big\{
\hat{y}_k^{(i)} \in \{\hat{y}_k^{(i)}\}_{k=1}^{K}
:
F(\hat{y}_k^{(i)}) \ge 1-\lambda
\Big\}.
\end{equation}
Namely, $\mathcal{C}_{\lambda}(x_i)$ retains all sampled responses whose confidence scores exceed the threshold $1-\lambda$. 
Under this parameterization, a larger $\lambda$ yields a larger prediction set, which is convenient for conformal risk calibration.

Based on $\mathcal{C}_{\lambda}(x_i)$, we define the set-level miscoverage loss for the $i$-th example as
\begin{equation}
\ell_i(\lambda)
:=
\mathbf{1}
\Big\{
\forall y \in \mathcal{C}_{\lambda}(x_i),\;
A(y, y_i^*) = 0
\Big\}.
\end{equation}
That is, $\ell_i(\lambda)=1$ if the prediction set contains no admissible answer aligned with the ground-truth, and $\ell_i(\lambda)=0$ otherwise. 
Since enlarging the prediction set can only reduce the probability of miscoverage, $\ell_i(\lambda)$ is non-increasing in $\lambda$. 
Moreover, as a binary loss, it satisfies
\[
\ell_i(\lambda)\in\{0,1\}\subset(-\infty,1].
\]

We then define the empirical calibration loss on average over the calibration set as
\begin{equation}
\widehat{\mathcal{L}}_n(\lambda)
:=
\frac{1}{n}\sum_{i=1}^{n}\ell_i(\lambda).
\end{equation}
Following prior conformal risk control frameworks~\citep{angelopoulos2023conformal,angelopoulos2024conformal,angelopoulos2024theoretical}, we calibrate the threshold by selecting
\begin{equation}
\begin{split}
\hat{\lambda}
&=
\inf\left\{
\lambda:
\frac{n}{n+1}\widehat{\mathcal{L}}_n(\lambda)+\frac{1}{n+1}\le \alpha
\right\} \\
&=
\inf\left\{
\lambda:
\widehat{\mathcal{L}}_n(\lambda)\le \alpha-\frac{1-\alpha}{n}
\right\},
\end{split}
\end{equation}
where $\alpha$ is the user-specified target risk level.

The calibrated threshold $\hat{\lambda}$ is then applied to a new test query $x_{n+1}$ to construct
\begin{equation}
\mathcal{C}_{\hat{\lambda}}(x_{n+1})
:=
\Big\{
\hat{y}_k^{(n+1)} \in \{\hat{y}_k^{(n+1)}\}_{k=1}^{K}
:
F(\hat{y}_k^{(n+1)}) \ge 1-\hat{\lambda}
\Big\}.
\end{equation}
The following theorem shows that the calibrated threshold yields the desired marginal coverage guarantee at test time.

\begin{theorem}[Coverage Guaranteed Threshold]
\label{thm:coverage-guarantee}
Assume that the augmented tuples for $n$ calibration samples and the test data, $\Big\{
(x_i, y_i^*, \{\hat{y}_k^{(i)}\}_{k=1}^{K})
\Big\}_{i=1}^{n+1}$, are exchangeable. Let $\hat{\lambda}$ be defined as above. Then, for any target risk level $\alpha \ge \alpha_l$, the calibrated prediction set $\mathcal{C}_{\hat{\lambda}}(x_{n+1})$ satisfies
\begin{equation}
\mathbb{E}\big[\ell_{n+1}(\hat{\lambda})\big]\le \alpha.
\end{equation}
Equivalently,
\begin{equation}
\Pr\Big(
\exists y\in\mathcal{C}_{\hat{\lambda}}(x_{n+1})
:
A(y,y_{n+1}^*)=1
\Big)
\ge 1-\alpha.
\end{equation}
\end{theorem}

For each fixed $\lambda\in[0,1]$, the loss $\ell_i(\lambda)$ is a measurable function of the augmented tuple $(x_i, y_i^*, \{\hat{y}_k^{(i)}\}_{k=1}^{K})$. 
Hence, by the exchangeability of the augmented tuples, the induced loss sequence, $\{\ell_i(\lambda)\}_{i=1}^{n+1}$, is also exchangeable for every fixed $\lambda$~\citep{angelopoulos2023conformal,angelopoulos2024theoretical}. 
Moreover, by construction, $\ell_i(\lambda)\in[0,1]$ and is non-increasing in $\lambda$. 
Therefore, we have
\begin{equation}
\begin{split}
    \mathbb{E}\big[\ell_{n+1}(\hat{\lambda})\big] &= \frac{1}{n+1} \sum_{i=1}^{n+1} \ell_i(\hat{\lambda})\\
    & = \frac{n}{n+1}\widehat{\mathcal{L}}_n(\hat{\lambda})+\frac{\ell_{n+1}(\hat{\lambda})}{n+1}\\
    & \leq \frac{n}{n+1}\widehat{\mathcal{L}}_n(\hat{\lambda})+\frac{1}{n+1}\\
    & \leq \alpha
\end{split}.
\end{equation}
This completes the proof of Theorem~\ref{thm:coverage-guarantee}. 

Finally, the condition $\alpha\ge \alpha_l$ ensures that the target risk level is feasible under finite sampling, as otherwise no downstream set construction rule can overcome the irreducible risk characterized by the MRL.

\paragraph{Reliability scoring function.}
Considering the confidence function $F$, we adopt a multi-view reliability scoring scheme that aggregates self-uncertainty, cross-sample consistency, and semantic consensus within the sampled candidate set. For the $i$-th query $x_i$ with sampled candidates $\{\hat{y}_k^{(i)}\}_{k=1}^{K}$, we first construct a pairwise semantic similarity matrix $\mathbf{S}^{(i)} \in [0,1]^{K\times K}$, where
\begin{equation}
    \mathbf{S}^{(i)}_{j,k}
    :=
    s\big((x_i,\hat{y}_j^{(i)}),(x_i,\hat{y}_k^{(i)})\big),
\end{equation}
and $s(\cdot,\cdot)$ denotes a semantic similarity function~\citep{reimers2019sentence}. 
Based on this matrix, we formulate the average consistency score of candidate $\hat{y}_j^{(i)}$ as
\begin{equation}
    \mathrm{AvgSim}_{j}^{(i)}
    :=
    \frac{1}{K-1}\sum_{k\neq j}\mathbf{S}^{(i)}_{j,k},
\end{equation}
measuring how well the candidate agrees with the remaining sampled responses~\citep{wang2023selfconsistency}.

To capture self-uncertainty, we leverage Shift-Attention-to-Relevance (SAR) to compute an uncertainty score~\citep{duan-etal-2024-shifting}, and define
\begin{equation}
    U_{j}^{(i)} := -\,\mathrm{TokenSAR}(x_i,\hat{y}_j^{(i)}),
\end{equation}
so that a larger value corresponds to lower uncertainty and hence higher reliability. Since the raw scales of uncertainty and consistency may vary across queries, we normalize them within each sampled candidate set using $z$-score normalization:
\begin{equation}
    \widetilde{U}_{j}^{(i)}
    :=
    \frac{U_{j}^{(i)}-\mu_i(U)}{\sigma_i(U)+\varepsilon},
    \quad
    \widetilde{S}_{j}^{(i)}
    :=
    \frac{\mathrm{AvgSim}_{j}^{(i)}-\mu_i(\mathrm{AvgSim})}{\sigma_i(\mathrm{AvgSim})+\varepsilon},
\end{equation}
where $\mu_i(\cdot)$ and $\sigma_i(\cdot)$ denote the mean and standard deviation computed over the $K$ sampled candidates for query $x_i$, and $\varepsilon>0$ is a small constant for numerical stability. We then combine the normalized uncertainty and consistency signals into a base quality score:
\begin{equation}
    Q_{j}^{(i)}
    :=
    \sigma\!\left(
    w_u \widetilde{U}_{j}^{(i)} + w_s \widetilde{S}_{j}^{(i)}
    \right),
\end{equation}
where $\sigma(\cdot)$ is the sigmoid function, and $w_u,w_s$ are weighting coefficients.

\begin{figure*}[!t]
    \centering
    \includegraphics[width=\linewidth]{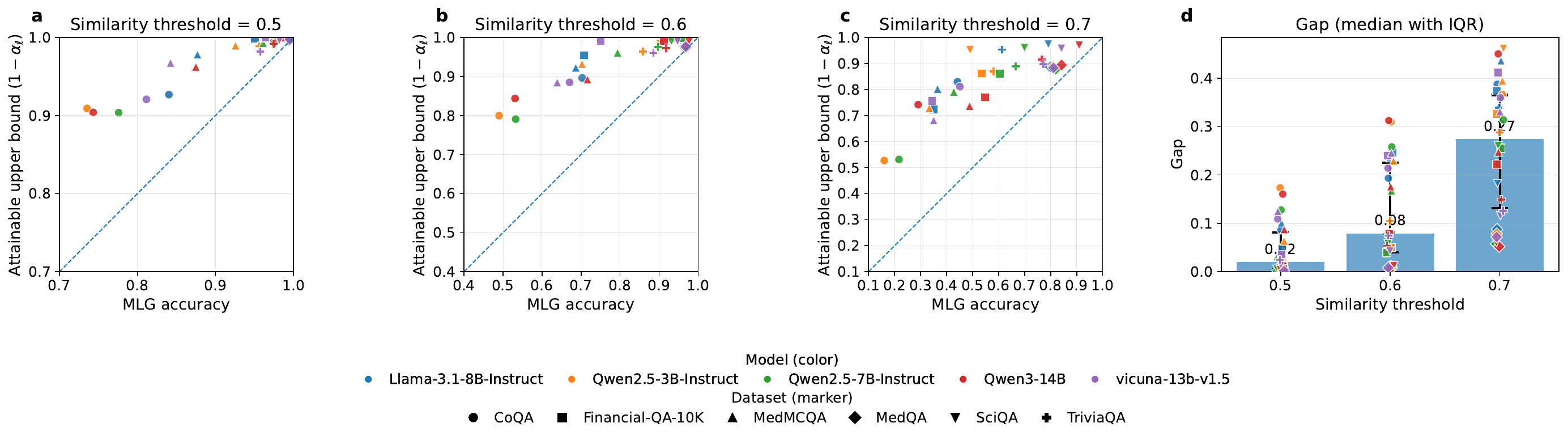}
    \caption{(a-c) Comparison between point-prediction accuracy and set-valued attainability under different semantic matching thresholds. The consistent gap shows that point prediction systematically under-utilizes admissible answers already present in the sampled candidate space, and (d) that this gap widens under stricter semantic evaluation.}
    \label{fig: rq1}
\end{figure*}

Finally, to capture semantic consensus, we partition the sampled candidates into semantically equivalent clusters via bidirectional NLI-based merging~\citep{kuhn2023semantic,wang2025word}. 
Let $c_j^{(i)}$ denote the cluster index of candidate $\hat{y}_j^{(i)}$, let $n_j^{(i)}$ be the size of its cluster, and let $n_{\max}^{(i)}$ be the size of the largest cluster. We define the consensus strength as
\begin{equation}
    \mathrm{CS}_{j}^{(i)}
    :=
    \left(
    \frac{n_j^{(i)}}{n_{\max}^{(i)}}
    \right)^{\gamma_{\mathrm{cons}}},
\end{equation}
where $\gamma_{\mathrm{cons}}>0$ controls the preference for larger consensus clusters. The final reliability score is then given by
\begin{equation}
    F\!\left(\hat{y}_j^{(i)};\{\hat{y}_k^{(i)}\}_{k=1}^{K}\right)
    :=
    \mathrm{CS}_{j}^{(i)} \cdot Q_{j}^{(i)}.
\end{equation}
For simplicity, we write this score as $F(\hat{y}_j^{(i)})$ in the sequel. A larger value of $F(\hat{y}_j^{(i)})$ indicates that the candidate is more reliable and should be ranked higher when constructing the thresholded prediction set.

Overall, our method extends conformal set-valued prediction to open-ended NLG tasks through a feasibility-aware calibrate-then-test pipeline. The first stage derives the MRL, which formalizes the feasibility boundary imposed by finite sampling. The second stage calibrates a reliability threshold from the held-out calibration set and applies it at test time to construct prediction sets with guaranteed marginal coverage whenever the target risk level is feasible. This formulation not only preserves finite-sample statistical validity, but also enables practical and efficient set-valued prediction for large language models in realistic open-ended settings.

\section{Experiments}
\subsection{Experimental Settings}
\paragraph{Datasets.} 
We evaluate our framework on six NLG benchmarks covering diverse generation settings, including open-domain, scientific, medical, and financial tasks: CoQA~\citep{reddy2019coqa}, TriviaQA~\citep{joshi2017triviaqa}, SciQA~\citep{lu2022learn}, MedQA~\citep{jin2021disease}, MedMCQA~\citep{pal2022medmcqa}, and Financial-QA-10K\footnote{\href{https://huggingface.co/datasets/virattt/financial-qa-10K}{https://huggingface.co/datasets/virattt/financial-qa-10K}}. 
Following prior work~\citep{wang-etal-2025-sconu}, we use 4,000 samples from the validation split of CoQA and 4,000 instances from the validation split of TriviaQA. Following the protocol in SAR~\citep{duan-etal-2024-shifting}, we convert SciQA, MedQA, and MedMCQA into open-ended generation tasks and use 4,000 samples from each dataset. For Financial-QA-10K, we use 4,000 samples from the training split. 

\paragraph{Models.} 
We consider five LLMs from three representative open-source model families across five parameter scales, including instruction-tuned Qwen2.5-3B-Instruct, Qwen2.5-7B-Instruct~\citep{bai2023qwen}, LLaMA-3.1-8B-Instruct~\citep{touvron2023llama}, Vicuna-13b-V1.5, and Qwen3-14B. 
All model weights are obtained from Hugging Face. Together, these models allow us to examine the effectiveness of our method across diverse settings. 

\paragraph{Evaluation Metrics.} 
(1) To validate the statistical validity of the data-driven set-valued predictor, we examine whether the empirical coverage rate averaged over the test set exceeds $1-\alpha$ whenever $\alpha \geq \alpha_l$. 
We justify that although set-valued prediction appears less directly comparable to point prediction, the comparison is in fact fair: from each prediction set, one can always select the candidate with the highest confidence score as the preferred response. 
Therefore, set-valued prediction preserves the usability of point prediction while additionally providing alternative admissible responses and a rigorous coverage guarantee. 
Existing studies have further shown that prediction sets are practically useful, especially in reliable generation and human-in-the-loop settings~\citep{hullman2025conformal}. 
(2) We also evaluate the average prediction set size (APSS) to demonstrate the uncertainty-awareness and predictive efficiency of our produced prediction sets~\citep{wang-etal-2024-conu,angelopoulos2024theoretical,angelopoulos2023conformal}.

\begin{figure*}[!t]
  \centering
  \begin{subfigure}[b]{0.195\textwidth}
    \centering
    \includegraphics[width=\textwidth]{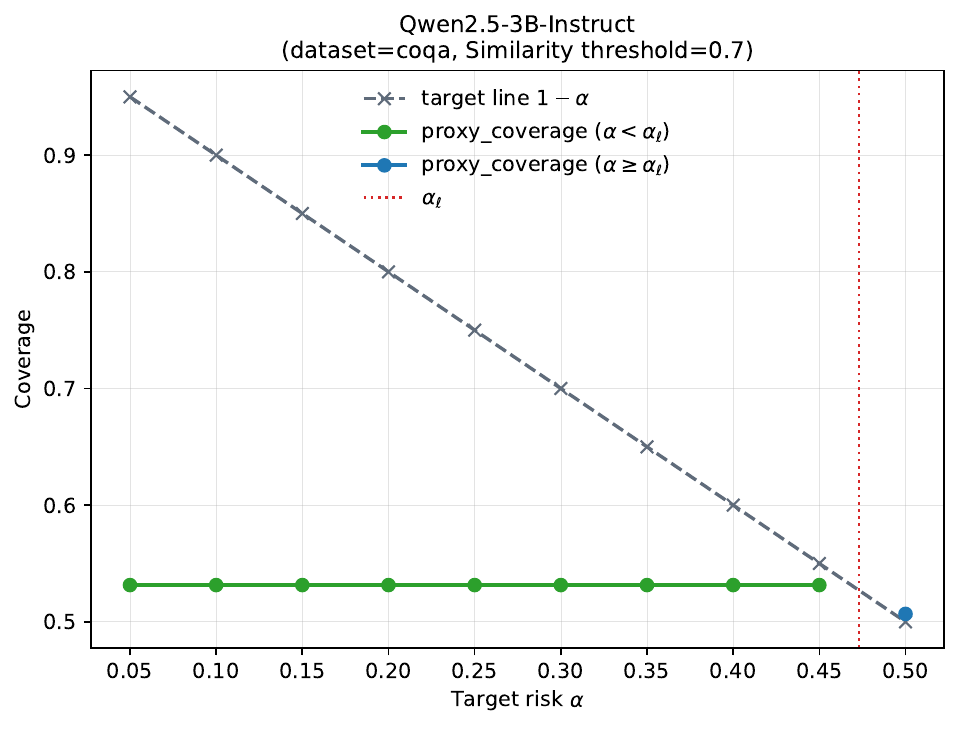}
    % \caption{Qwen2.5-3B}
  \end{subfigure}
  \hfill
  \begin{subfigure}[b]{0.195\textwidth}
    \centering
    \includegraphics[width=\textwidth]{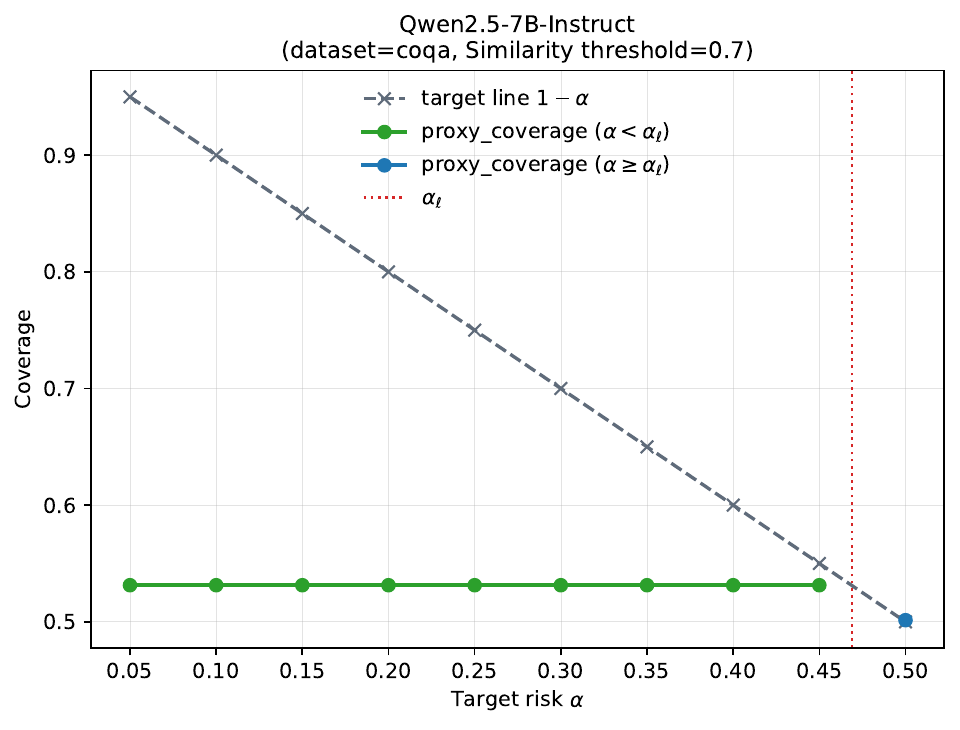}
    % \caption{Qwen2.5-7B}
  \end{subfigure}
  \hfill
  \begin{subfigure}[b]{0.195\textwidth}
    \centering
    \includegraphics[width=\textwidth]{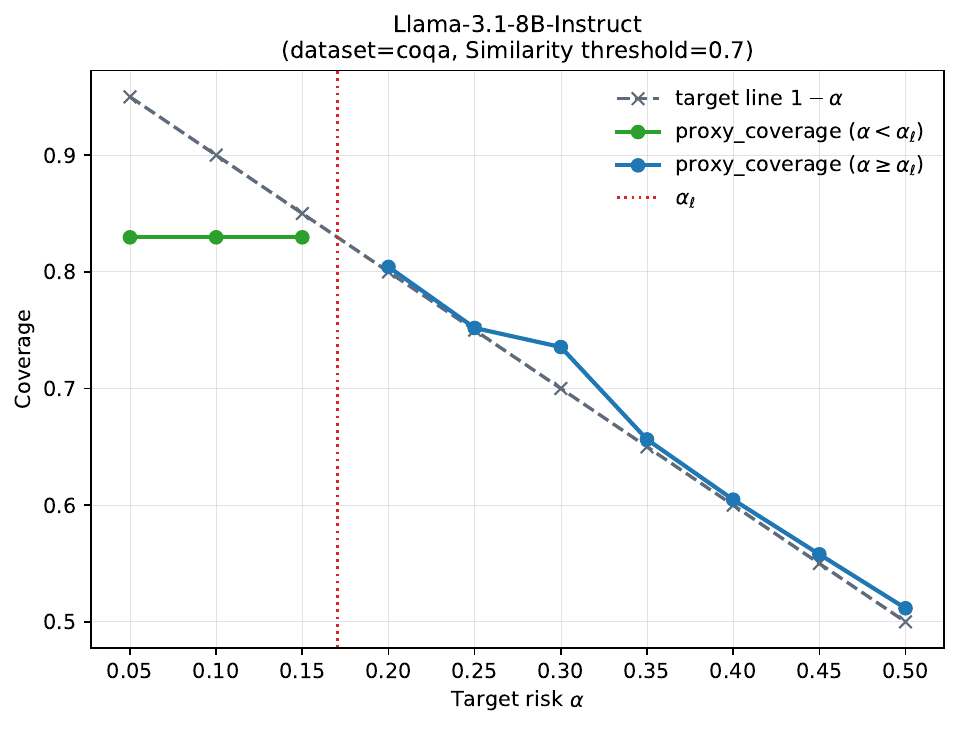}
    % \caption{LLaMA-3.1-8B}
  \end{subfigure}
  \hfill
  \begin{subfigure}[b]{0.195\textwidth}
    \centering
    \includegraphics[width=\textwidth]{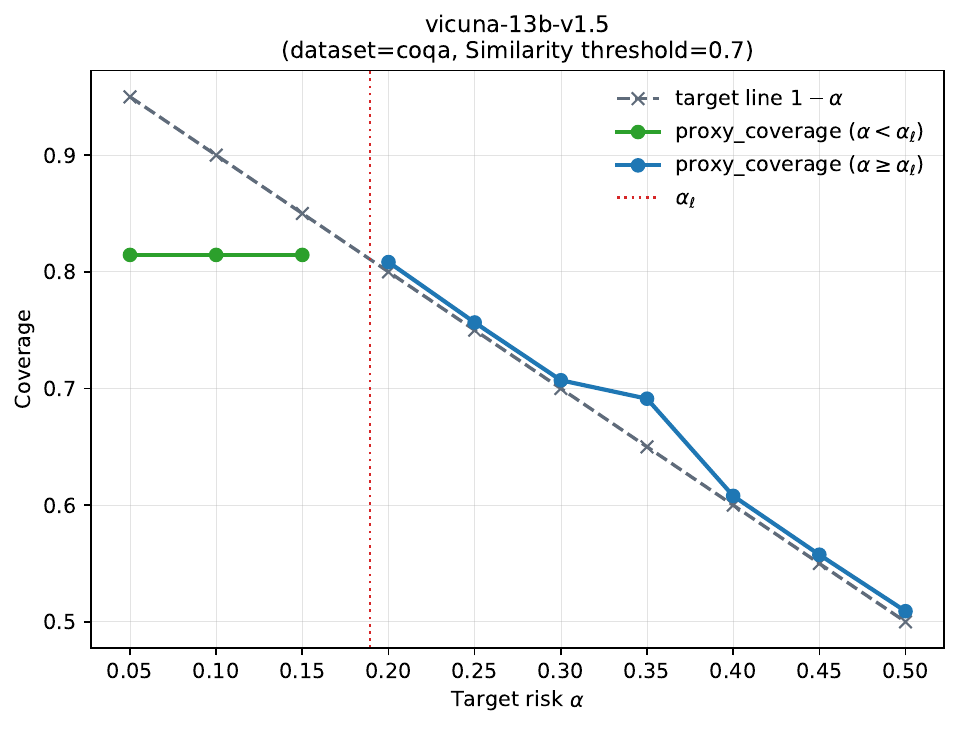}
    % \caption{Vicuna-13B}
  \end{subfigure}
  \hfill
  \begin{subfigure}[b]{0.195\textwidth}
    \centering
    \includegraphics[width=\textwidth]{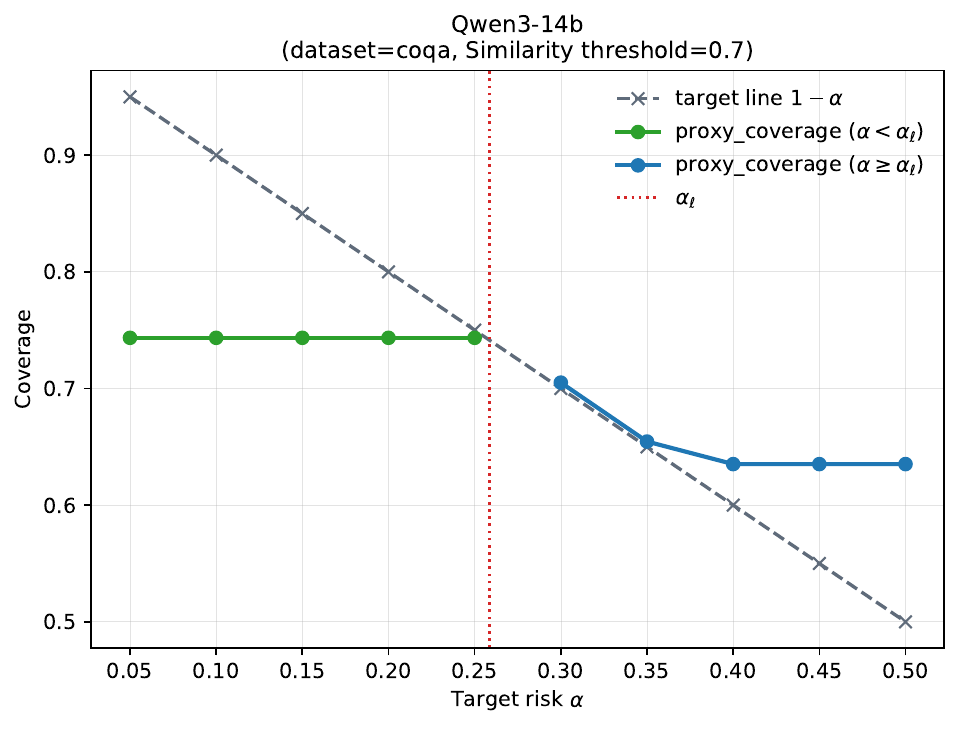}
    % \caption{Qwen3-14B}
  \end{subfigure}

  \begin{subfigure}[b]{0.195\textwidth}
    \centering
    \includegraphics[width=\textwidth]{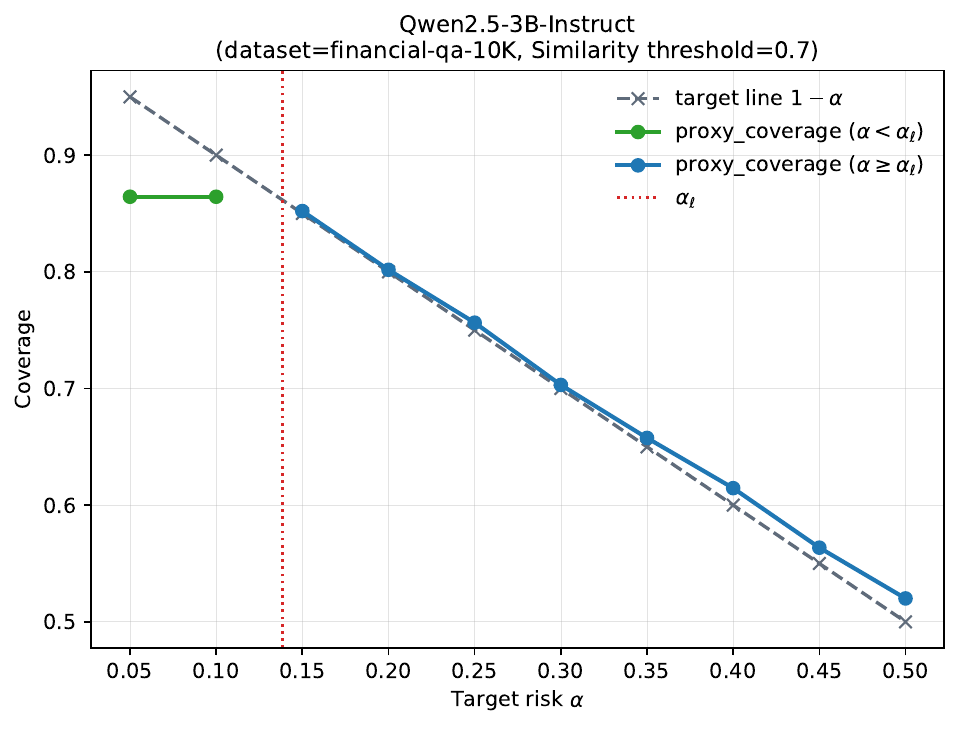}
    % \caption{Qwen2.5-3B}
  \end{subfigure}
  \hfill
  \begin{subfigure}[b]{0.195\textwidth}
    \centering
    \includegraphics[width=\textwidth]{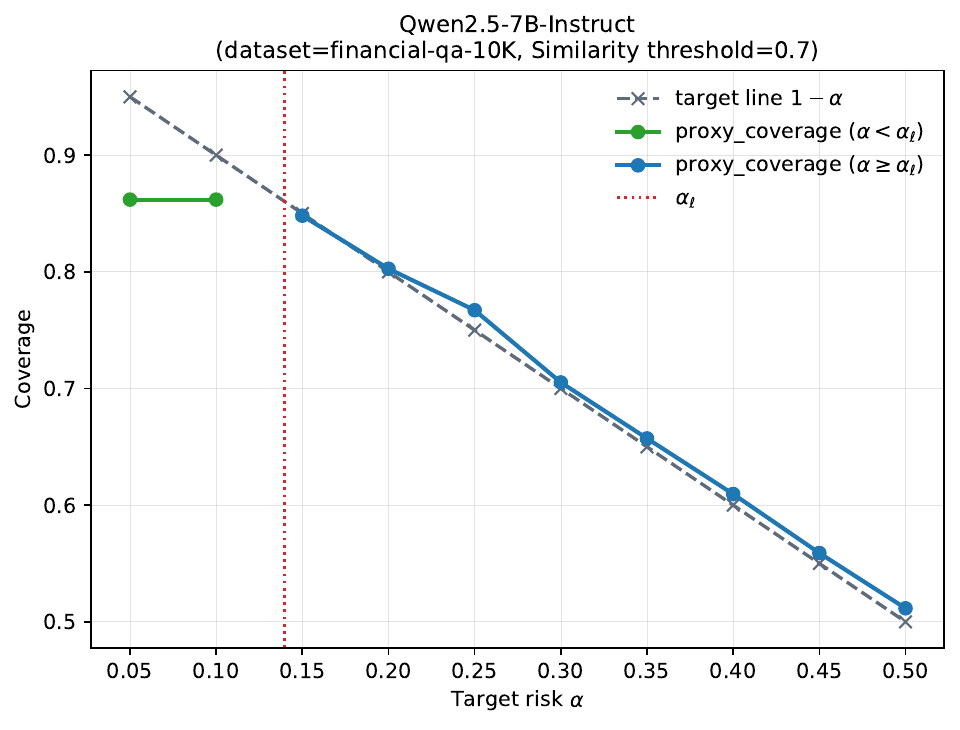}
    % \caption{Qwen2.5-7B}
  \end{subfigure}
  \hfill
  \begin{subfigure}[b]{0.195\textwidth}
    \centering
    \includegraphics[width=\textwidth]{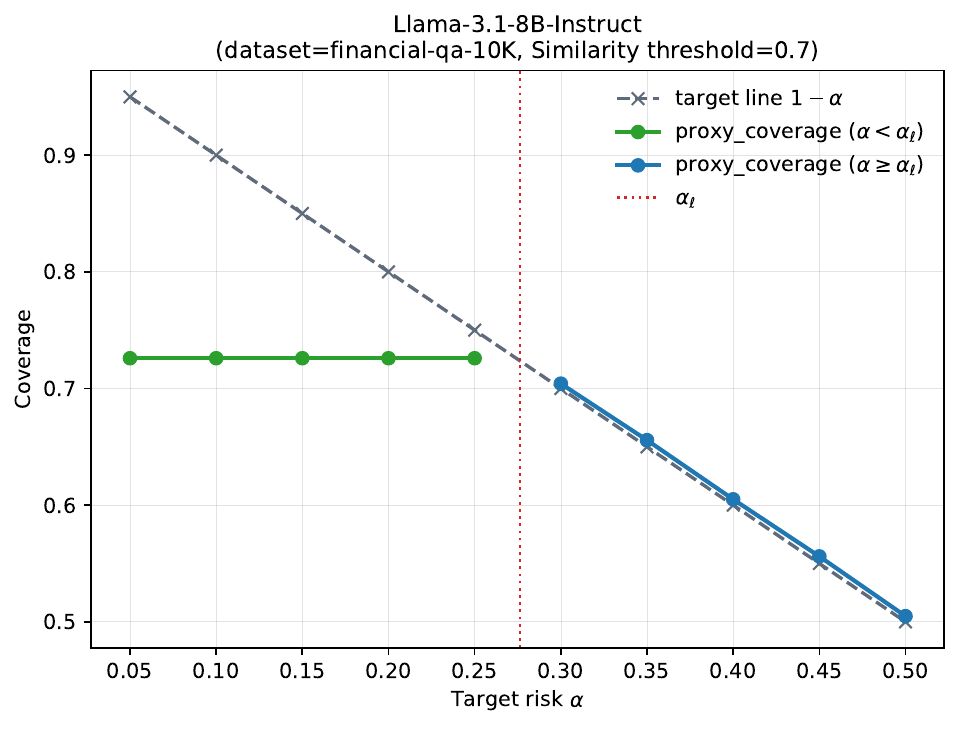}
    % \caption{LLaMA-3.1-8B}
  \end{subfigure}
  \hfill
  \begin{subfigure}[b]{0.195\textwidth}
    \centering
    \includegraphics[width=\textwidth]{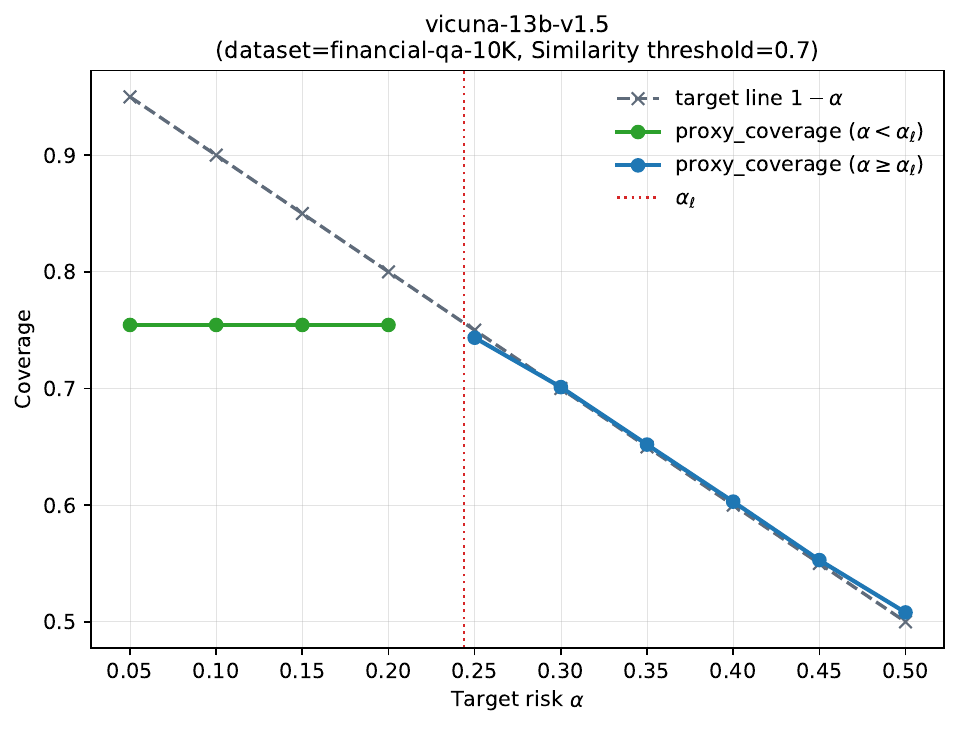}
    % \caption{Vicuna-13B}
  \end{subfigure}
  \hfill
  \begin{subfigure}[b]{0.195\textwidth}
    \centering
    \includegraphics[width=\textwidth]{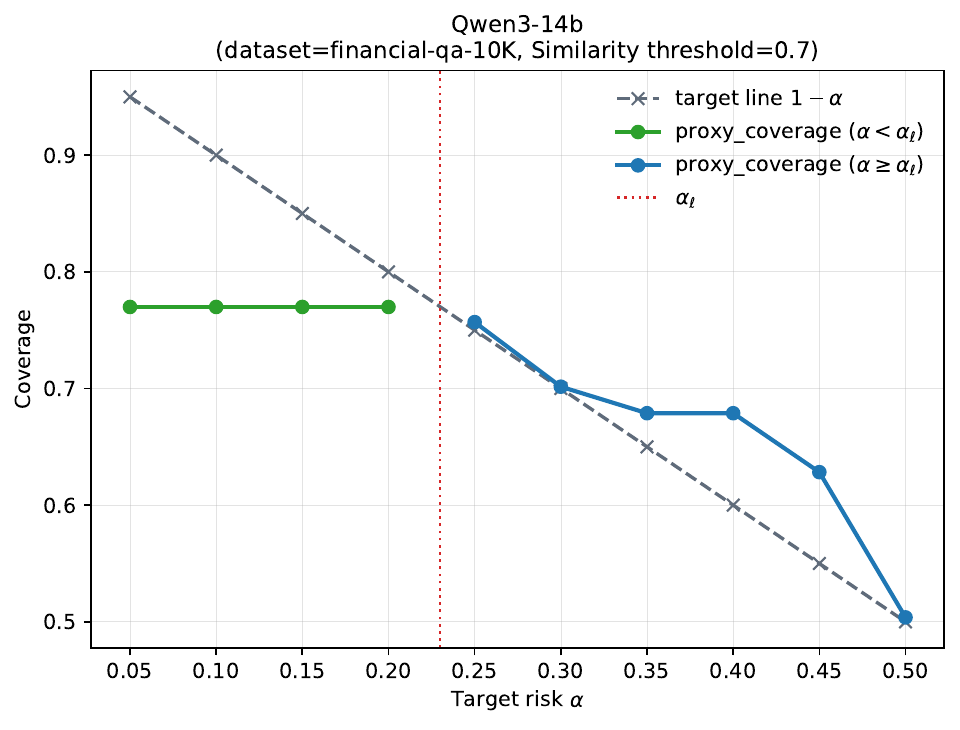}
    % \caption{Qwen3-14B}
  \end{subfigure}

  \begin{subfigure}[b]{0.195\textwidth}
    \centering
    \includegraphics[width=\textwidth]{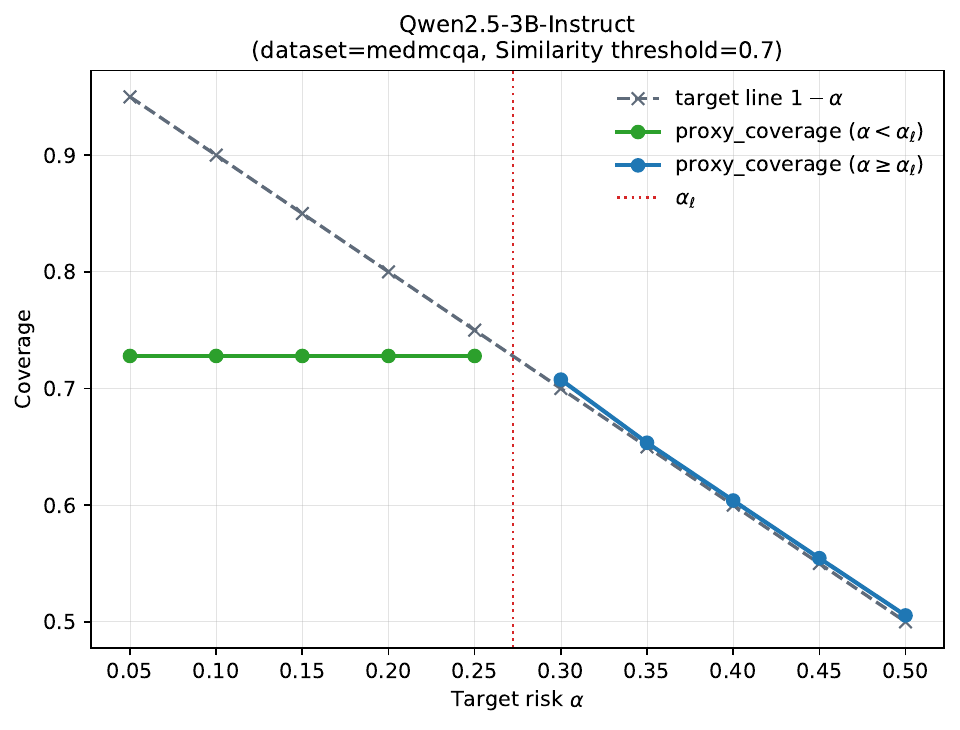}
    % \caption{Qwen2.5-3B}
  \end{subfigure}
  \hfill
  \begin{subfigure}[b]{0.195\textwidth}
    \centering
    \includegraphics[width=\textwidth]{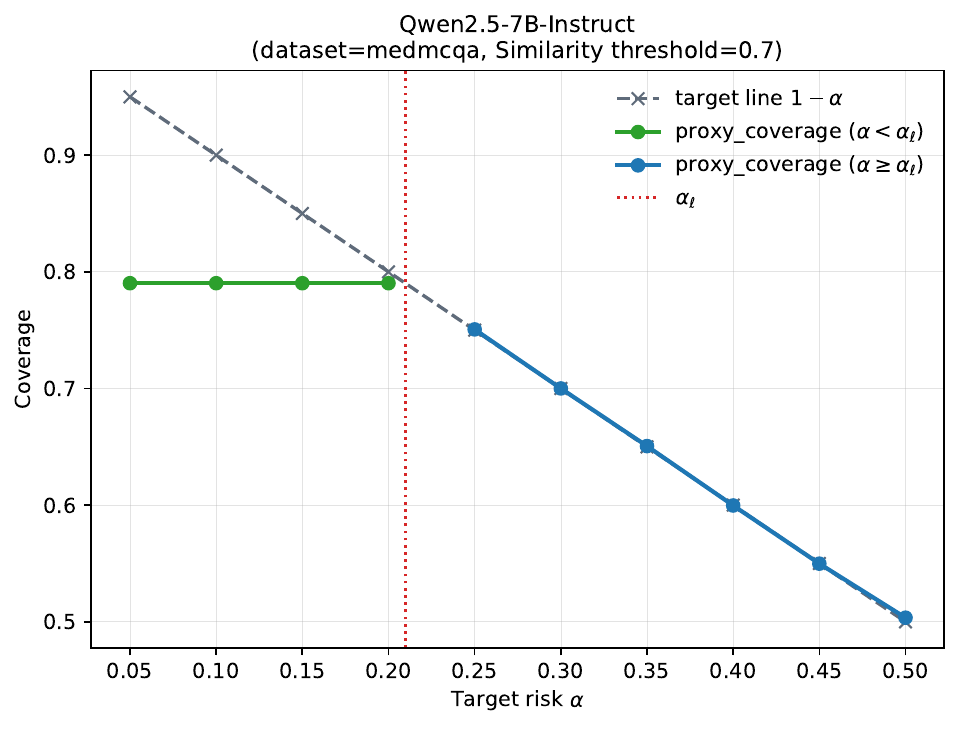}
    % \caption{Qwen2.5-7B}
  \end{subfigure}
  \hfill
  \begin{subfigure}[b]{0.195\textwidth}
    \centering
    \includegraphics[width=\textwidth]{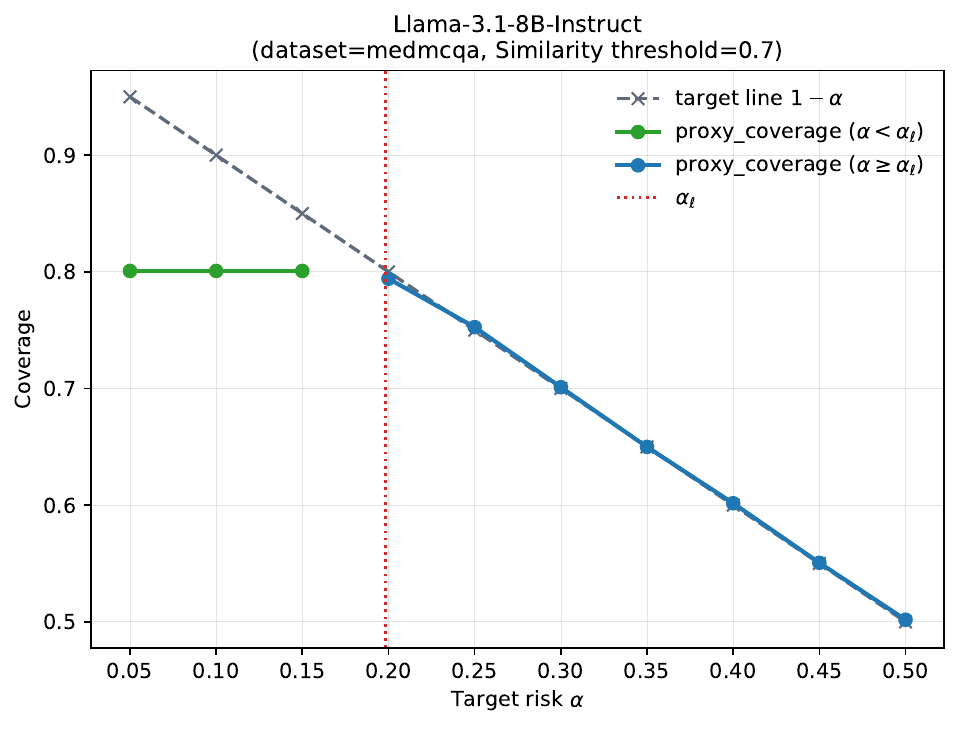}
    % \caption{LLaMA-3.1-8B}
  \end{subfigure}
  \hfill
  \begin{subfigure}[b]{0.195\textwidth}
    \centering
    \includegraphics[width=\textwidth]{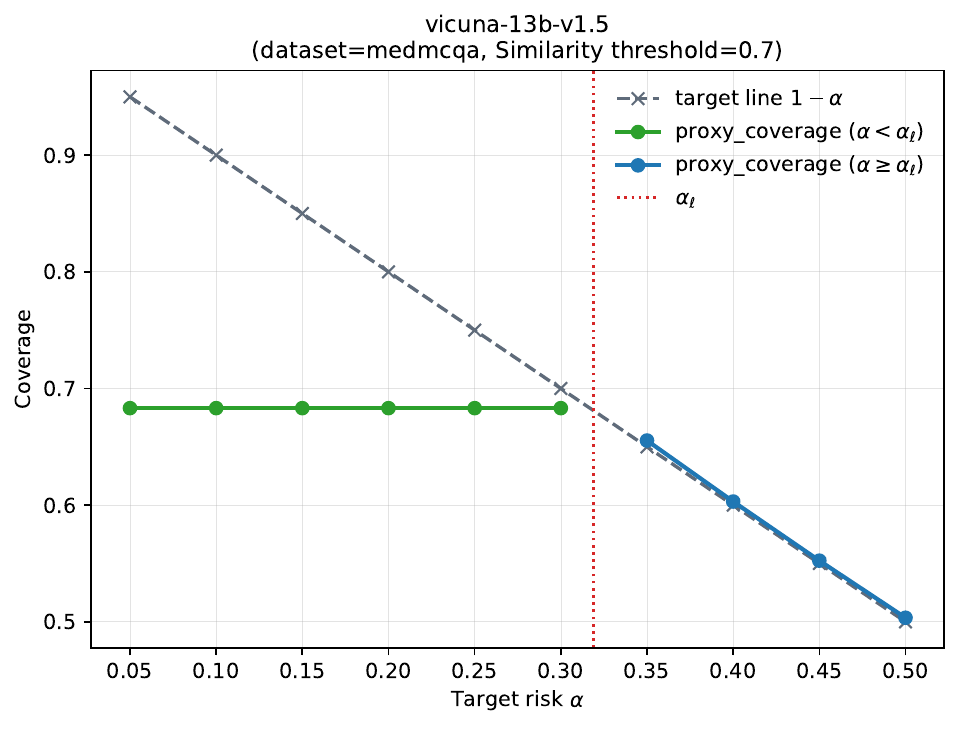}
    % \caption{Vicuna-13B}
  \end{subfigure}
  \hfill
  \begin{subfigure}[b]{0.195\textwidth}
    \centering
    \includegraphics[width=\textwidth]{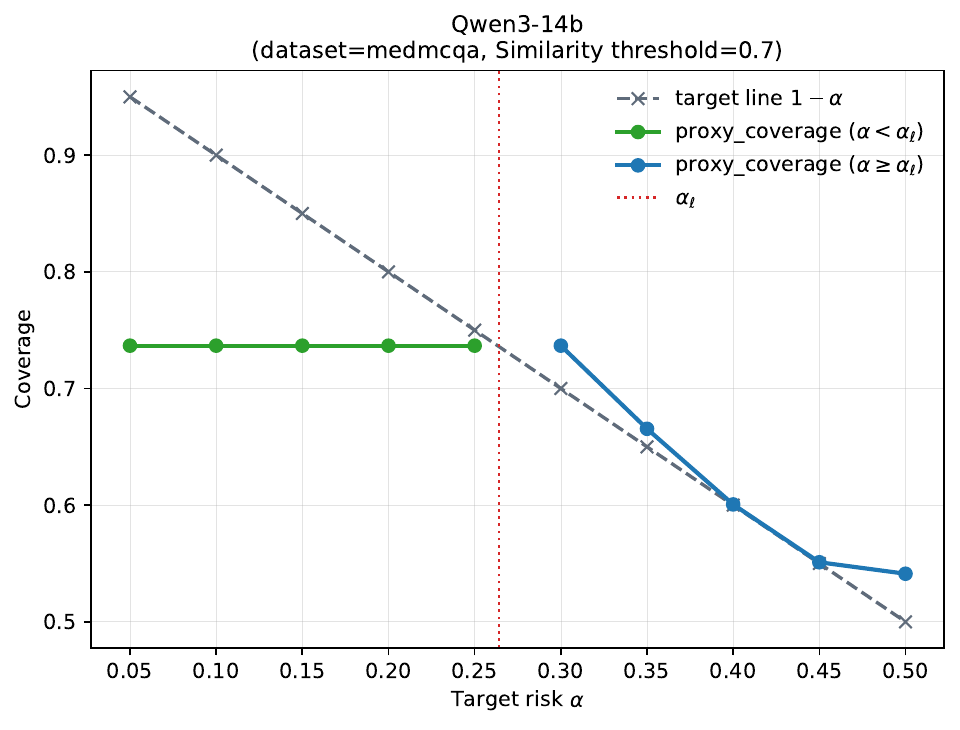}
    % \caption{Qwen3-14B}
  \end{subfigure}

  \begin{subfigure}[b]{0.195\textwidth}
    \centering
    \includegraphics[width=\textwidth]{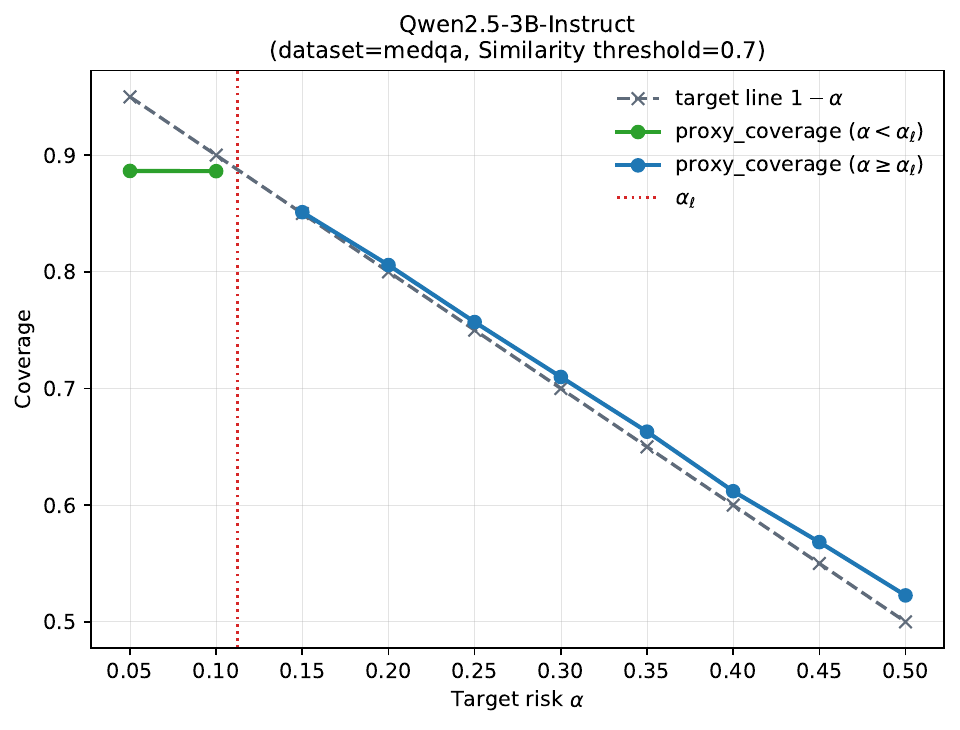}
    % \caption{Qwen2.5-3B}
  \end{subfigure}
  \hfill
  \begin{subfigure}[b]{0.195\textwidth}
    \centering
    \includegraphics[width=\textwidth]{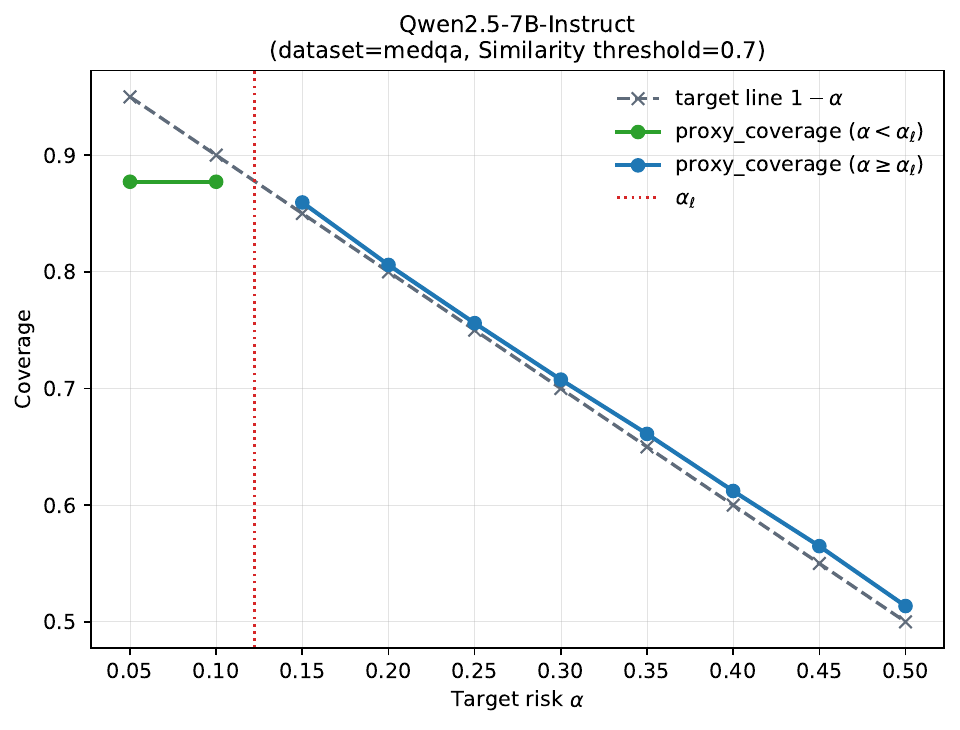}
    % \caption{Qwen2.5-7B}
  \end{subfigure}
  \hfill
  \begin{subfigure}[b]{0.195\textwidth}
    \centering
    \includegraphics[width=\textwidth]{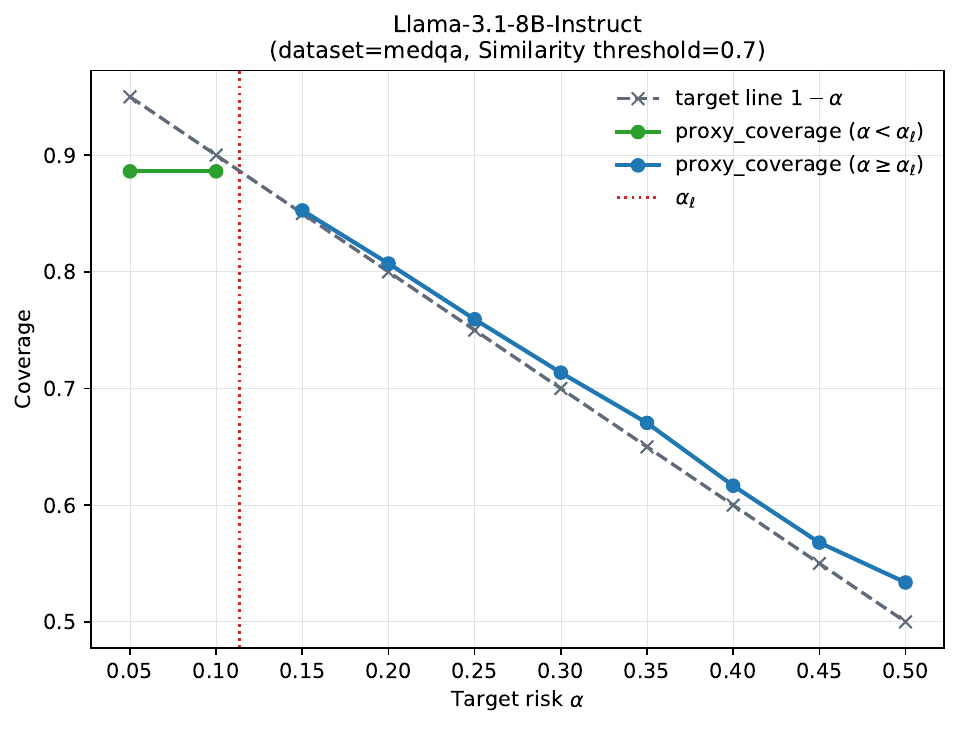}
    % \caption{LLaMA-3.1-8B}
  \end{subfigure}
  \hfill
  \begin{subfigure}[b]{0.195\textwidth}
    \centering
    \includegraphics[width=\textwidth]{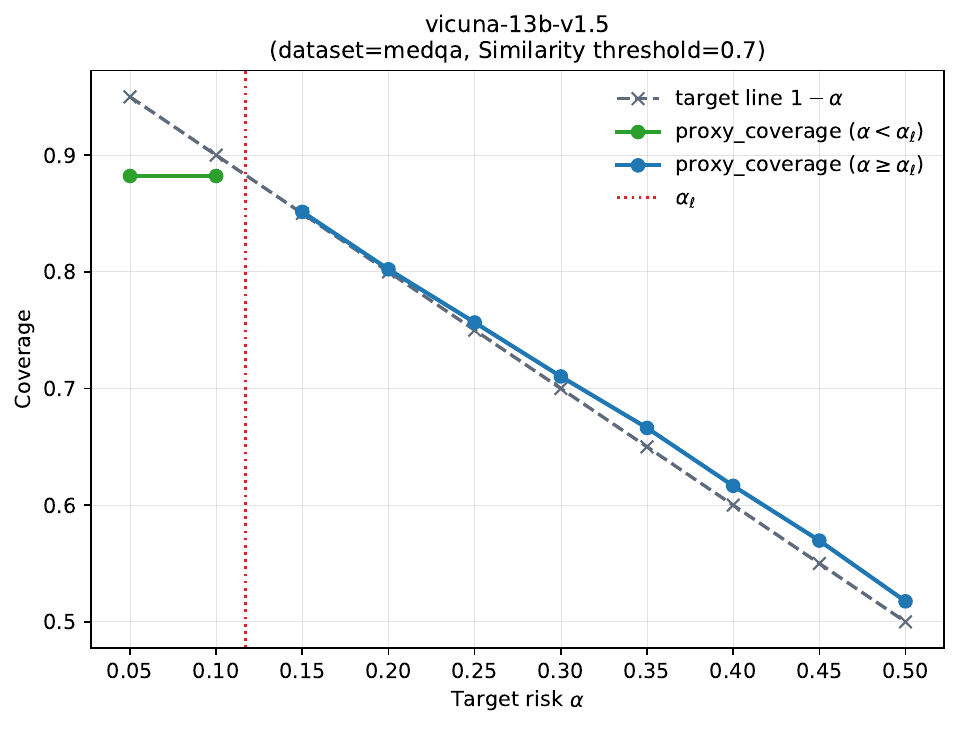}
    % \caption{Vicuna-13B}
  \end{subfigure}
  \hfill
  \begin{subfigure}[b]{0.195\textwidth}
    \centering
    \includegraphics[width=\textwidth]{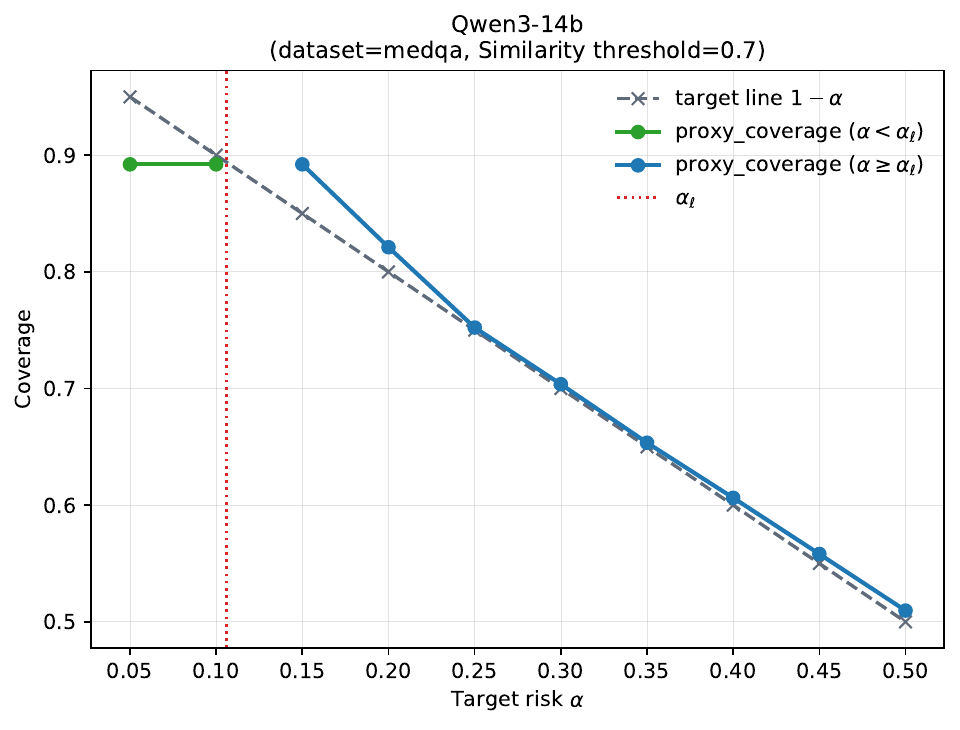}
    % \caption{Qwen3-14B}
  \end{subfigure}

  \begin{subfigure}[b]{0.195\textwidth}
    \centering
    \includegraphics[width=\textwidth]{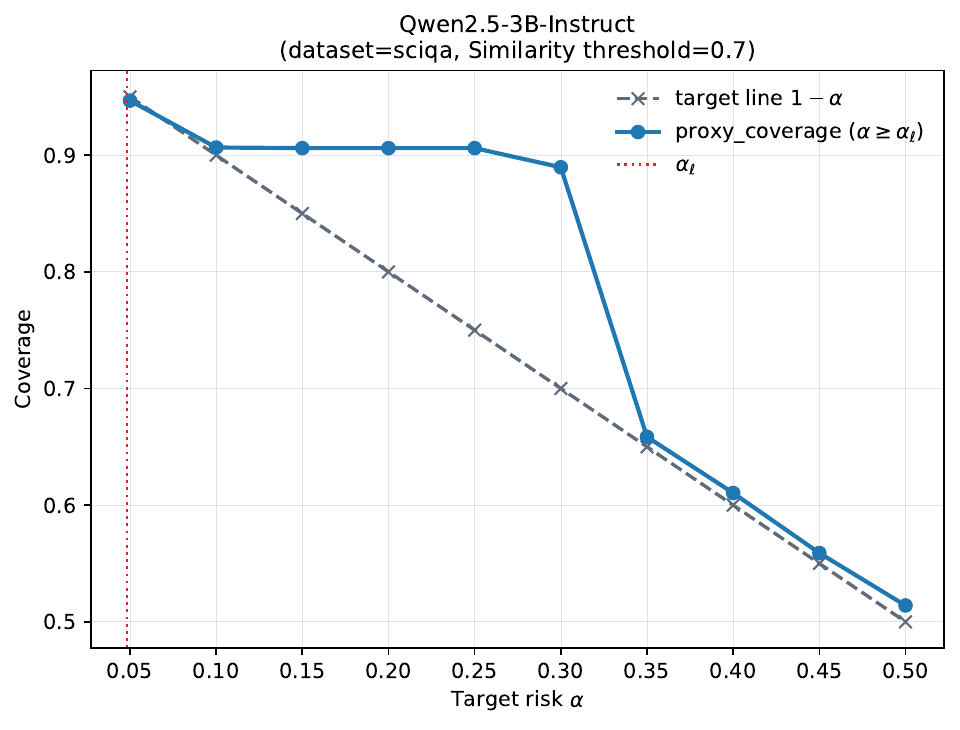}
    % \caption{Qwen2.5-3B}
  \end{subfigure}
  \hfill
  \begin{subfigure}[b]{0.195\textwidth}
    \centering
    \includegraphics[width=\textwidth]{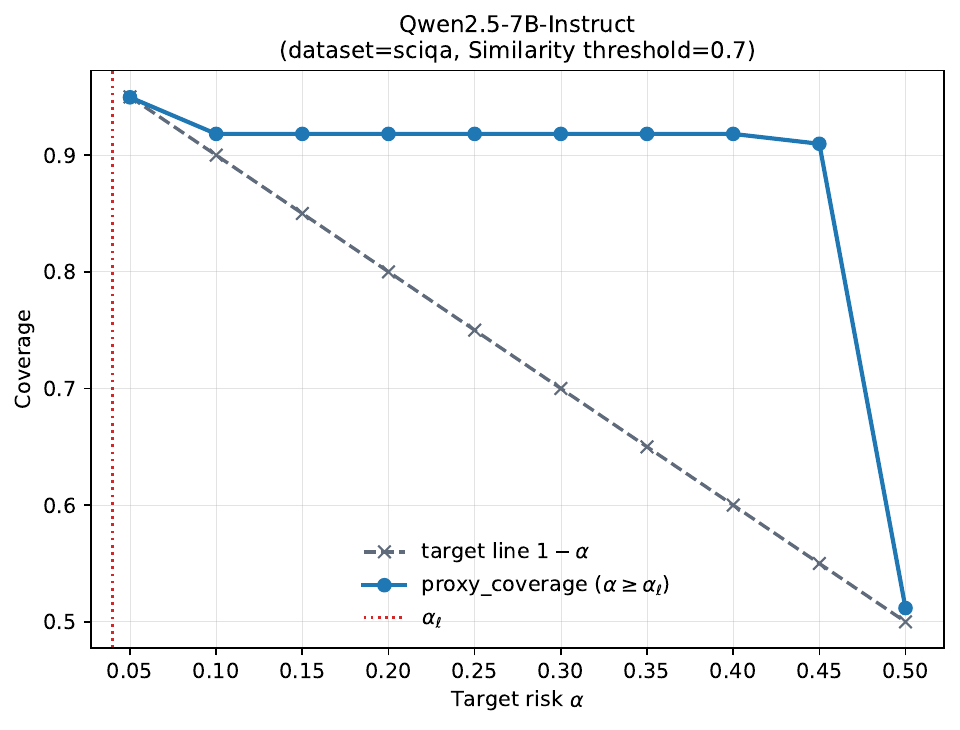}
    % \caption{Qwen2.5-7B}
  \end{subfigure}
  \hfill
  \begin{subfigure}[b]{0.195\textwidth}
    \centering
    \includegraphics[width=\textwidth]{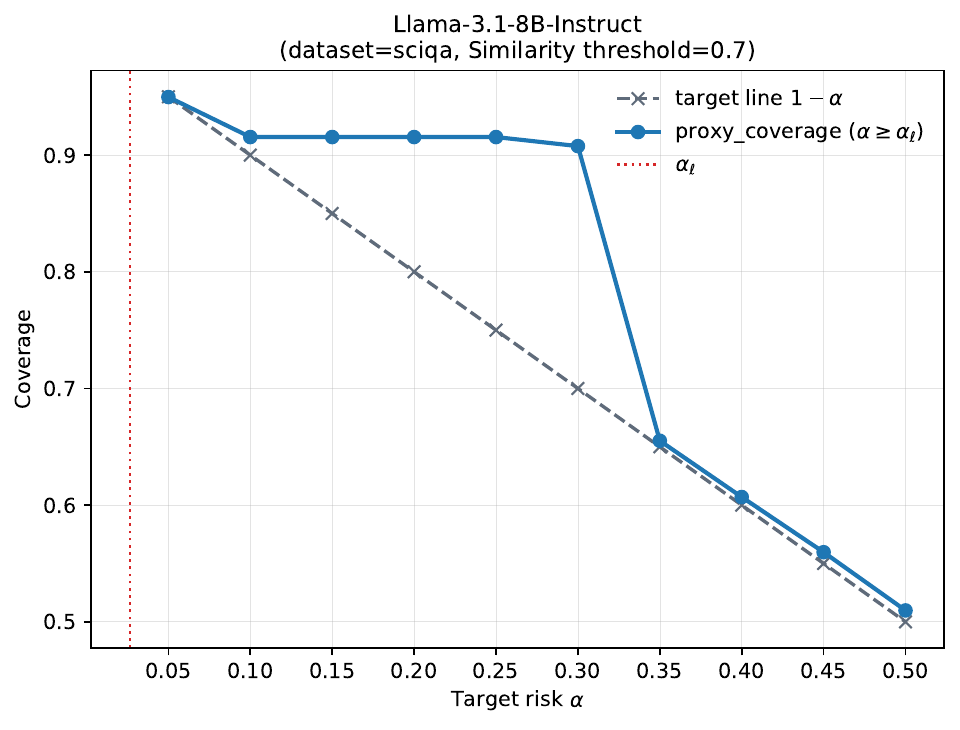}
    % \caption{LLaMA-3.1-8B}
  \end{subfigure}
  \hfill
  \begin{subfigure}[b]{0.195\textwidth}
    \centering
    \includegraphics[width=\textwidth]{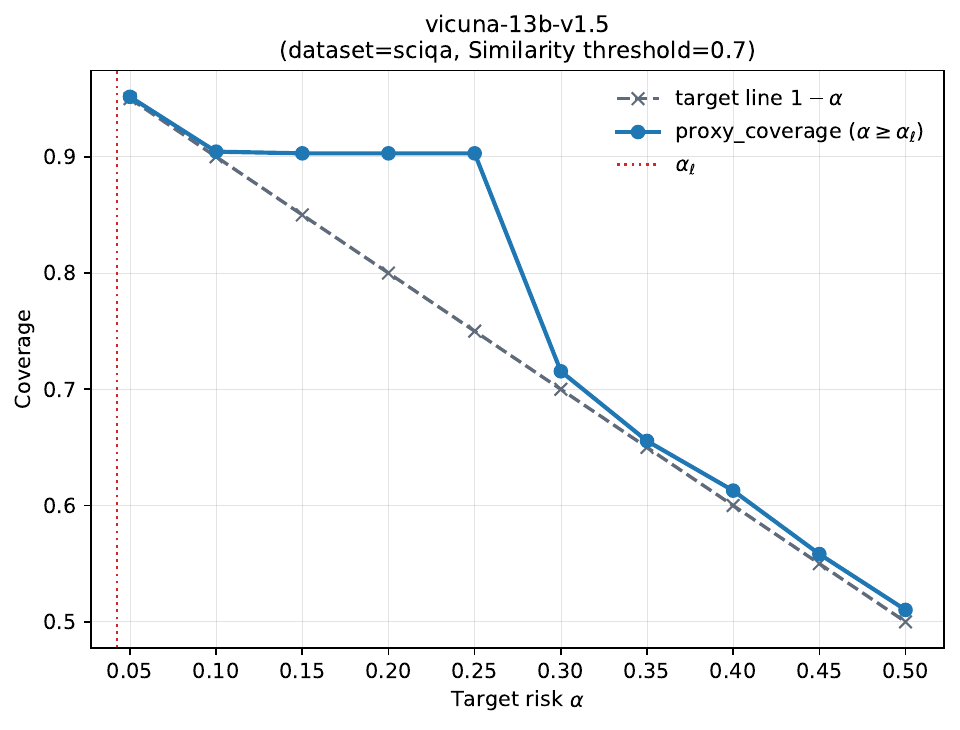}
    % \caption{Vicuna-13B}
  \end{subfigure}
  \hfill
  \begin{subfigure}[b]{0.195\textwidth}
    \centering
    \includegraphics[width=\textwidth]{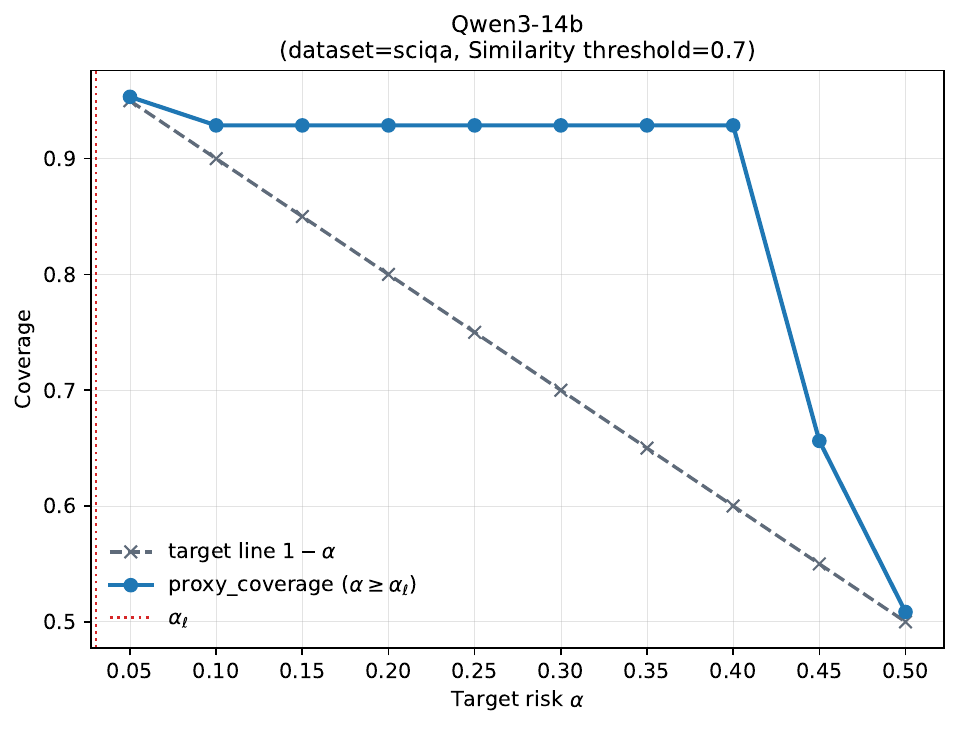}
    % \caption{Qwen3-14B}
  \end{subfigure}

  \begin{subfigure}[b]{0.195\textwidth}
    \centering
    \includegraphics[width=\textwidth]{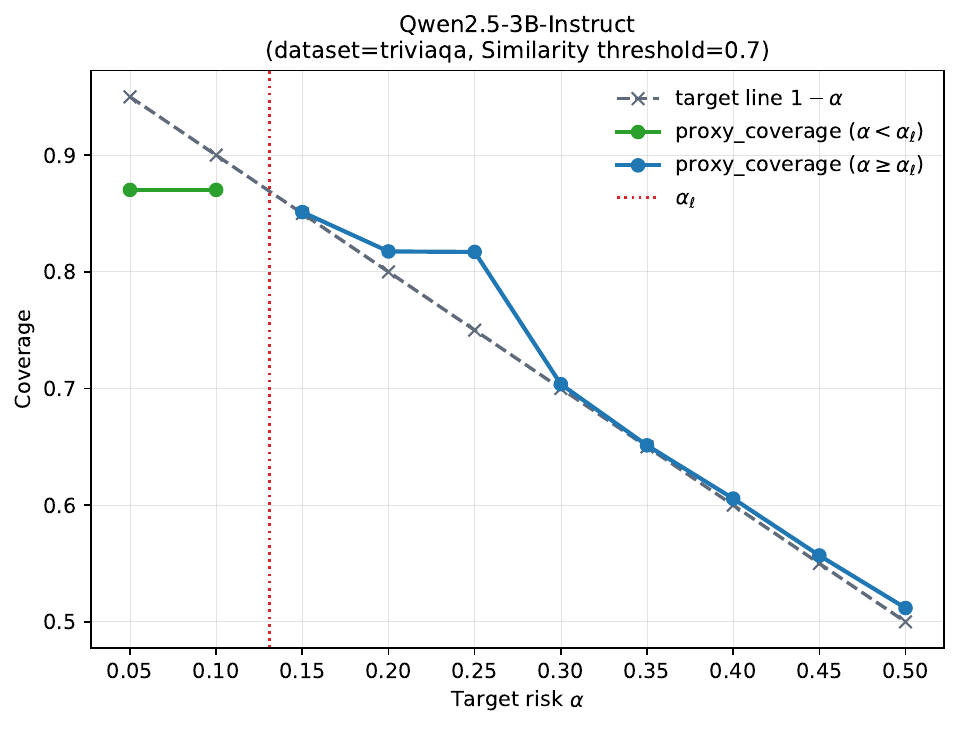}
    \caption{Qwen2.5-3B}
  \end{subfigure}
  \hfill
  \begin{subfigure}[b]{0.195\textwidth}
    \centering
    \includegraphics[width=\textwidth]{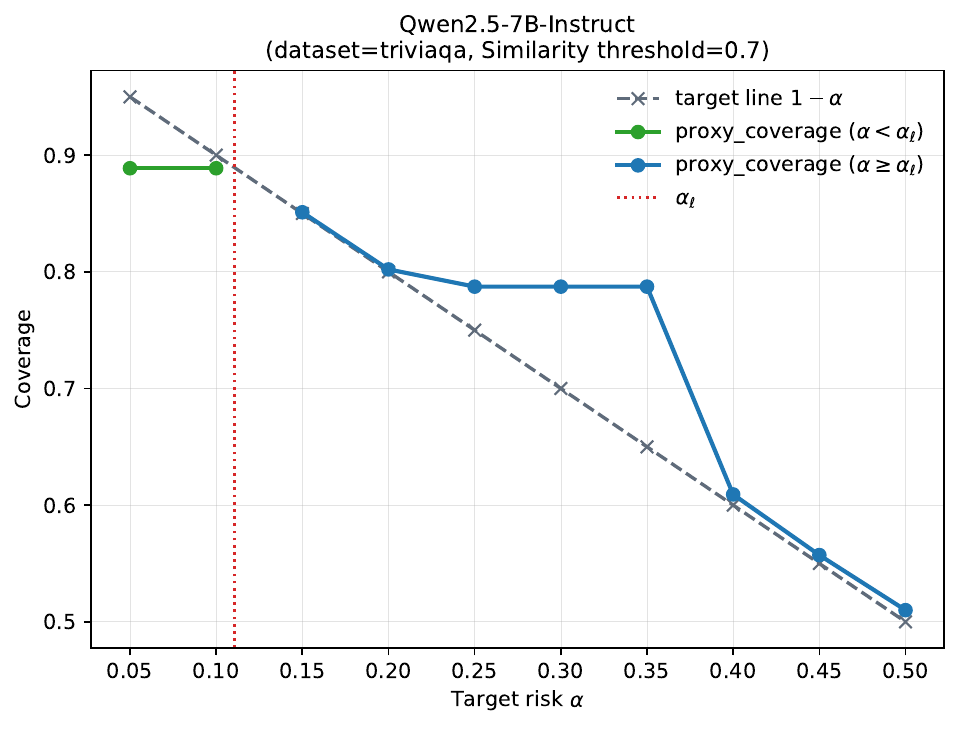}
    \caption{Qwen2.5-7B}
  \end{subfigure}
  \hfill
  \begin{subfigure}[b]{0.195\textwidth}
    \centering
    \includegraphics[width=\textwidth]{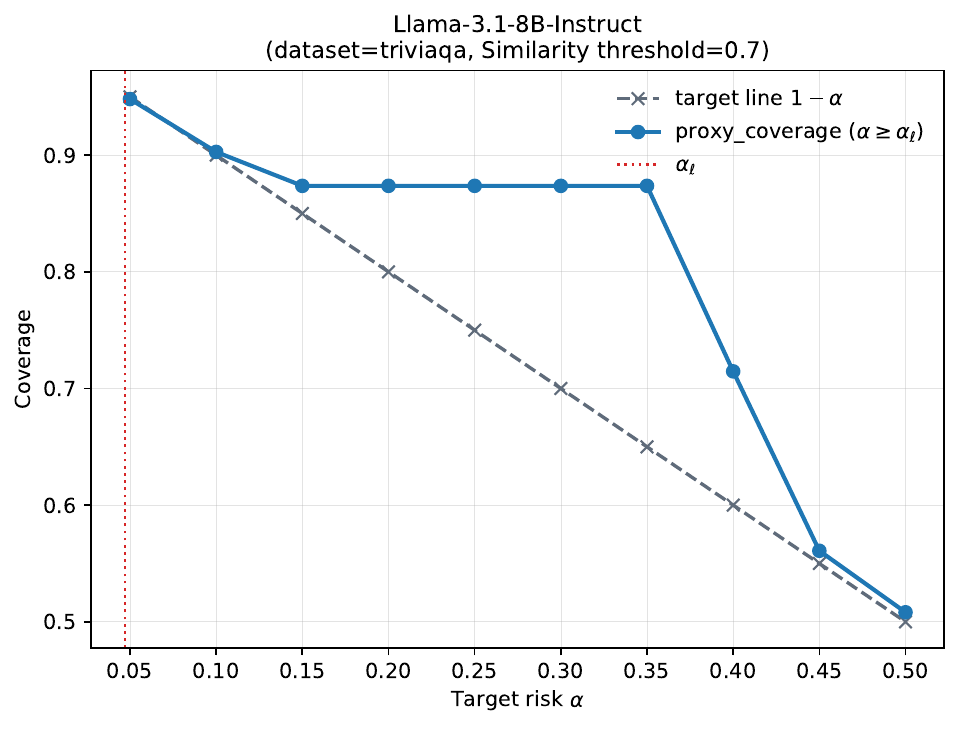}
    \caption{LLaMA-3.1-8B}
  \end{subfigure}
  \hfill
  \begin{subfigure}[b]{0.195\textwidth}
    \centering
    \includegraphics[width=\textwidth]{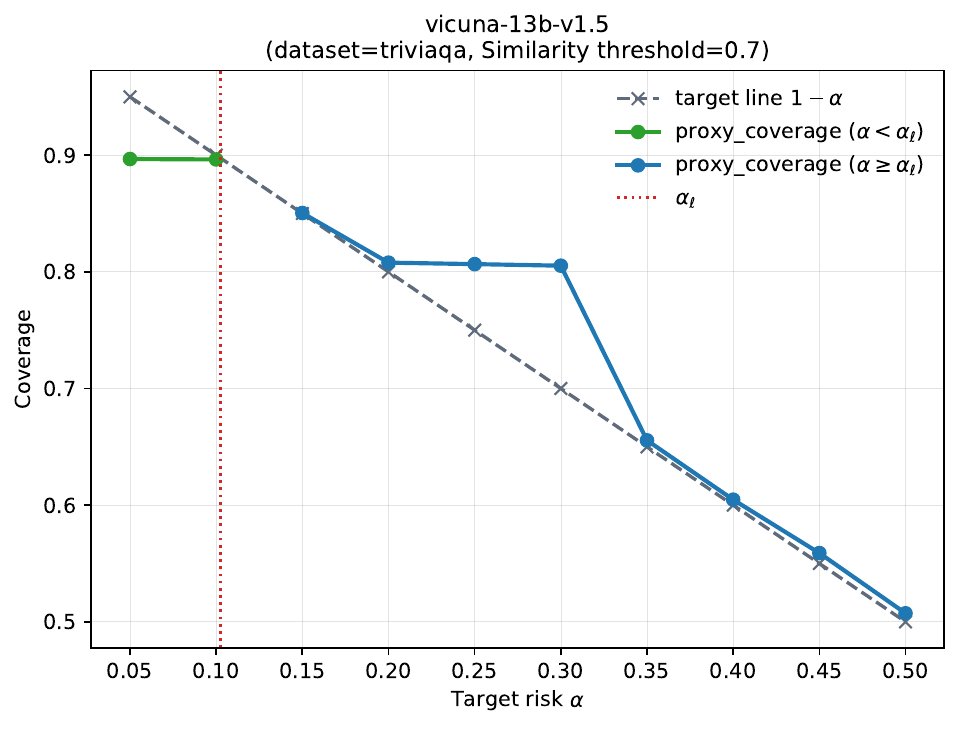}
    \caption{Vicuna-13B}
  \end{subfigure}
  \hfill
  \begin{subfigure}[b]{0.195\textwidth}
    \centering
    \includegraphics[width=\textwidth]{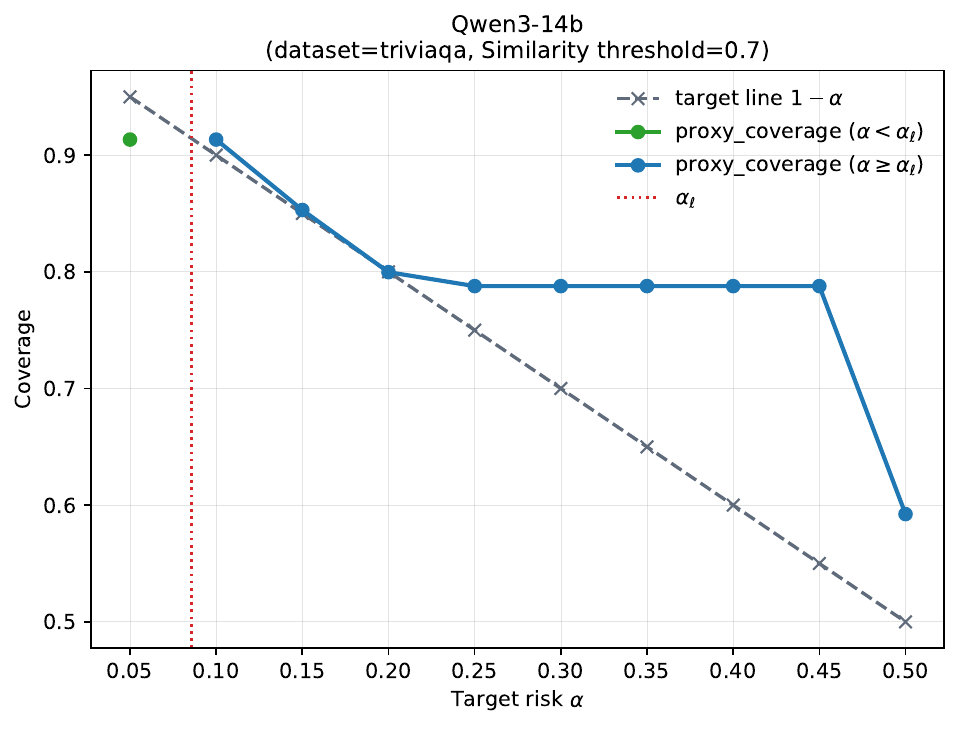}
    \caption{Qwen3-14B}
  \end{subfigure}

      \caption{Coverage Guarantees on six NLG benchmarks utilizing five LLMs. The threshold of sentence similarity is fixed at 0.7.}
  \label{fig: Coverage Guarantees (0.7)}
\end{figure*}

\paragraph{Hyperparameters.} 
Following recent research~\citep{duan-etal-2024-shifting,wang2025word,wang-etal-2024-conu}, we employ beam search to generate the MLG. 
For the sentence similarity, we utilize a cross-encoder model provided by the SentenceTransformers library~\citep{reimers2019sentence}, with RoBERTa-large~\citep{liu2019roberta} as the backbone. 
For the bidirectional entailment, we leverage DeBERTa-large-mnli as the NLI classifier~\citep{he2020deberta}. 
The global random seed for all experimental results is fixed at 10. 
The calibration-test split ratio is set to 0.5 by default. 
We repeated this random splitting process 100 times and
reported the average results~\citep{quach2024conformal}. 
Regarding generation configurations, the point
prediction baseline used \texttt{num-beams=5}. For candidate pool sampling, we set
K = 20, \texttt{temperature=1.0} and \texttt{top-p=0.9}. 
During the threshold calibration process, the search step size for the $\lambda$ was set to 0.01. 
For the admission function $A$, we estimate the sentence similarity between the candidate answer and the ground-truth by default. 

\subsection{Evaluation Results}
We organize our evaluation around four questions: (i) whether point prediction is intrinsically insufficient for open-ended generation, (ii) whether the proposed feasibility-aware framework attains valid coverage in the feasible regime, (iii) how the feasibility boundary changes with sampling and semantic criteria, and (iv) whether the resulting prediction sets are efficient and robust in practice. 

\paragraph{1) Point prediction is intrinsically limited in open-ended generation.}
We first ask a more fundamental question before evaluating calibration itself: is point prediction an adequate abstraction for open-ended generation? To answer this, Figure~\ref{fig: rq1} compares the semantic accuracy of MLG with the attainability upper bound $1-\alpha_l$ of the sampled candidate pool under different semantic matching thresholds. Here, $1-\alpha_l$ represents the maximum achievable coverage if one simply returns the entire sampled candidate set, and therefore quantifies the recoverable headroom already present in the generation space.

\begin{figure*}[!t]
  \centering
  \begin{subfigure}[b]{0.195\textwidth}
    \centering
    \includegraphics[width=\textwidth]{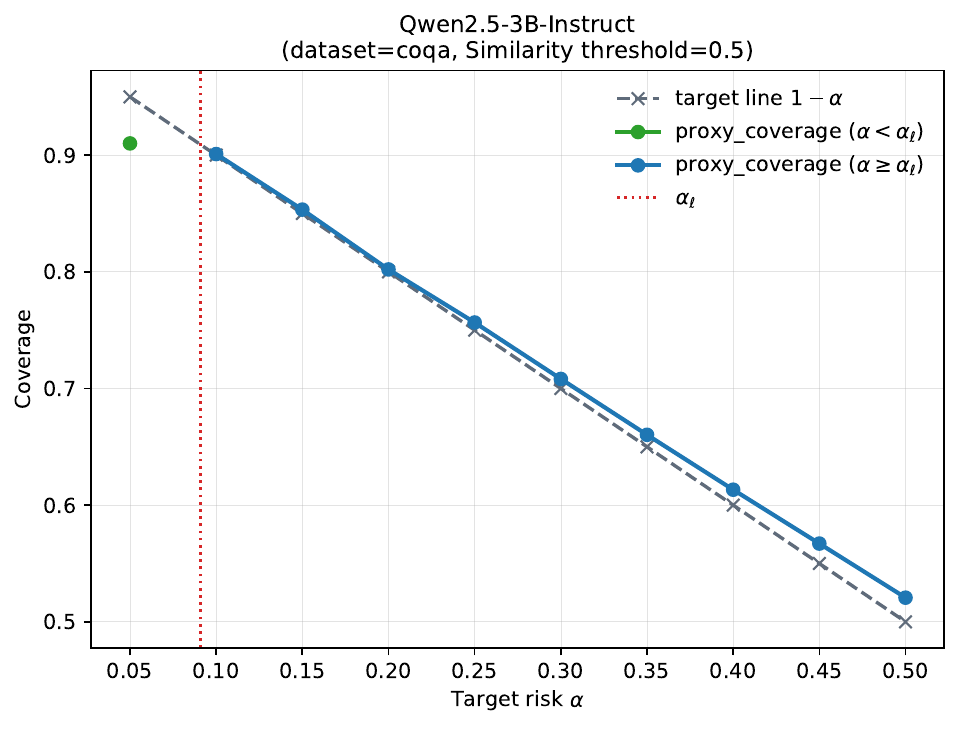}
    % \caption{Qwen2.5-3B}
  \end{subfigure}
  \hfill
  \begin{subfigure}[b]{0.195\textwidth}
    \centering
    \includegraphics[width=\textwidth]{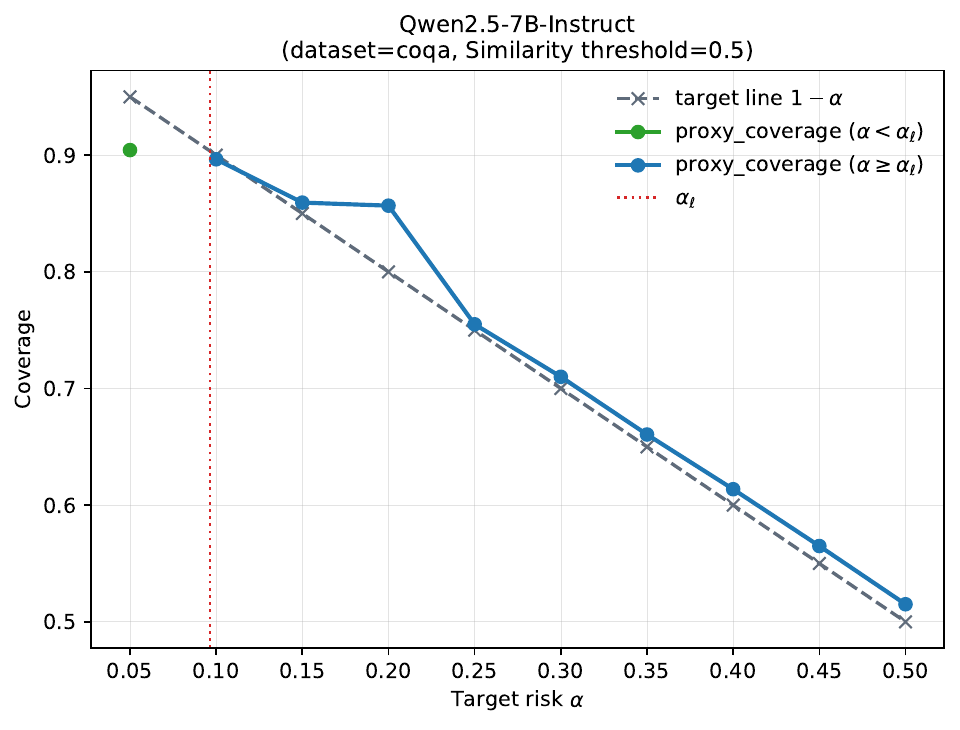}
    % \caption{Qwen2.5-7B}
  \end{subfigure}
  \hfill
  \begin{subfigure}[b]{0.195\textwidth}
    \centering
    \includegraphics[width=\textwidth]{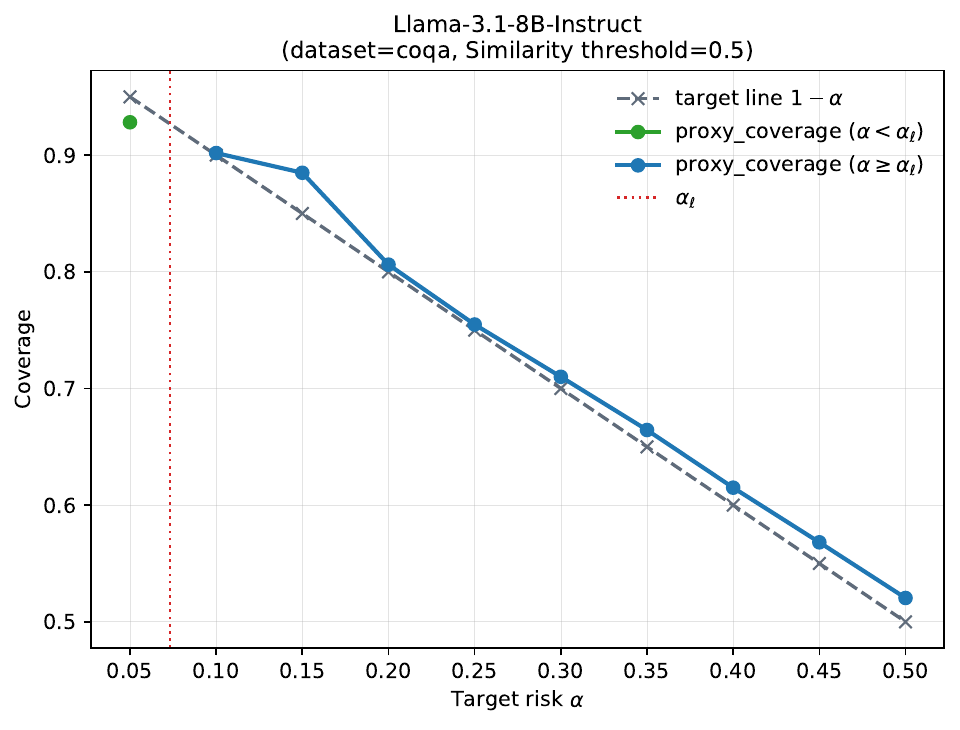}
    % \caption{LLaMA-3.1-8B}
  \end{subfigure}
  \hfill
  \begin{subfigure}[b]{0.195\textwidth}
    \centering
    \includegraphics[width=\textwidth]{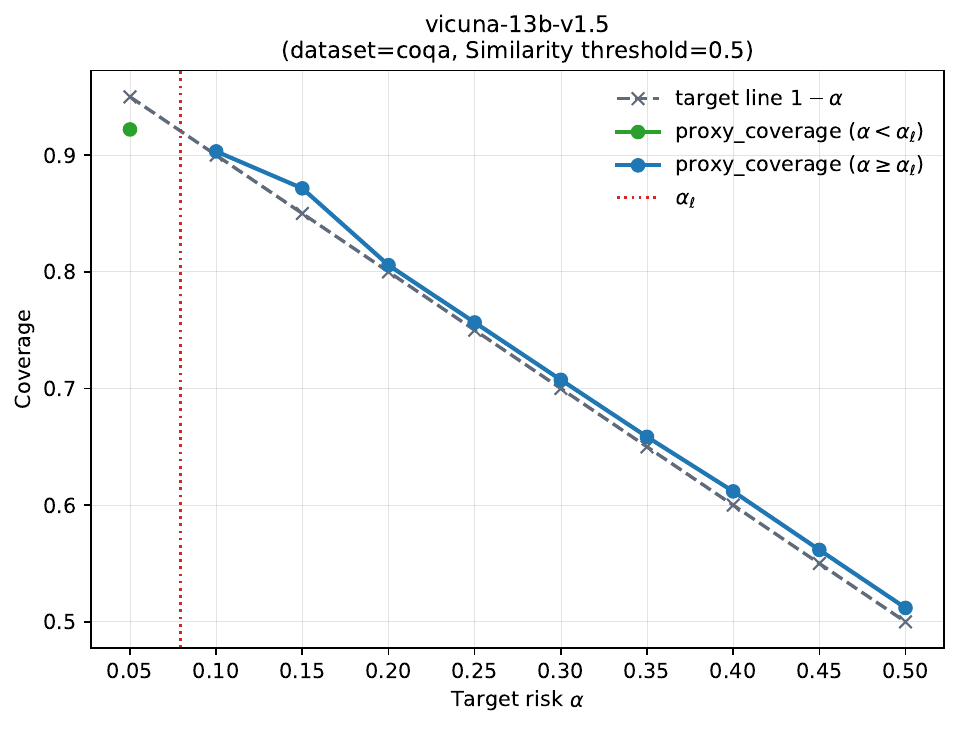}
    % \caption{Vicuna-13B}
  \end{subfigure}
  \hfill
  \begin{subfigure}[b]{0.195\textwidth}
    \centering
    \includegraphics[width=\textwidth]{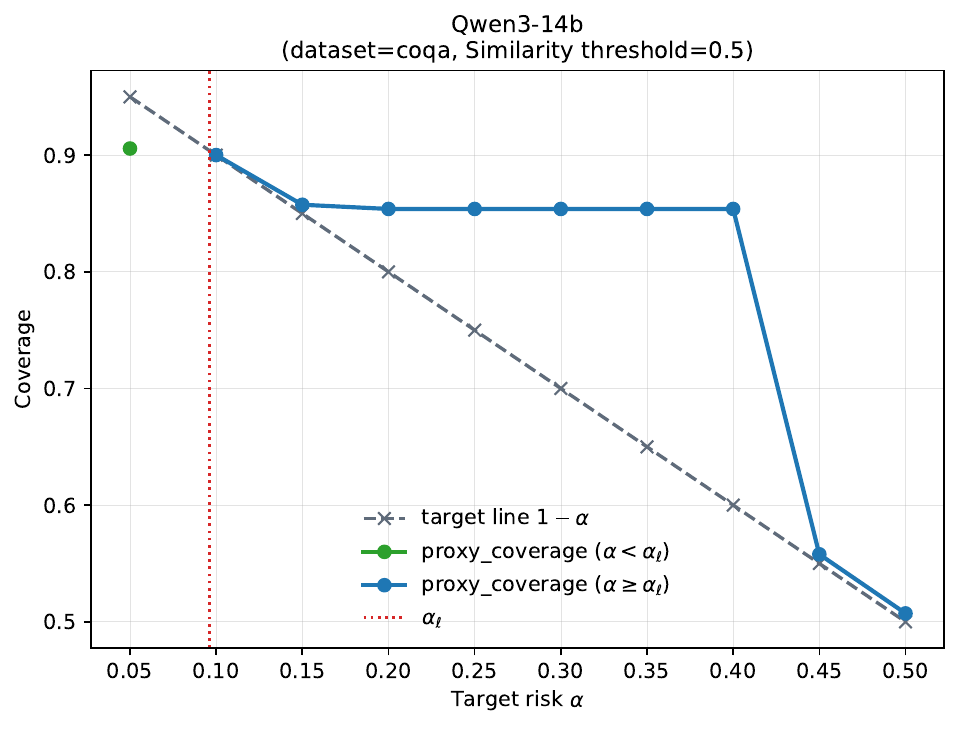}
    % \caption{Qwen3-14B}
  \end{subfigure}

  \begin{subfigure}[b]{0.195\textwidth}
    \centering
    \includegraphics[width=\textwidth]{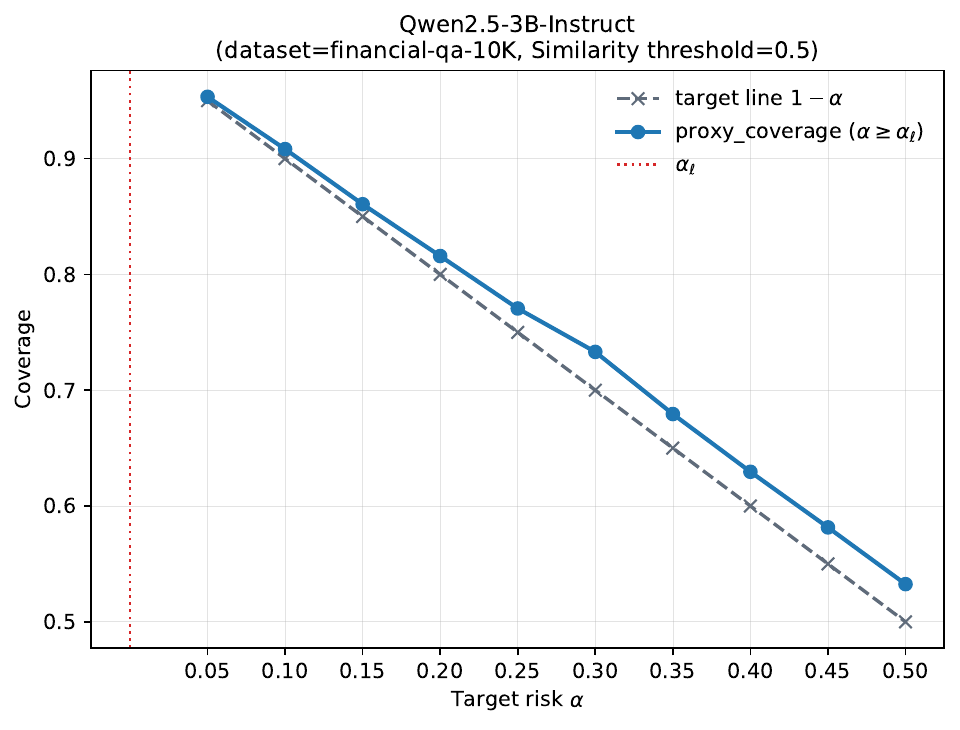}
    % \caption{Qwen2.5-3B}
  \end{subfigure}
  \hfill
  \begin{subfigure}[b]{0.195\textwidth}
    \centering
    \includegraphics[width=\textwidth]{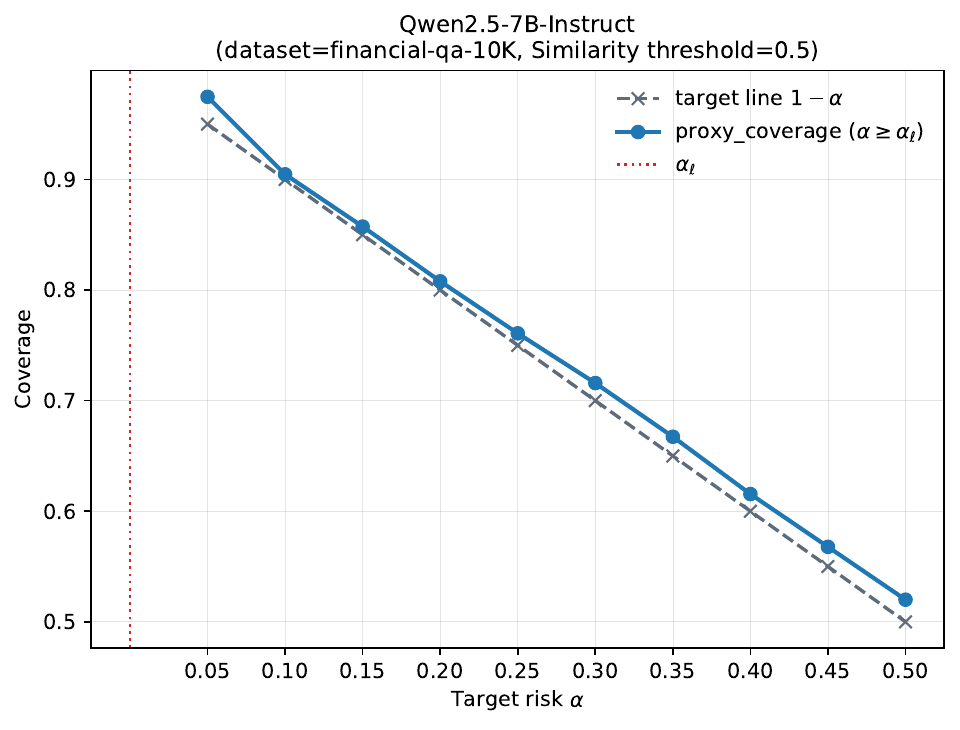}
    % \caption{Qwen2.5-7B}
  \end{subfigure}
  \hfill
  \begin{subfigure}[b]{0.195\textwidth}
    \centering
    \includegraphics[width=\textwidth]{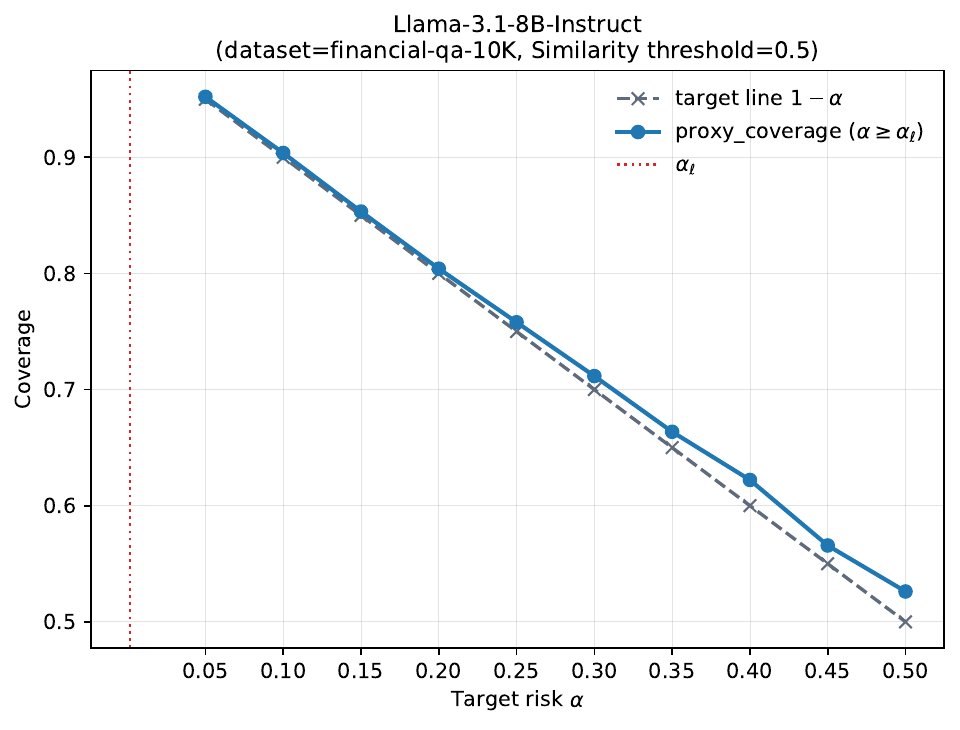}
    % \caption{LLaMA-3.1-8B}
  \end{subfigure}
  \hfill
  \begin{subfigure}[b]{0.195\textwidth}
    \centering
    \includegraphics[width=\textwidth]{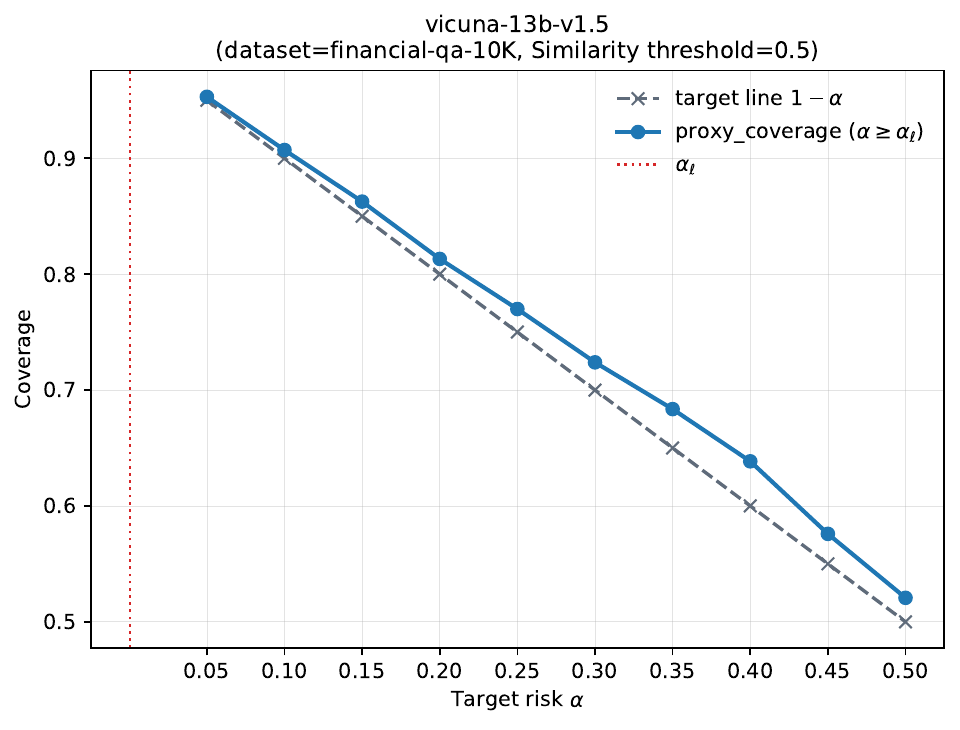}
    % \caption{Vicuna-13B}
  \end{subfigure}
  \hfill
  \begin{subfigure}[b]{0.195\textwidth}
    \centering
    \includegraphics[width=\textwidth]{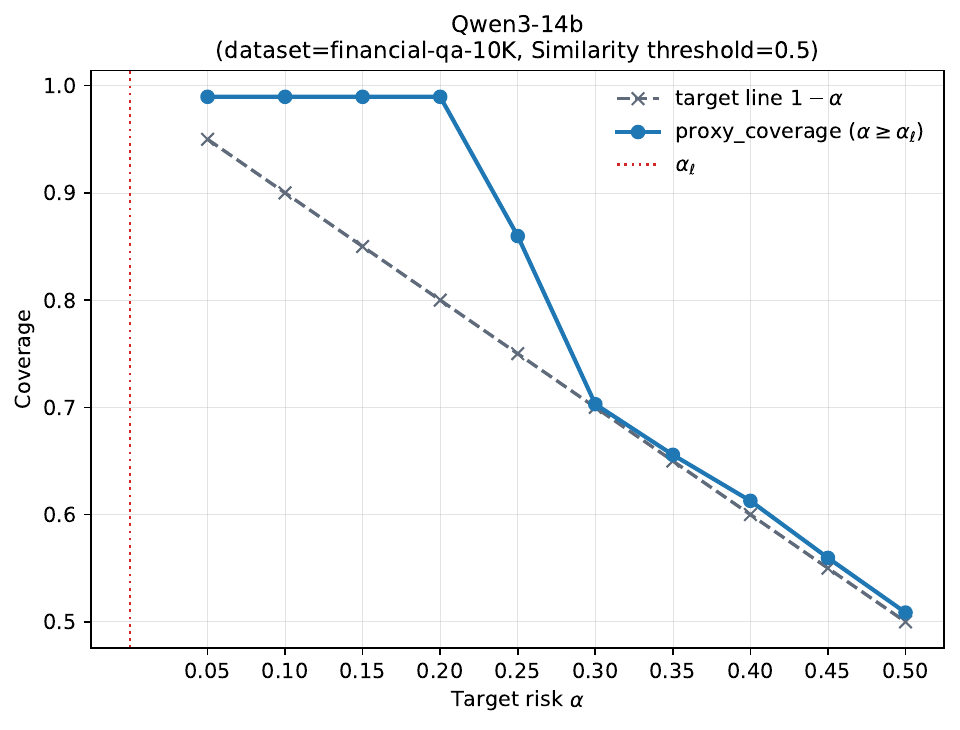}
    % \caption{Qwen3-14B}
  \end{subfigure}

  \begin{subfigure}[b]{0.195\textwidth}
    \centering
    \includegraphics[width=\textwidth]{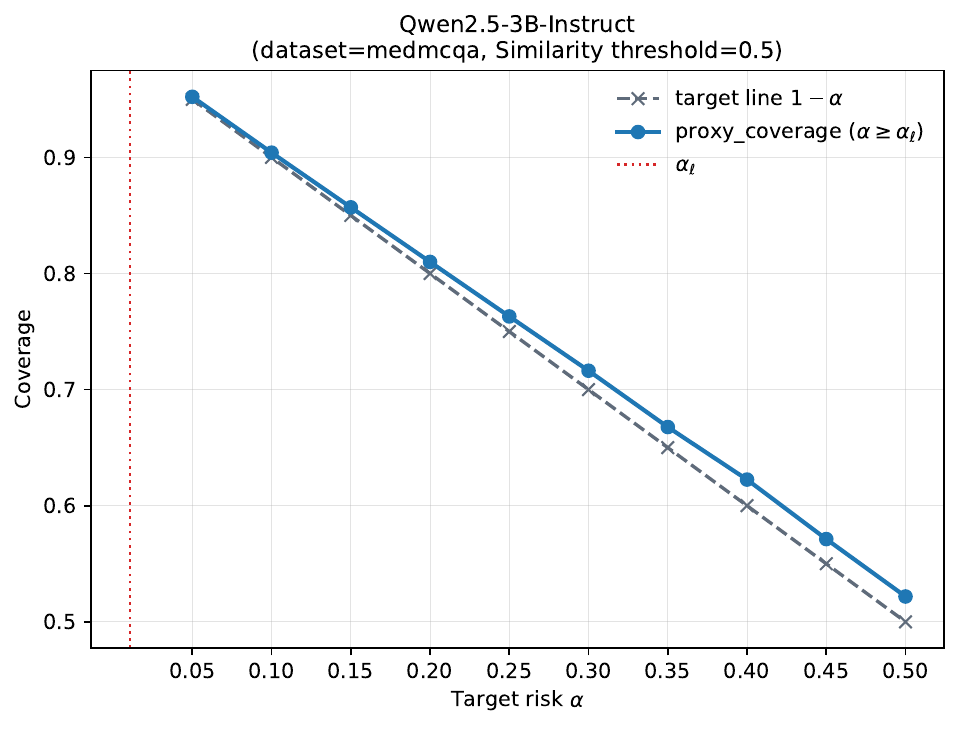}
    % \caption{Qwen2.5-3B}
  \end{subfigure}
  \hfill
  \begin{subfigure}[b]{0.195\textwidth}
    \centering
    \includegraphics[width=\textwidth]{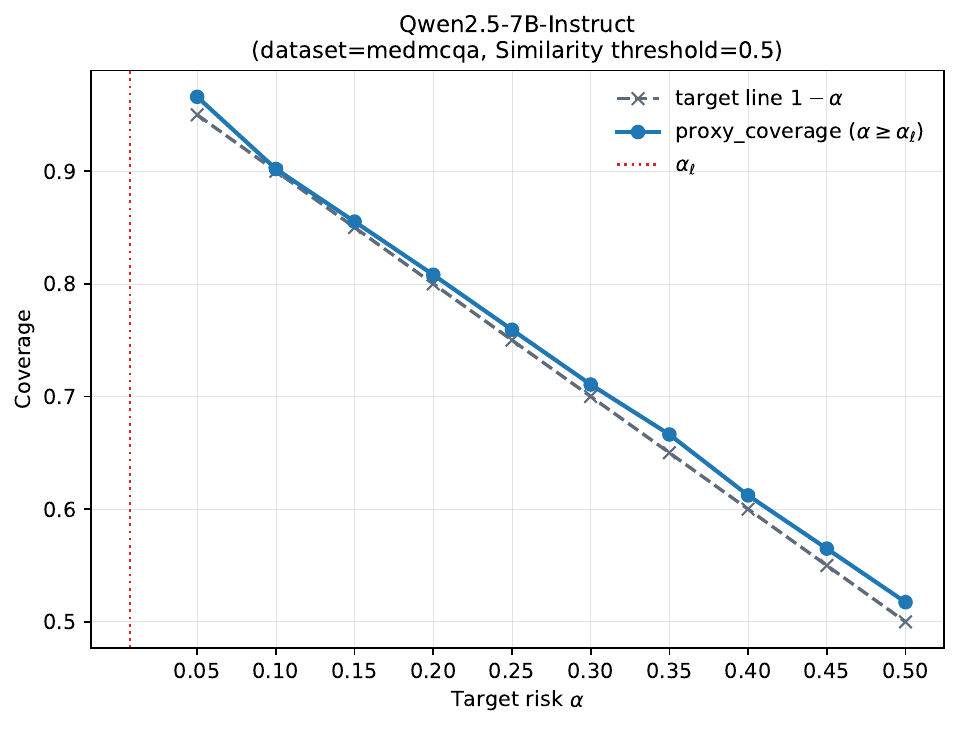}
    % \caption{Qwen2.5-7B}
  \end{subfigure}
  \hfill
  \begin{subfigure}[b]{0.195\textwidth}
    \centering
    \includegraphics[width=\textwidth]{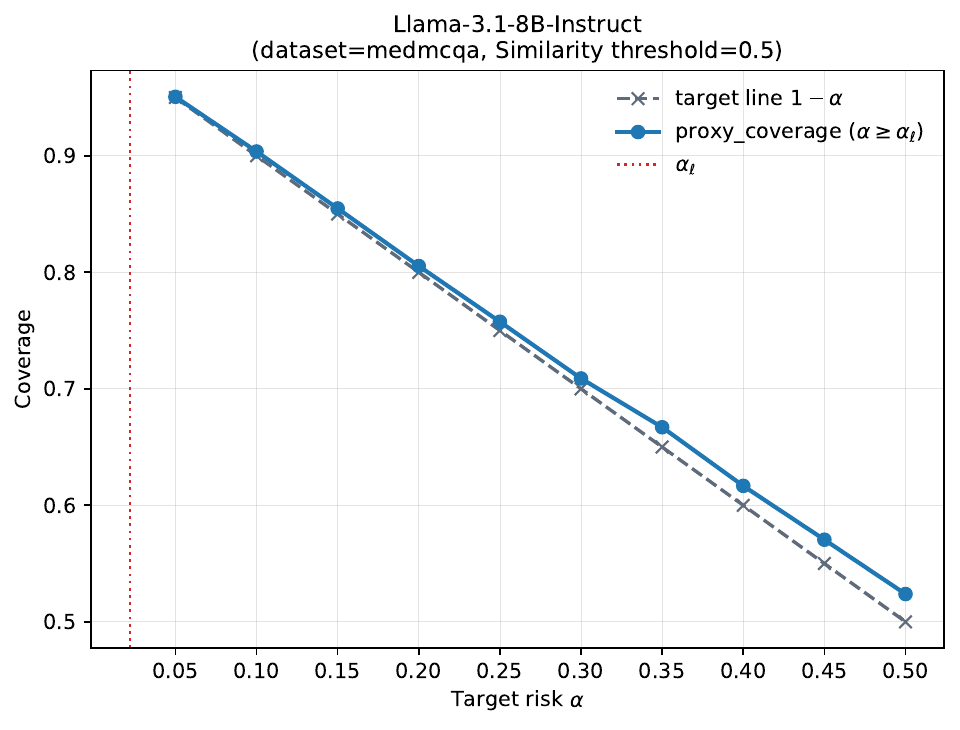}
    % \caption{LLaMA-3.1-8B}
  \end{subfigure}
  \hfill
  \begin{subfigure}[b]{0.195\textwidth}
    \centering
    \includegraphics[width=\textwidth]{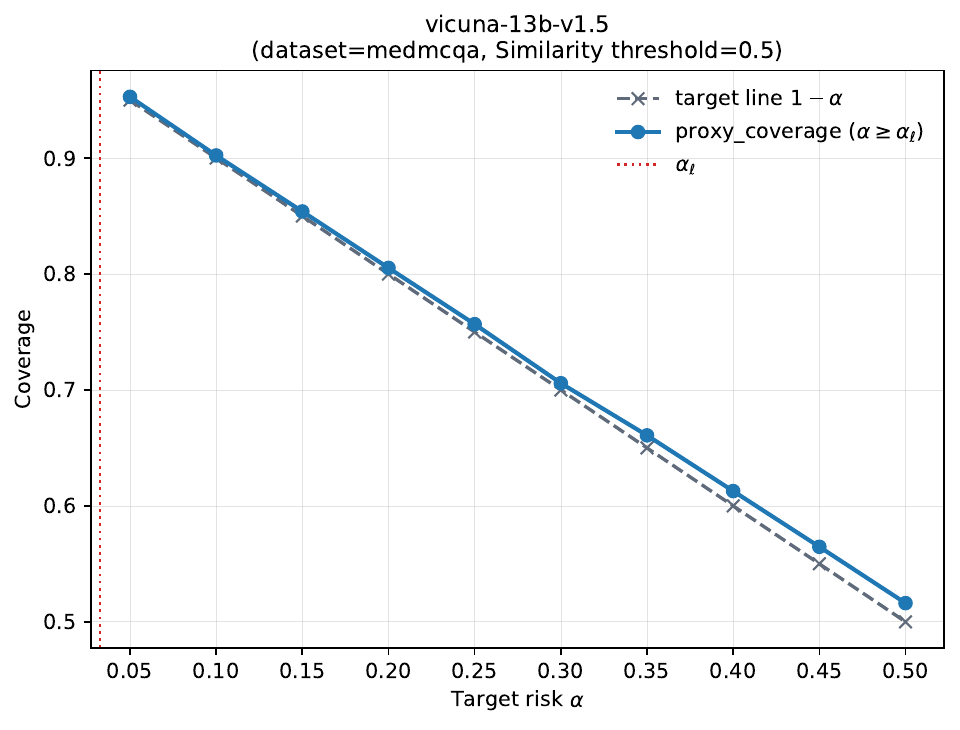}
    % \caption{Vicuna-13B}
  \end{subfigure}
  \hfill
  \begin{subfigure}[b]{0.195\textwidth}
    \centering
    \includegraphics[width=\textwidth]{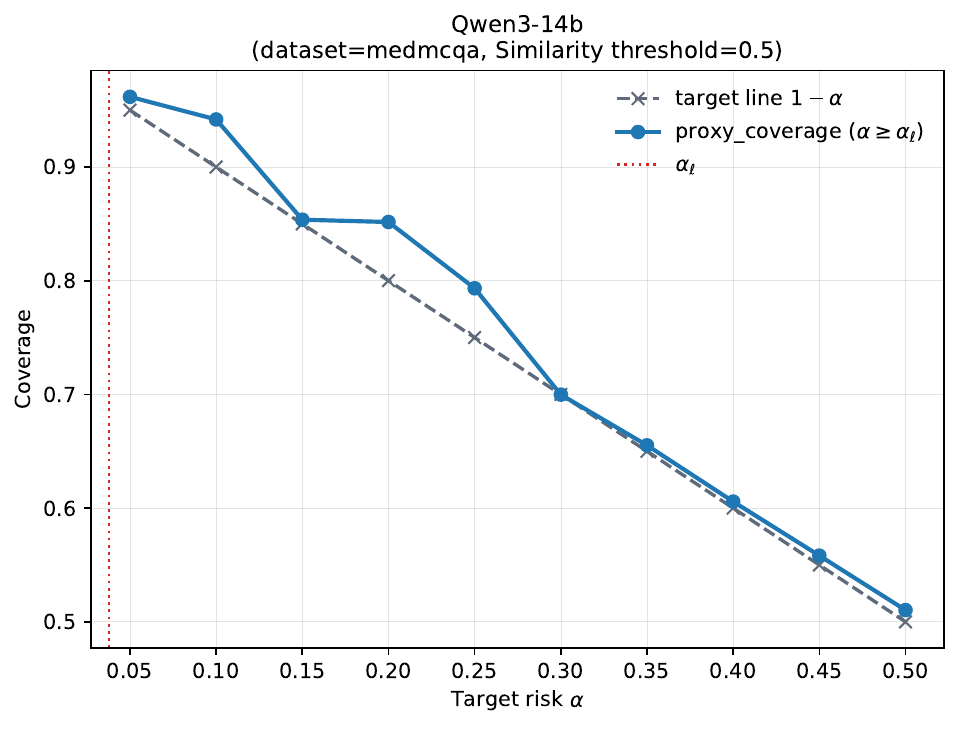}
    % \caption{Qwen3-14B}
  \end{subfigure}

  \begin{subfigure}[b]{0.195\textwidth}
    \centering
    \includegraphics[width=\textwidth]{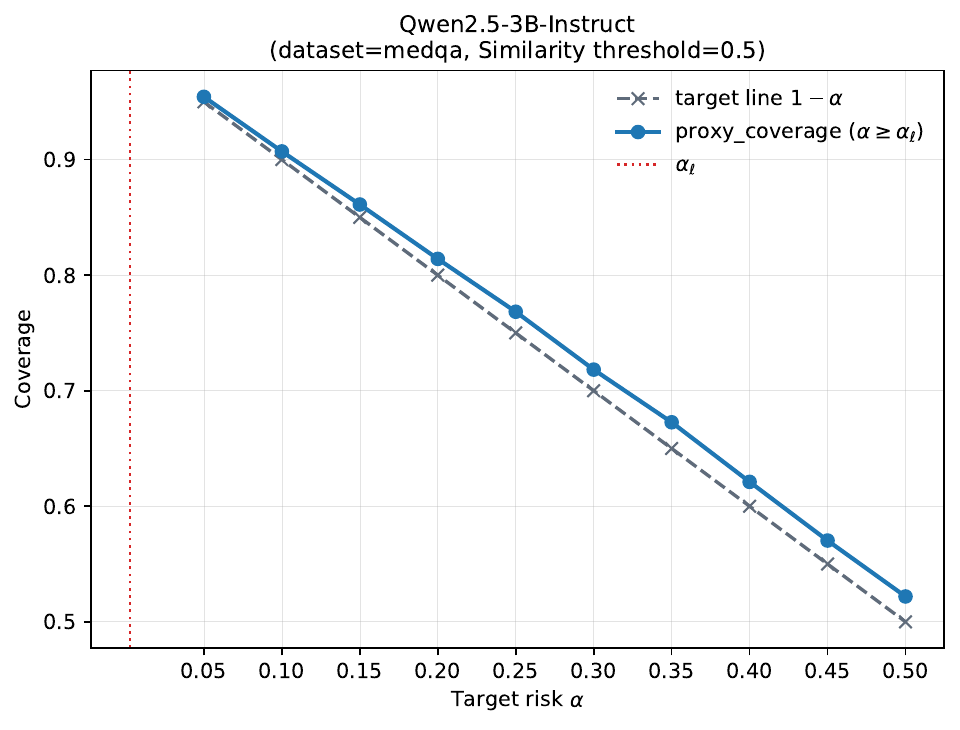}
    % \caption{Qwen2.5-3B}
  \end{subfigure}
  \hfill
  \begin{subfigure}[b]{0.195\textwidth}
    \centering
    \includegraphics[width=\textwidth]{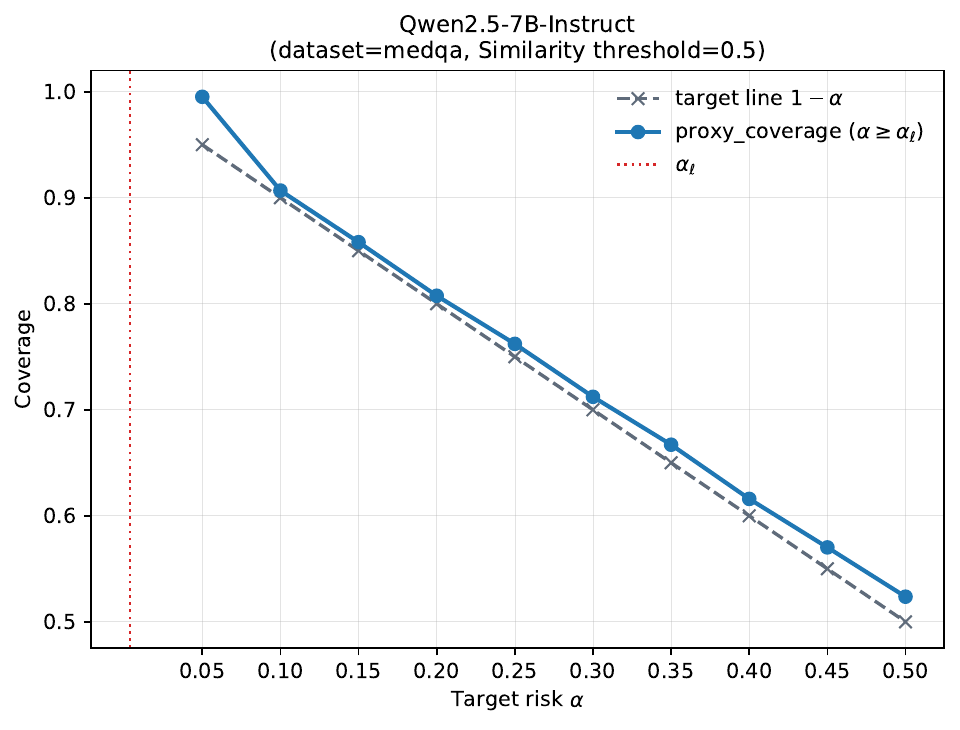}
    % \caption{Qwen2.5-7B}
  \end{subfigure}
  \hfill
  \begin{subfigure}[b]{0.195\textwidth}
    \centering
    \includegraphics[width=\textwidth]{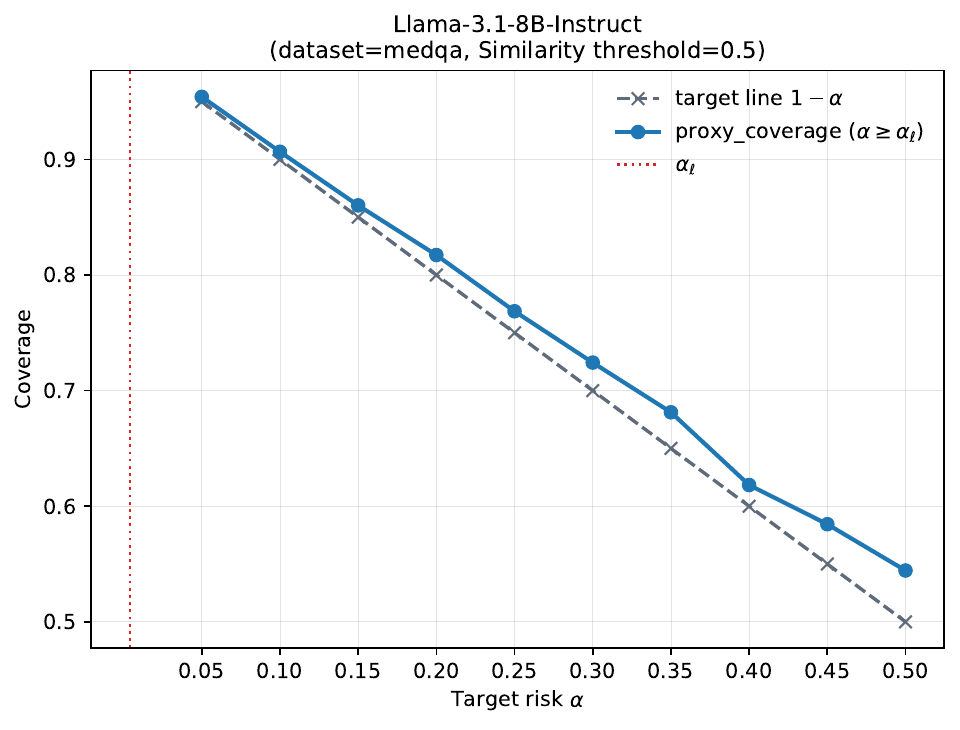}
    % \caption{LLaMA-3.1-8B}
  \end{subfigure}
  \hfill
  \begin{subfigure}[b]{0.195\textwidth}
    \centering
    \includegraphics[width=\textwidth]{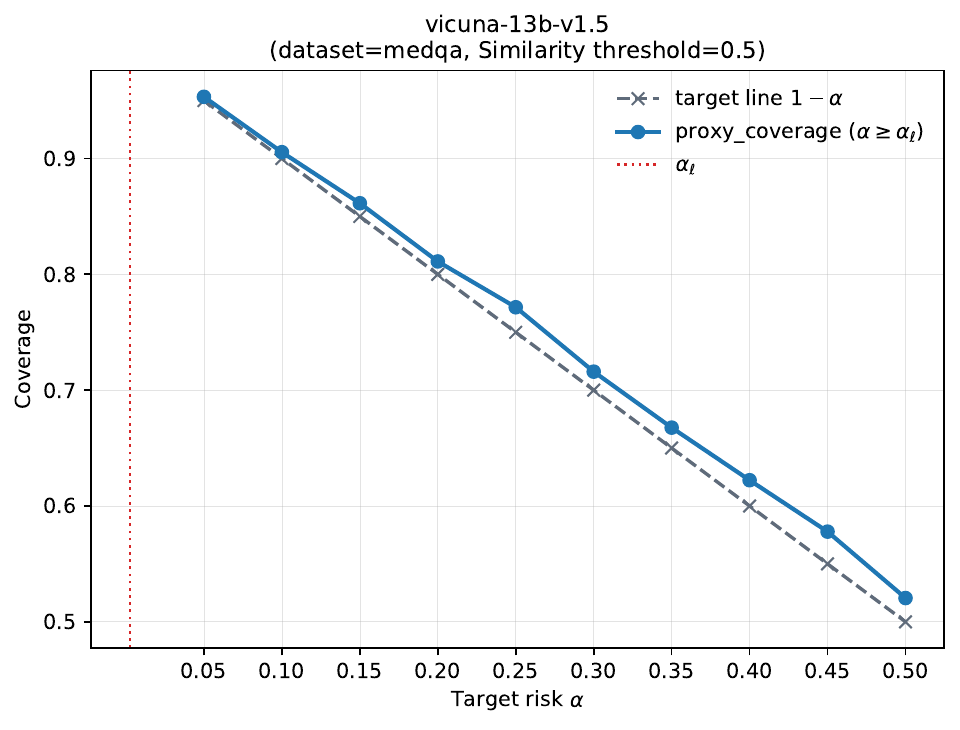}
    % \caption{Vicuna-13B}
  \end{subfigure}
  \hfill
  \begin{subfigure}[b]{0.195\textwidth}
    \centering
    \includegraphics[width=\textwidth]{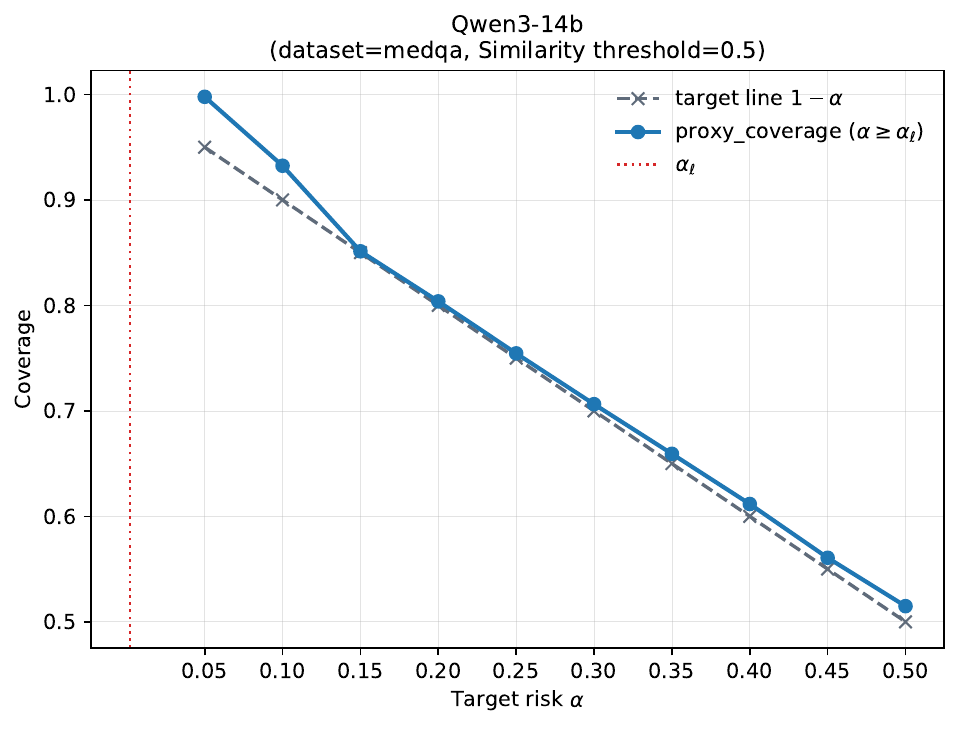}
    % \caption{Qwen3-14B}
  \end{subfigure}

  \begin{subfigure}[b]{0.195\textwidth}
    \centering
    \includegraphics[width=\textwidth]{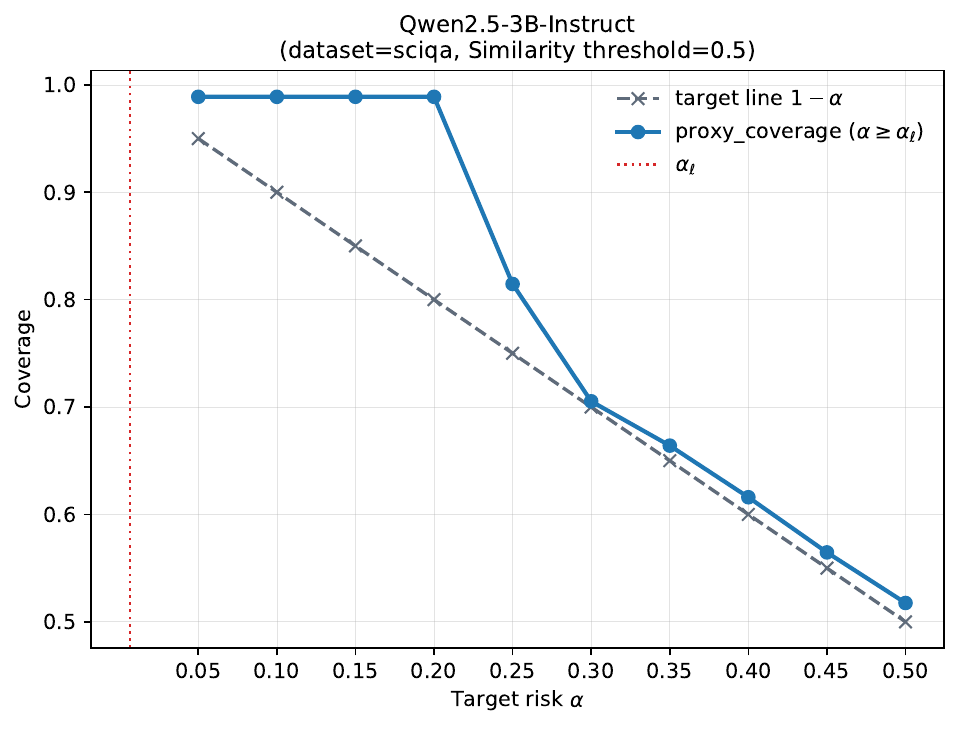}
    % \caption{Qwen2.5-3B}
  \end{subfigure}
  \hfill
  \begin{subfigure}[b]{0.195\textwidth}
    \centering
    \includegraphics[width=\textwidth]{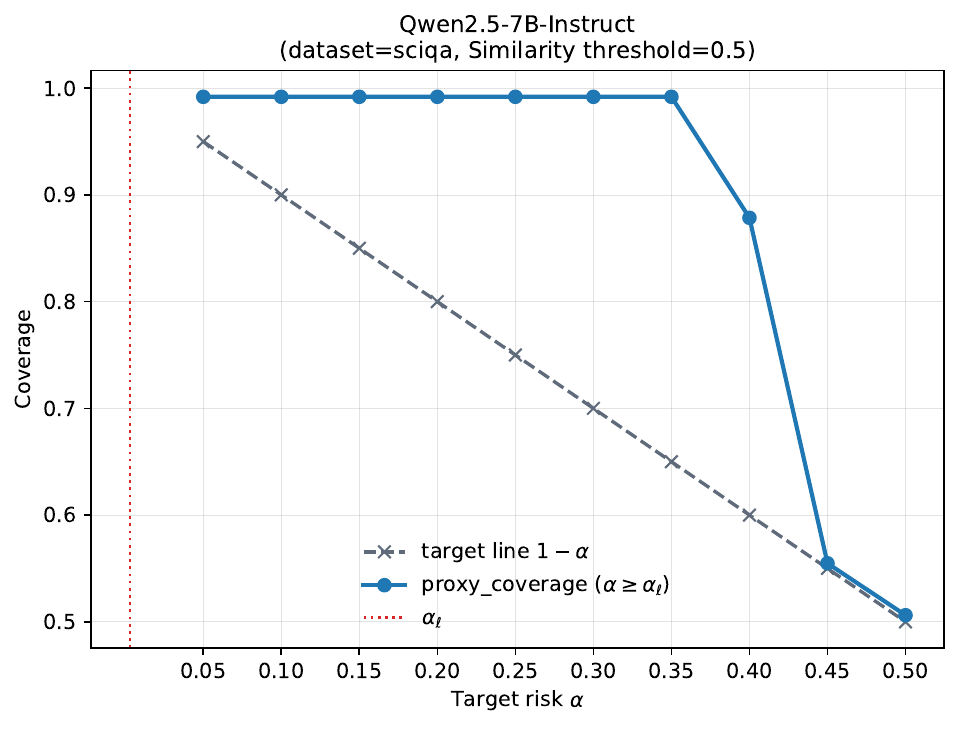}
    % \caption{Qwen2.5-7B}
  \end{subfigure}
  \hfill
  \begin{subfigure}[b]{0.195\textwidth}
    \centering
    \includegraphics[width=\textwidth]{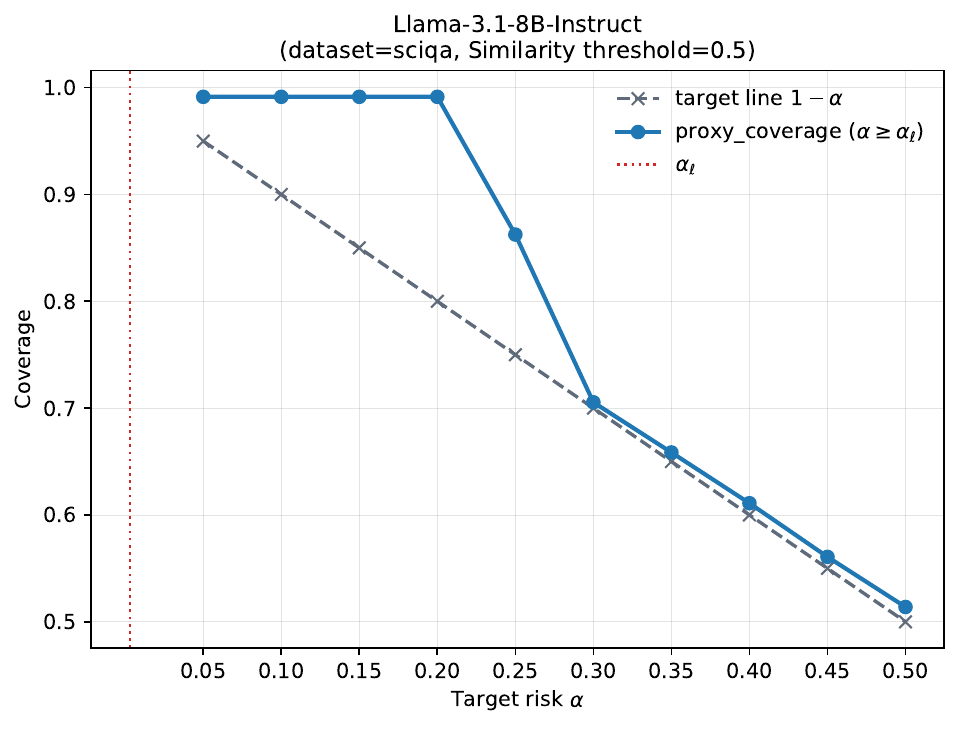}
    % \caption{LLaMA-3.1-8B}
  \end{subfigure}
  \hfill
  \begin{subfigure}[b]{0.195\textwidth}
    \centering
    \includegraphics[width=\textwidth]{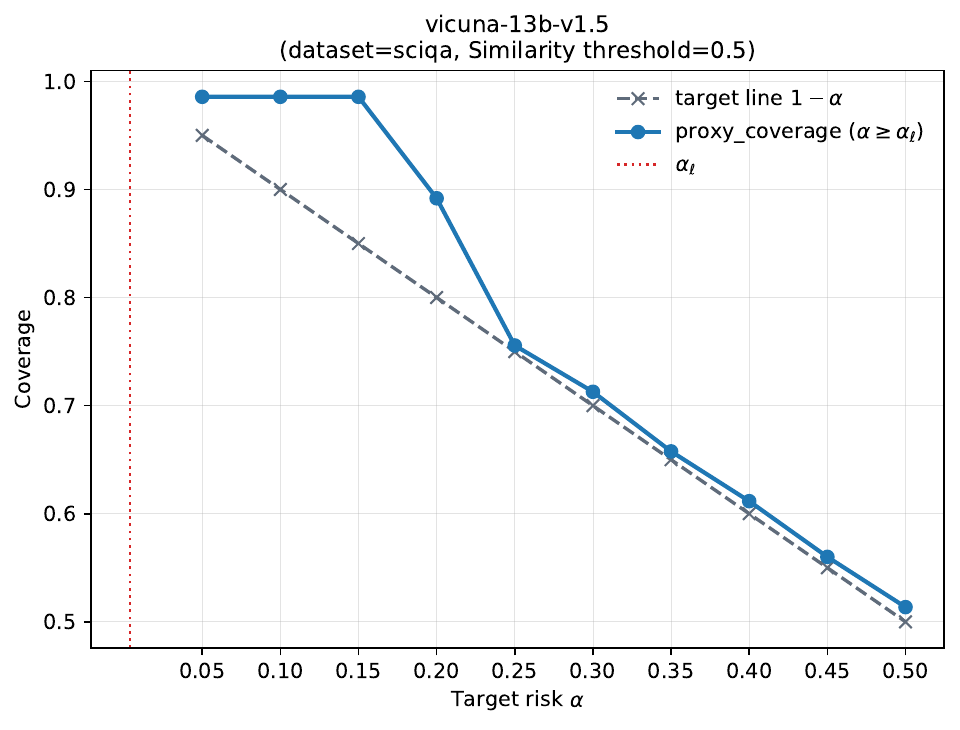}
    % \caption{Vicuna-13B}
  \end{subfigure}
  \hfill
  \begin{subfigure}[b]{0.195\textwidth}
    \centering
    \includegraphics[width=\textwidth]{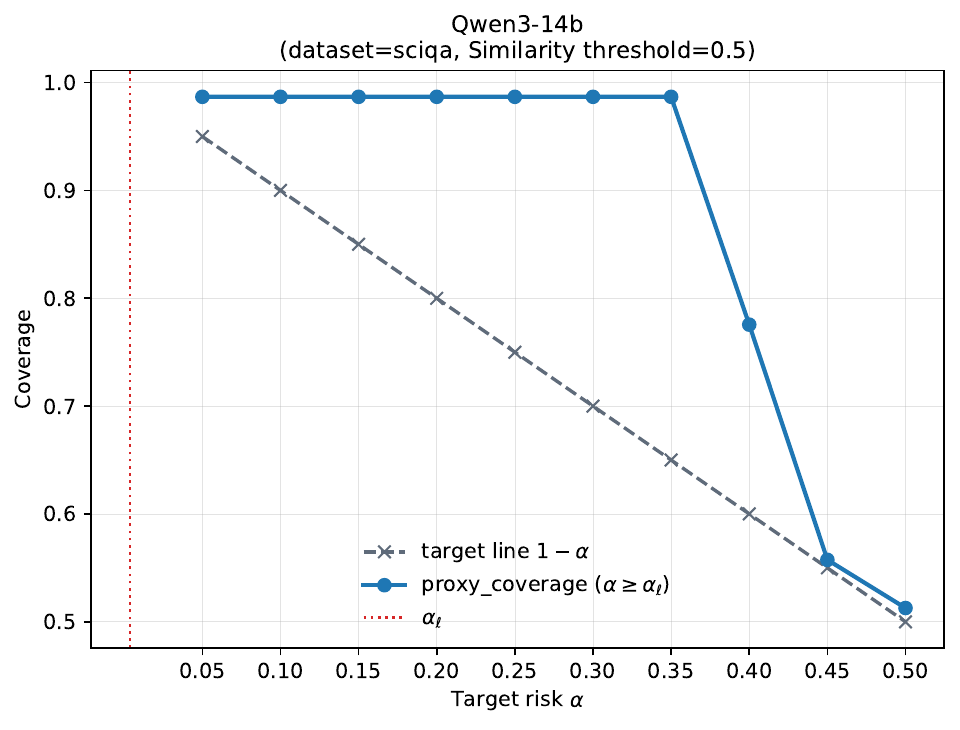}
    % \caption{Qwen3-14B}
  \end{subfigure}

  \begin{subfigure}[b]{0.195\textwidth}
    \centering
    \includegraphics[width=\textwidth]{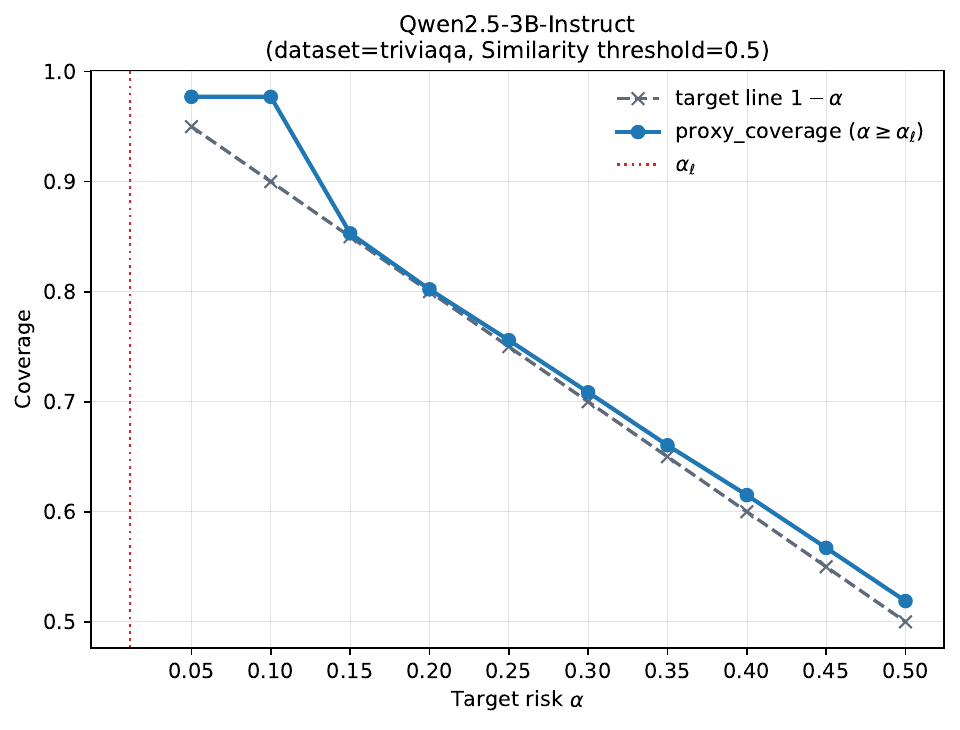}
    \caption{Qwen2.5-3B}
  \end{subfigure}
  \hfill
  \begin{subfigure}[b]{0.195\textwidth}
    \centering
    \includegraphics[width=\textwidth]{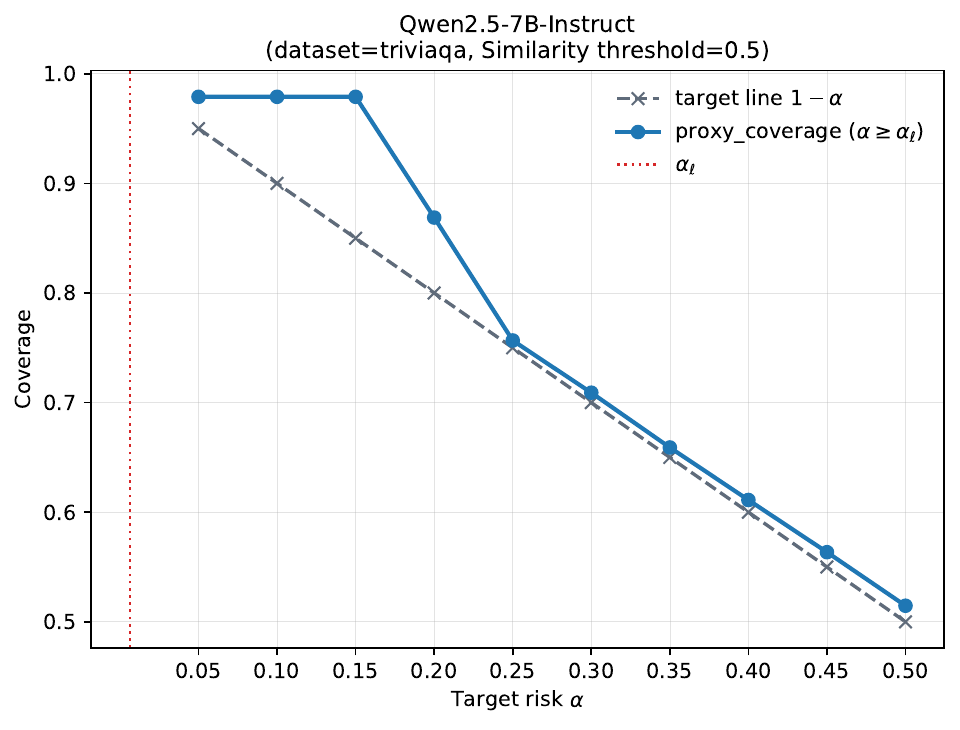}
    \caption{Qwen2.5-7B}
  \end{subfigure}
  \hfill
  \begin{subfigure}[b]{0.195\textwidth}
    \centering
    \includegraphics[width=\textwidth]{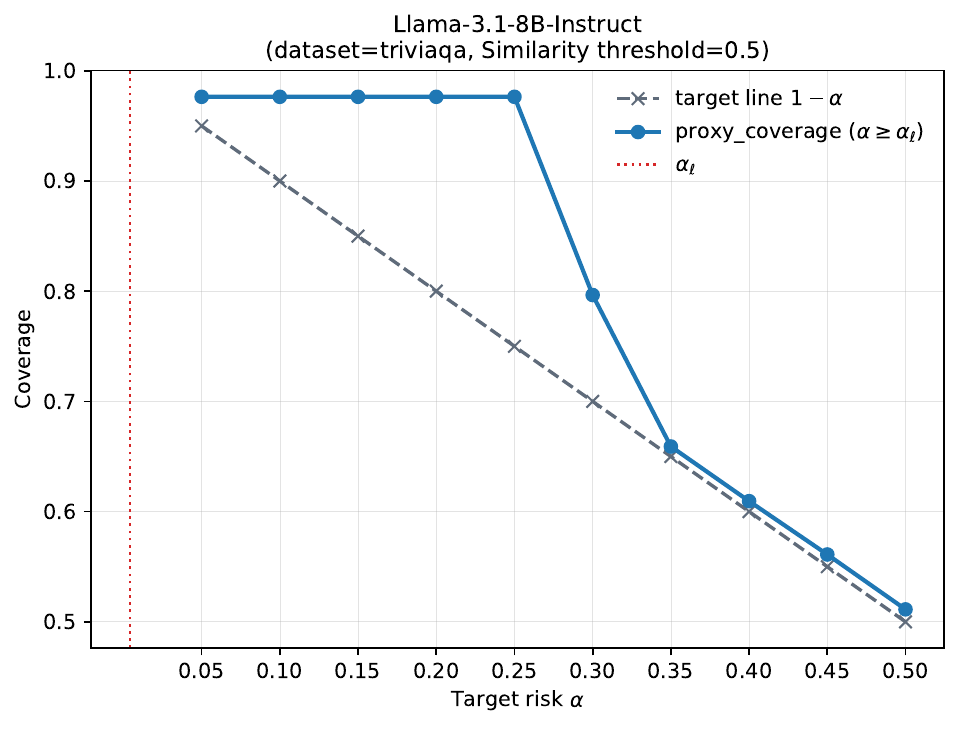}
    \caption{LLaMA-3.1-8B}
  \end{subfigure}
  \hfill
  \begin{subfigure}[b]{0.195\textwidth}
    \centering
    \includegraphics[width=\textwidth]{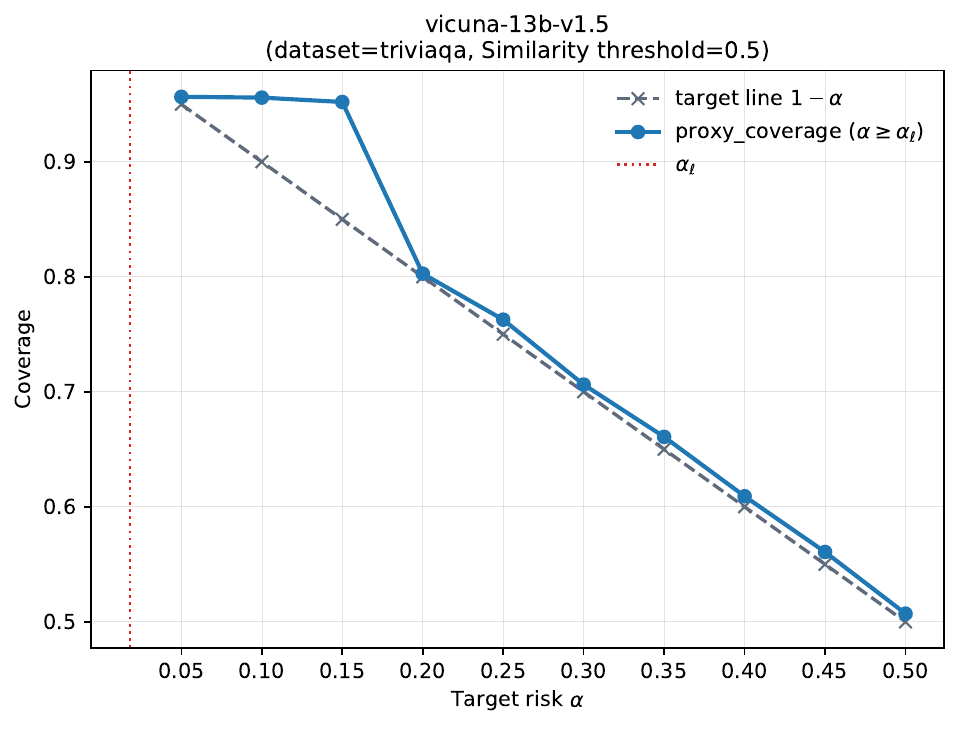}
    \caption{Vicuna-13B}
  \end{subfigure}
  \hfill
  \begin{subfigure}[b]{0.195\textwidth}
    \centering
    \includegraphics[width=\textwidth]{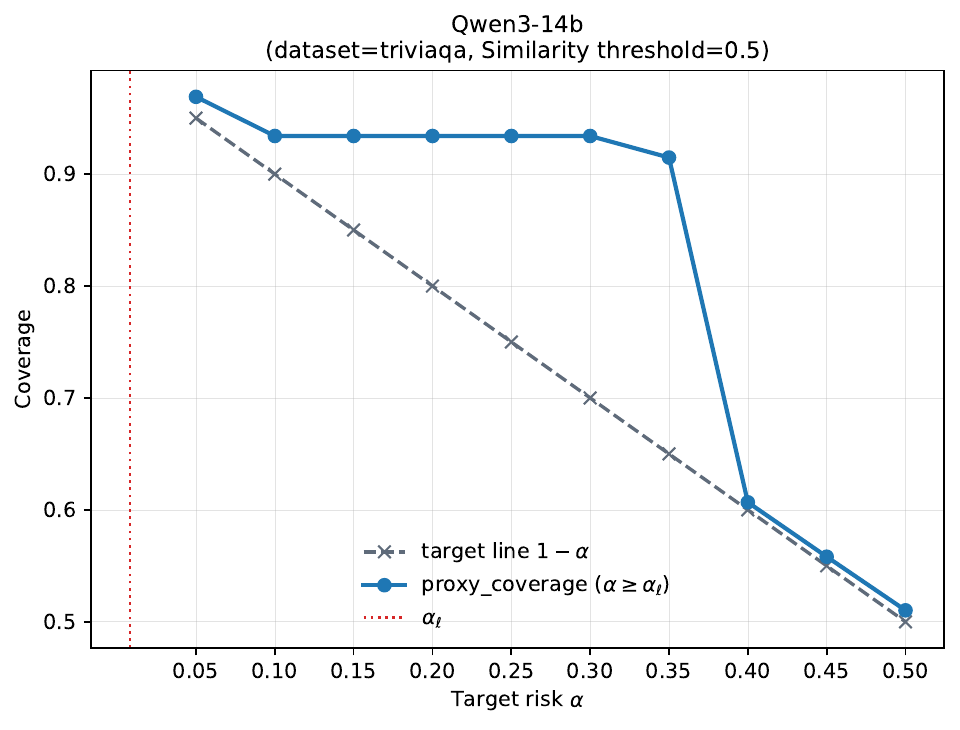}
    \caption{Qwen3-14B}
  \end{subfigure}

      \caption{Coverage Guarantees on six NLG benchmarks utilizing five LLMs. The threshold of sentence similarity is fixed at 0.5.}
  \label{fig: Coverage Guarantees (0.5)}
\end{figure*}

A consistent pattern emerges across datasets and models: the attainability upper bound is generally higher than the accuracy of MLG, indicating that admissible answers often already exist in the sampled candidate pool even when the top-ranked response is incorrect. This shows that the weakness of point prediction is not merely that the model ``does not know'' the answer, but rather that a single committed response under-utilizes the semantic diversity already available in the sampled space. 
Moreover, the gap widens as the semantic criterion becomes stricter, suggesting that point prediction degrades more rapidly than the candidate-space attainability under demanding semantic evaluation. These results provide direct empirical motivation for moving from point prediction to set-valued prediction.

\begin{figure*}[!t]
  \centering
  \begin{subfigure}[b]{0.195\textwidth}
    \centering
    \includegraphics[width=\textwidth]{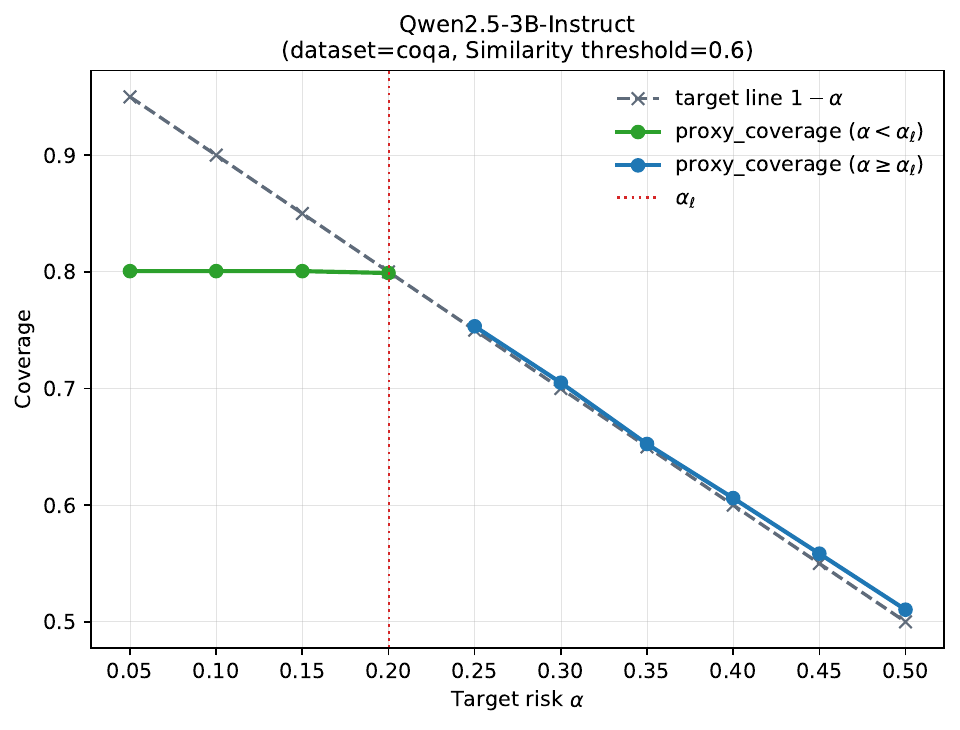}
    % \caption{Qwen2.5-3B}
  \end{subfigure}
  \hfill
  \begin{subfigure}[b]{0.195\textwidth}
    \centering
    \includegraphics[width=\textwidth]{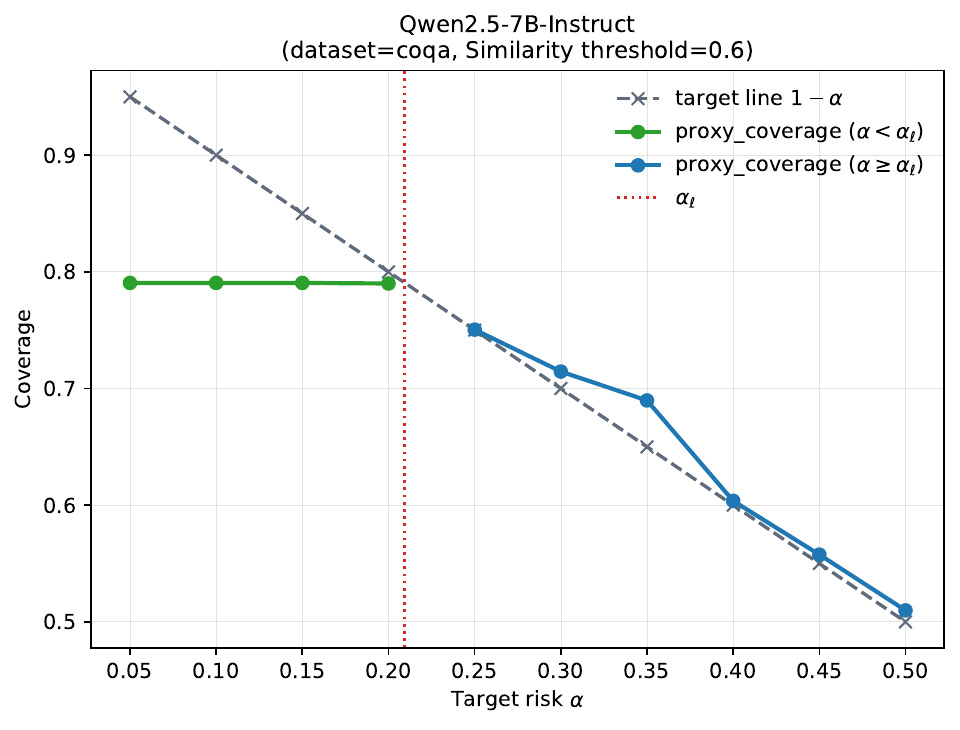}
    % \caption{Qwen2.5-7B}
  \end{subfigure}
  \hfill
  \begin{subfigure}[b]{0.195\textwidth}
    \centering
    \includegraphics[width=\textwidth]{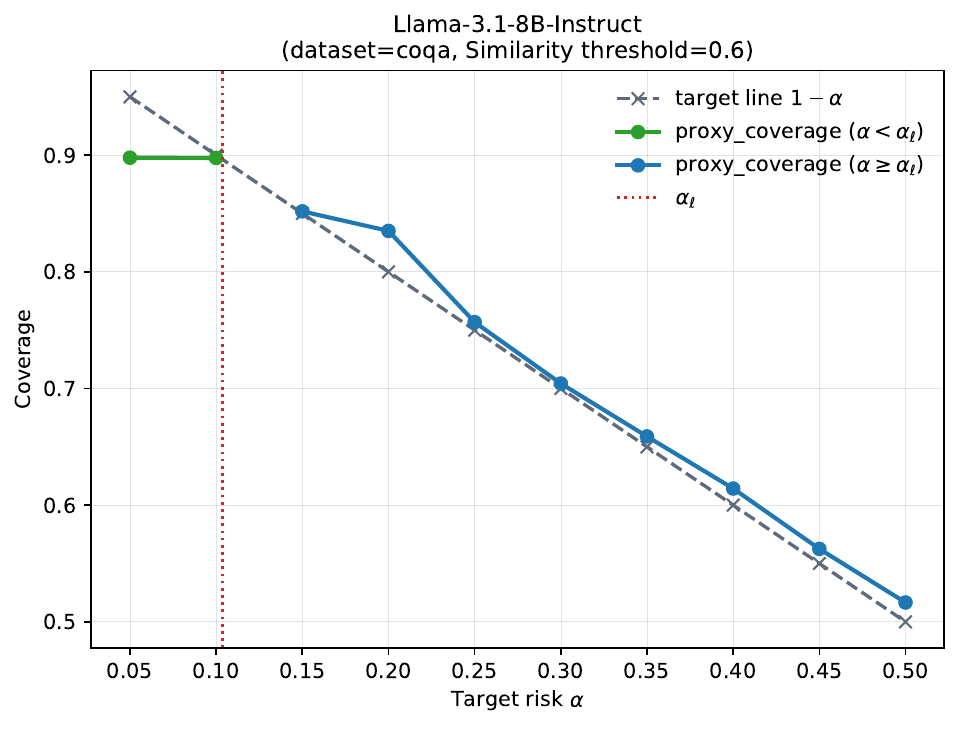}
    % \caption{LLaMA-3.1-8B}
  \end{subfigure}
  \hfill
  \begin{subfigure}[b]{0.195\textwidth}
    \centering
    \includegraphics[width=\textwidth]{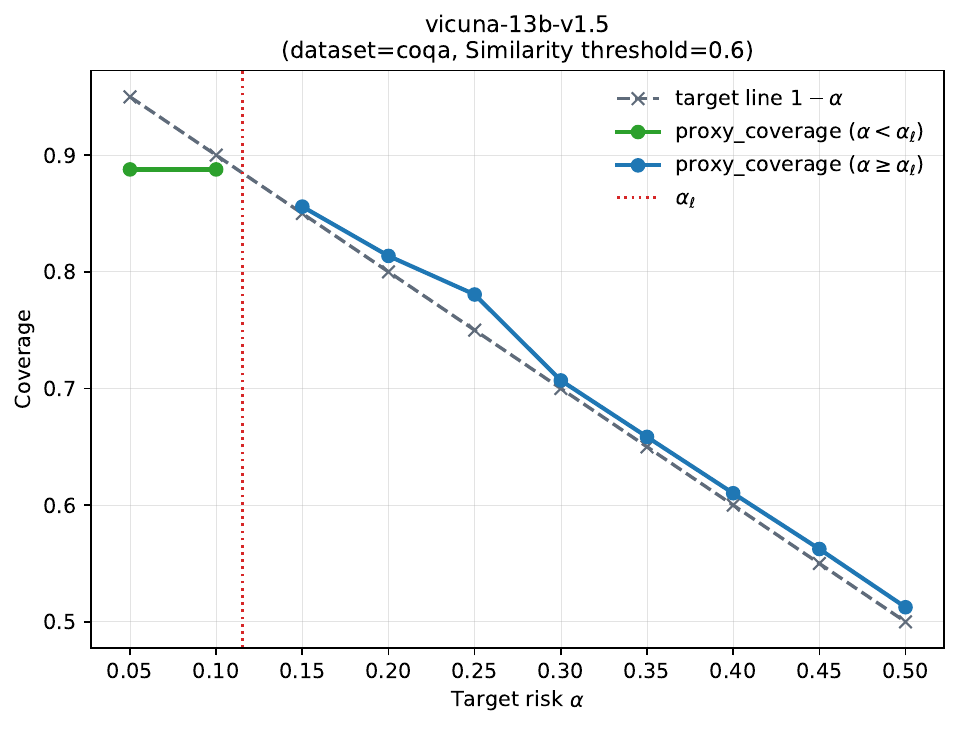}
    % \caption{Vicuna-13B}
  \end{subfigure}
  \hfill
  \begin{subfigure}[b]{0.195\textwidth}
    \centering
    \includegraphics[width=\textwidth]{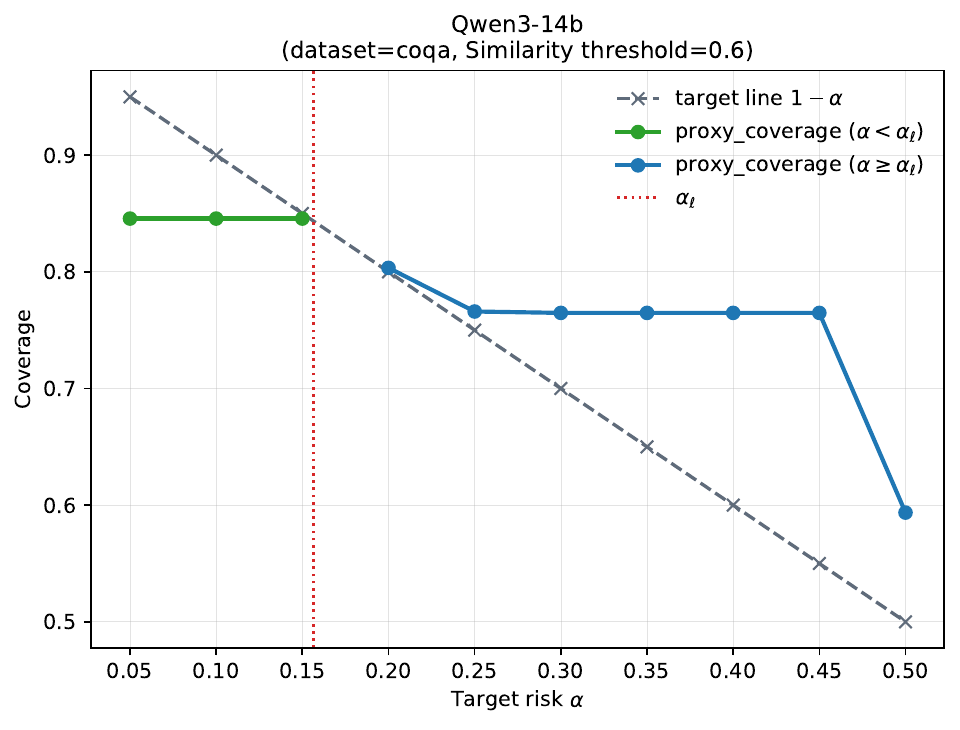}
    % \caption{Qwen3-14B}
  \end{subfigure}

  \begin{subfigure}[b]{0.195\textwidth}
    \centering
    \includegraphics[width=\textwidth]{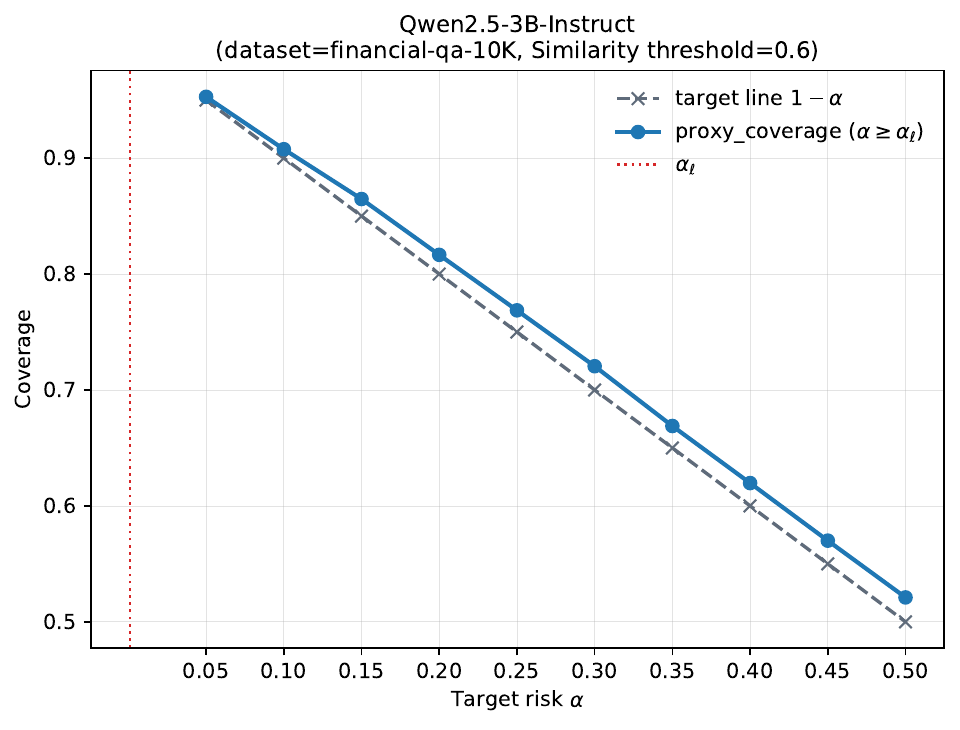}
    % \caption{Qwen2.5-3B}
  \end{subfigure}
  \hfill
  \begin{subfigure}[b]{0.195\textwidth}
    \centering
    \includegraphics[width=\textwidth]{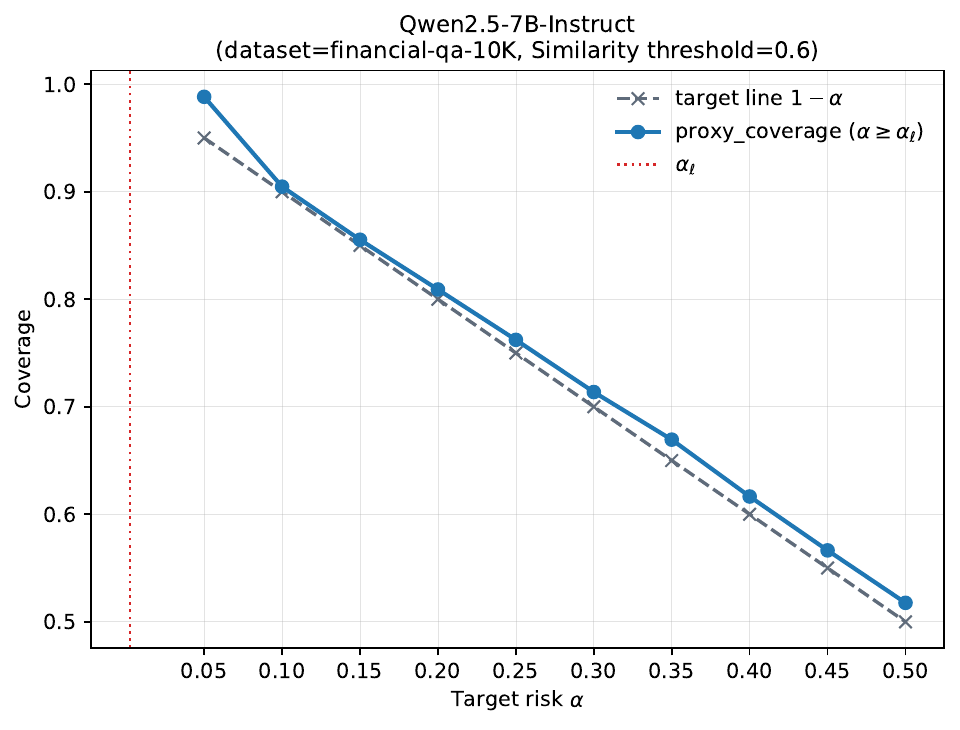}
    % \caption{Qwen2.5-7B}
  \end{subfigure}
  \hfill
  \begin{subfigure}[b]{0.195\textwidth}
    \centering
    \includegraphics[width=\textwidth]{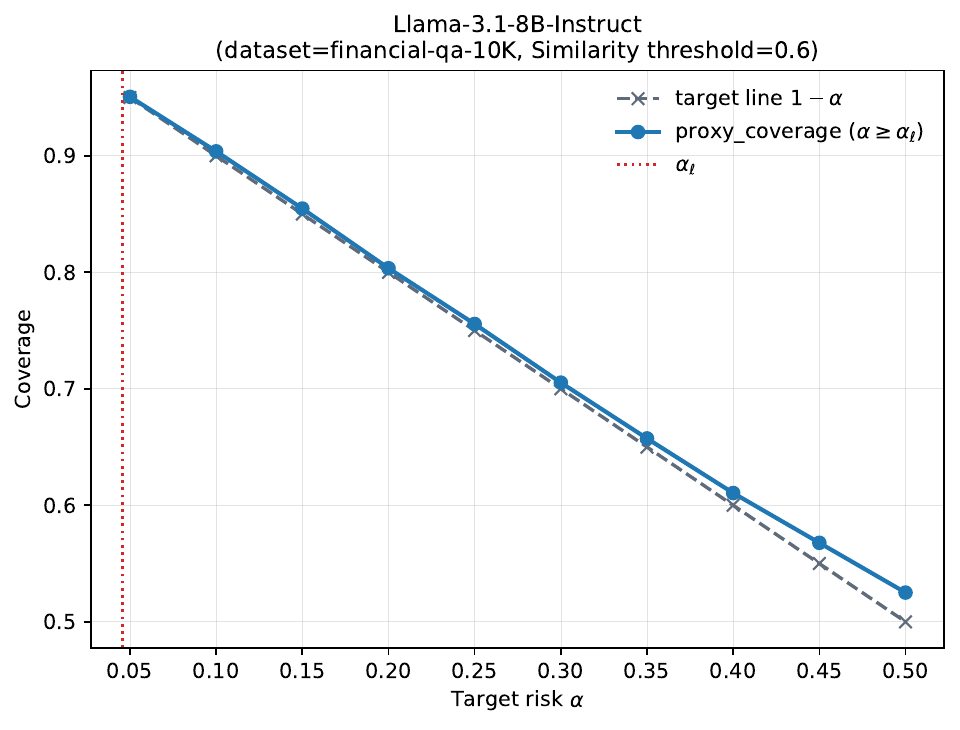}
    % \caption{LLaMA-3.1-8B}
  \end{subfigure}
  \hfill
  \begin{subfigure}[b]{0.195\textwidth}
    \centering
    \includegraphics[width=\textwidth]{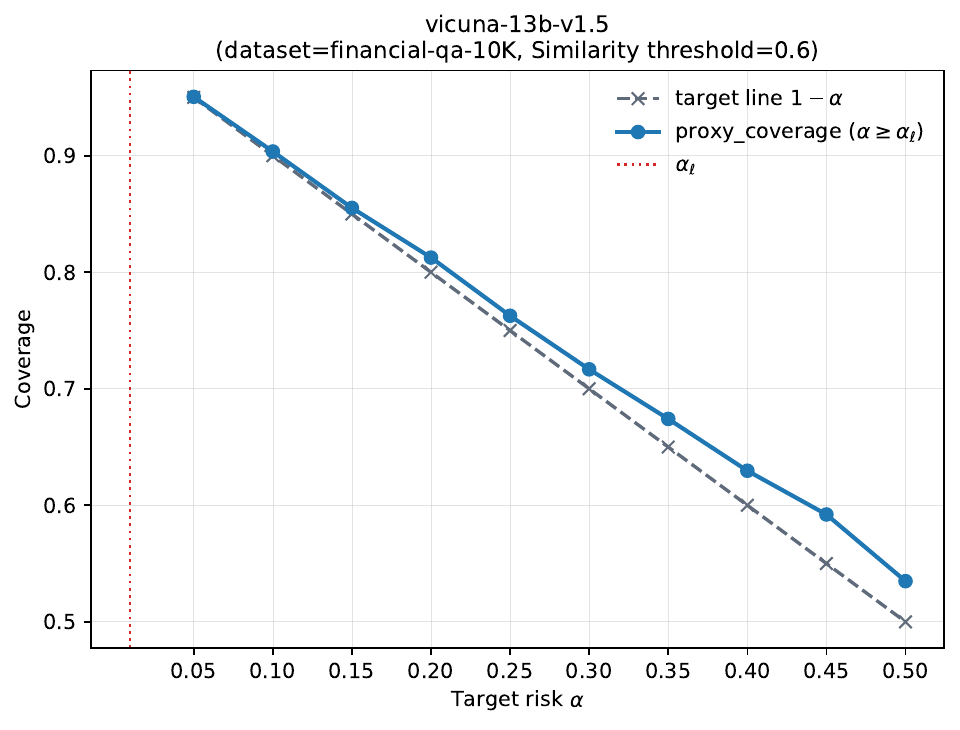}
    % \caption{Vicuna-13B}
  \end{subfigure}
  \hfill
  \begin{subfigure}[b]{0.195\textwidth}
    \centering
    \includegraphics[width=\textwidth]{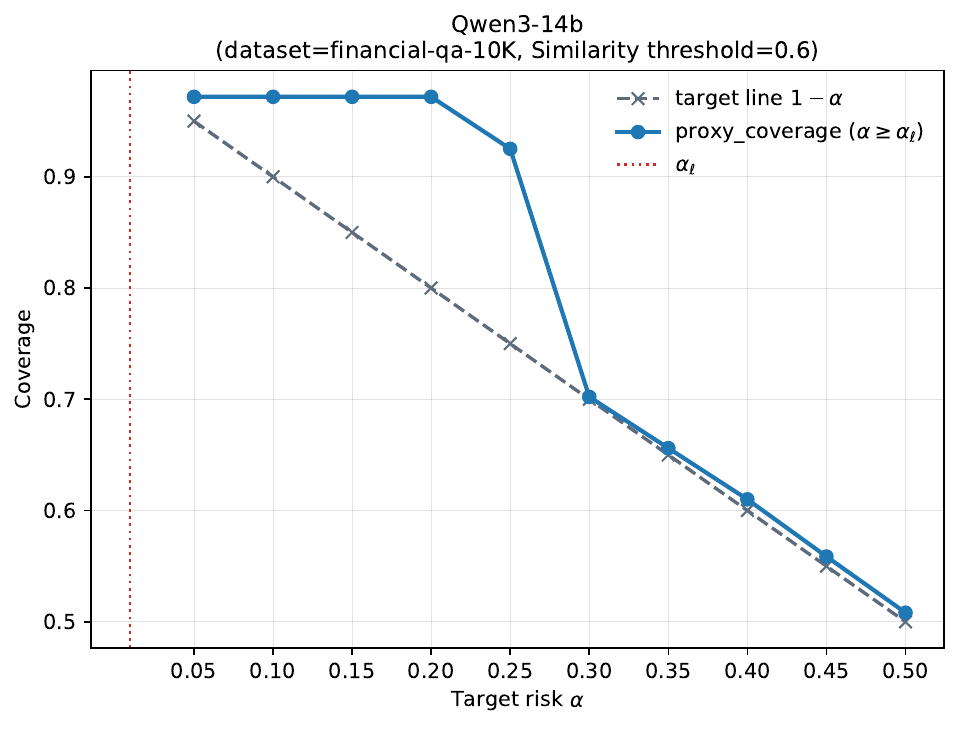}
    % \caption{Qwen3-14B}
  \end{subfigure}

  \begin{subfigure}[b]{0.195\textwidth}
    \centering
    \includegraphics[width=\textwidth]{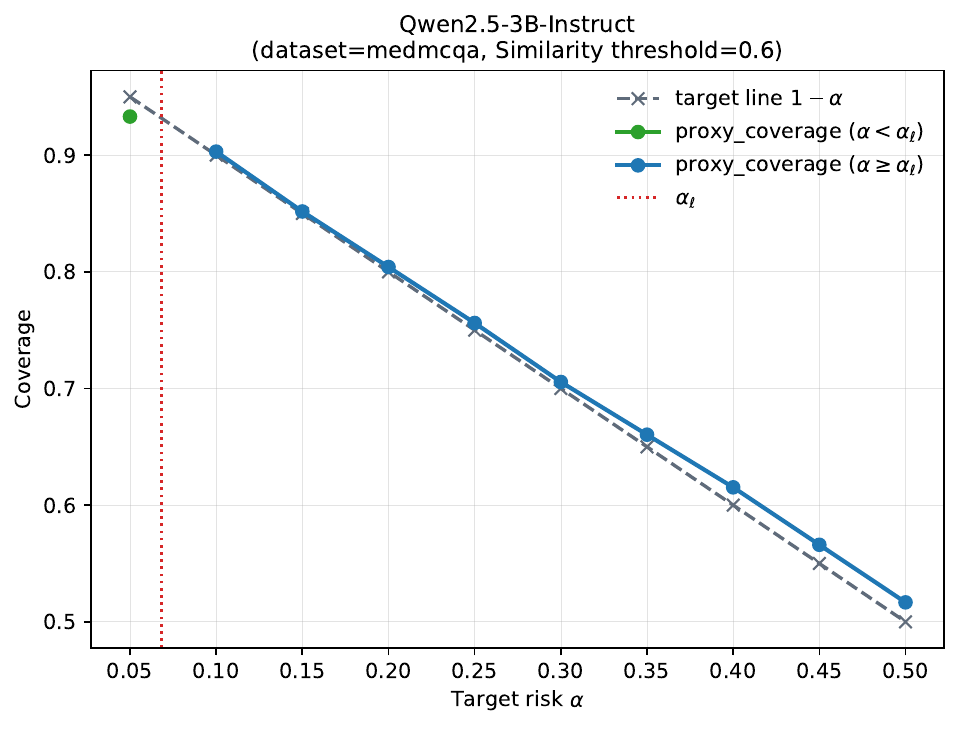}
    % \caption{Qwen2.5-3B}
  \end{subfigure}
  \hfill
  \begin{subfigure}[b]{0.195\textwidth}
    \centering
    \includegraphics[width=\textwidth]{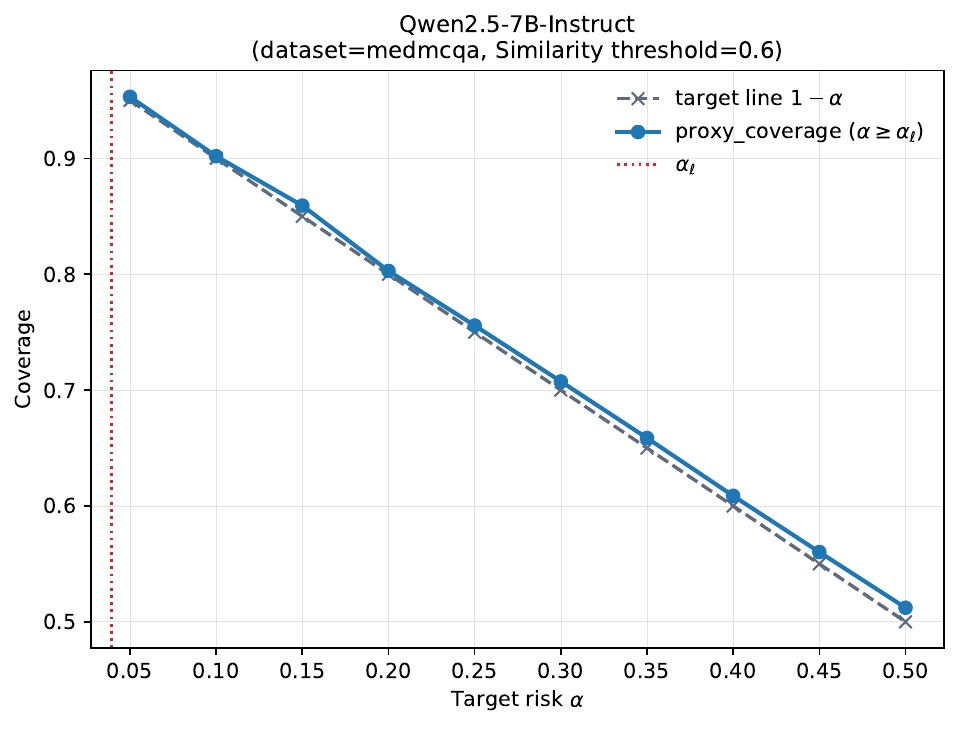}
    % \caption{Qwen2.5-7B}
  \end{subfigure}
  \hfill
  \begin{subfigure}[b]{0.195\textwidth}
    \centering
    \includegraphics[width=\textwidth]{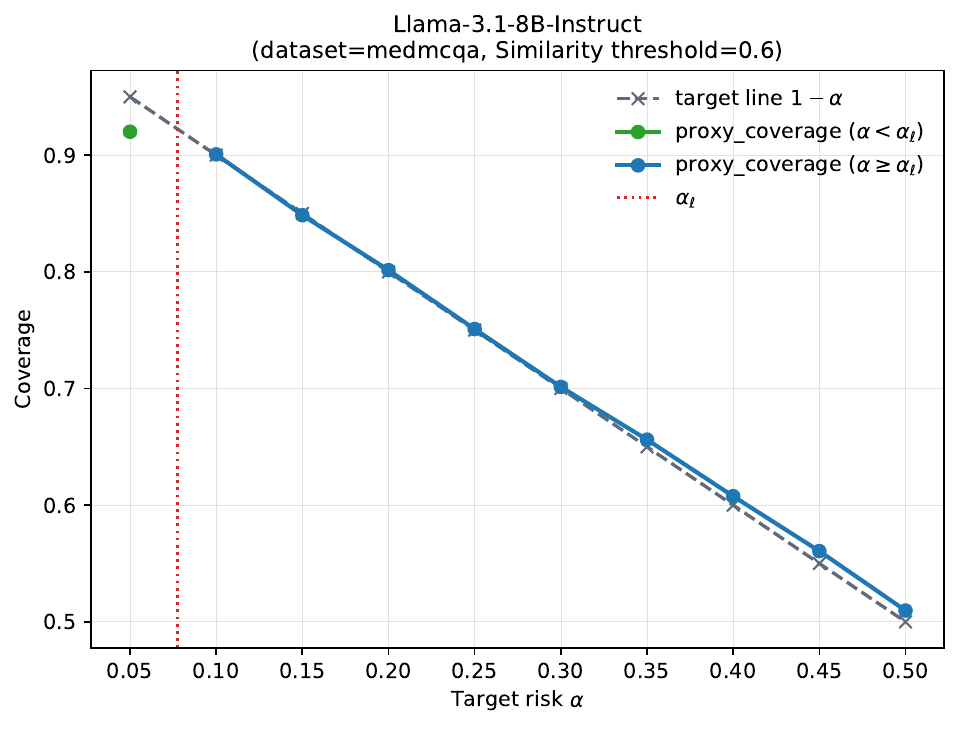}
    % \caption{LLaMA-3.1-8B}
  \end{subfigure}
  \hfill
  \begin{subfigure}[b]{0.195\textwidth}
    \centering
    \includegraphics[width=\textwidth]{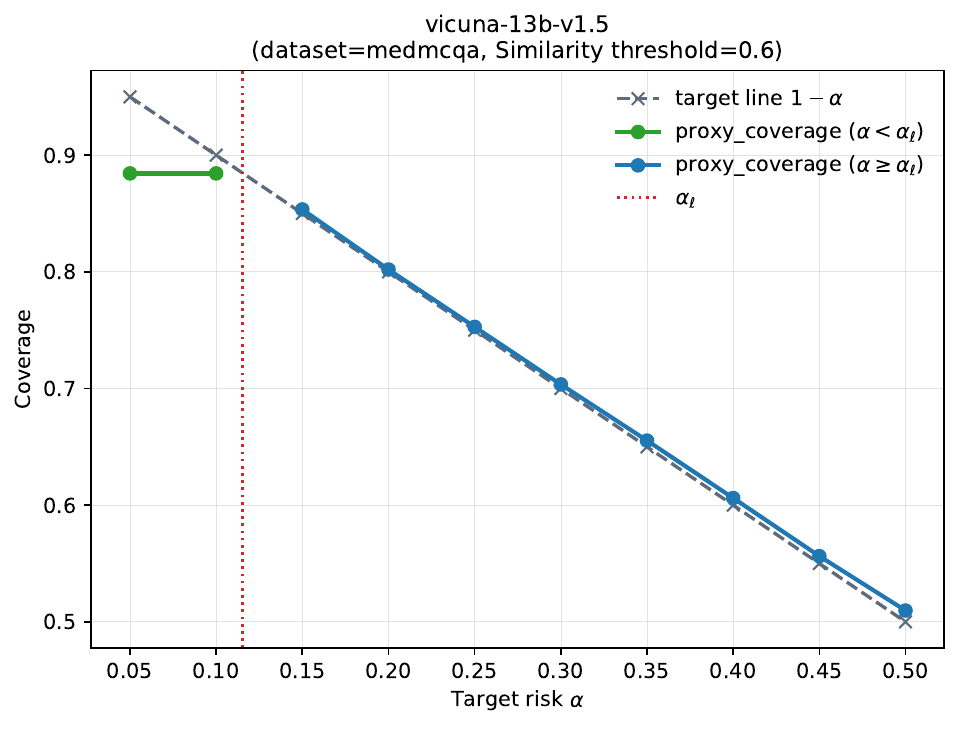}
    % \caption{Vicuna-13B}
  \end{subfigure}
  \hfill
  \begin{subfigure}[b]{0.195\textwidth}
    \centering
    \includegraphics[width=\textwidth]{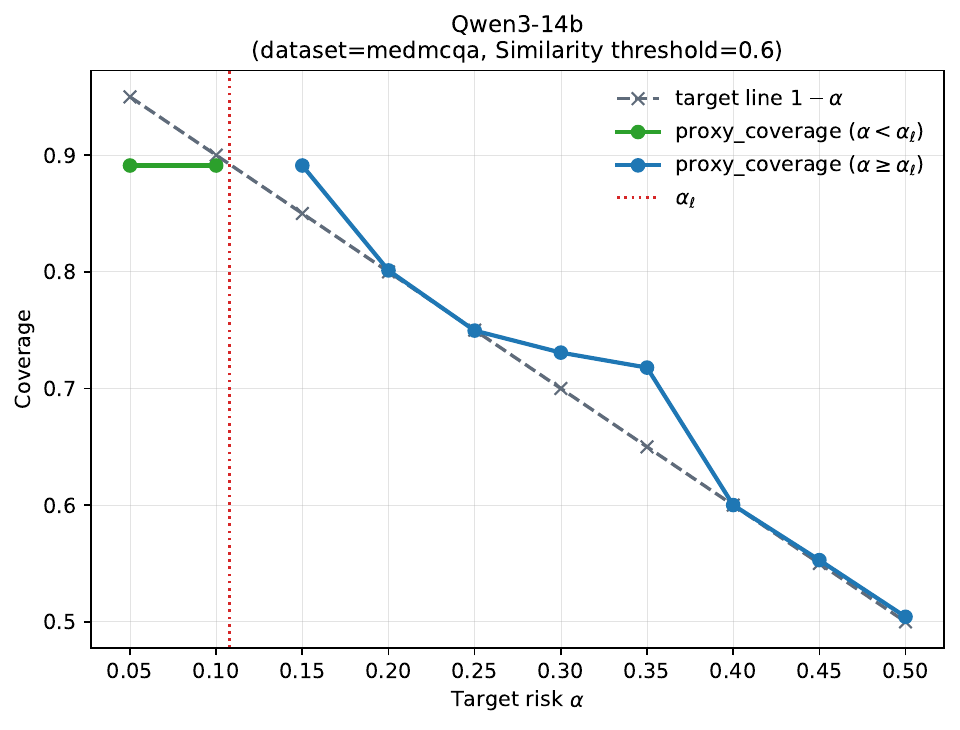}
    % \caption{Qwen3-14B}
  \end{subfigure}

  \begin{subfigure}[b]{0.195\textwidth}
    \centering
    \includegraphics[width=\textwidth]{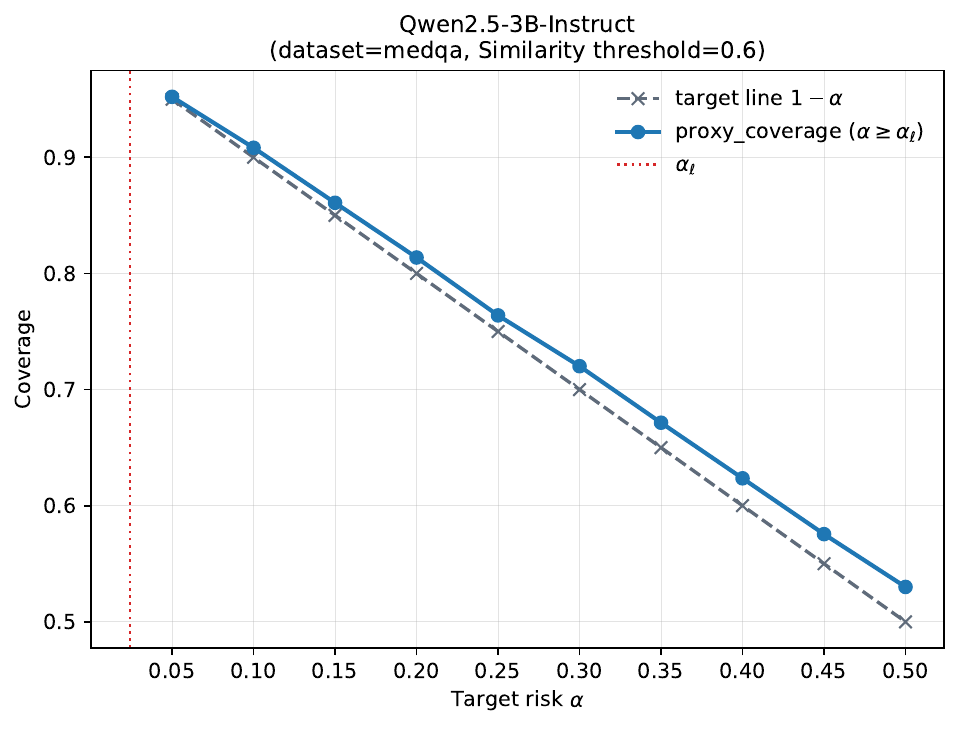}
    % \caption{Qwen2.5-3B}
  \end{subfigure}
  \hfill
  \begin{subfigure}[b]{0.195\textwidth}
    \centering
    \includegraphics[width=\textwidth]{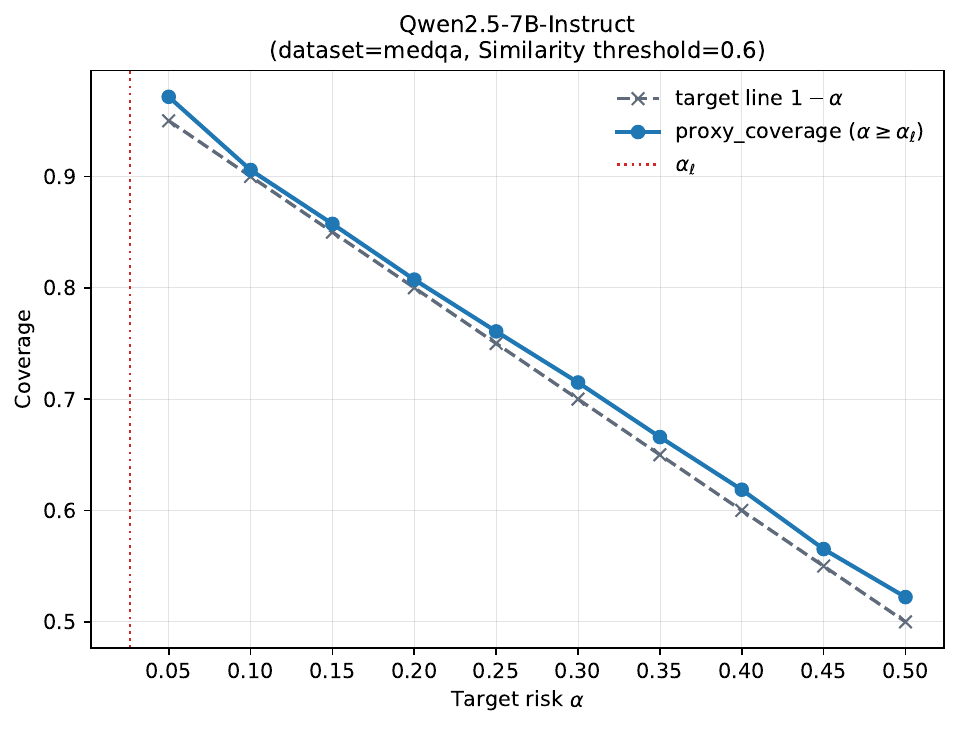}
    % \caption{Qwen2.5-7B}
  \end{subfigure}
  \hfill
  \begin{subfigure}[b]{0.195\textwidth}
    \centering
    \includegraphics[width=\textwidth]{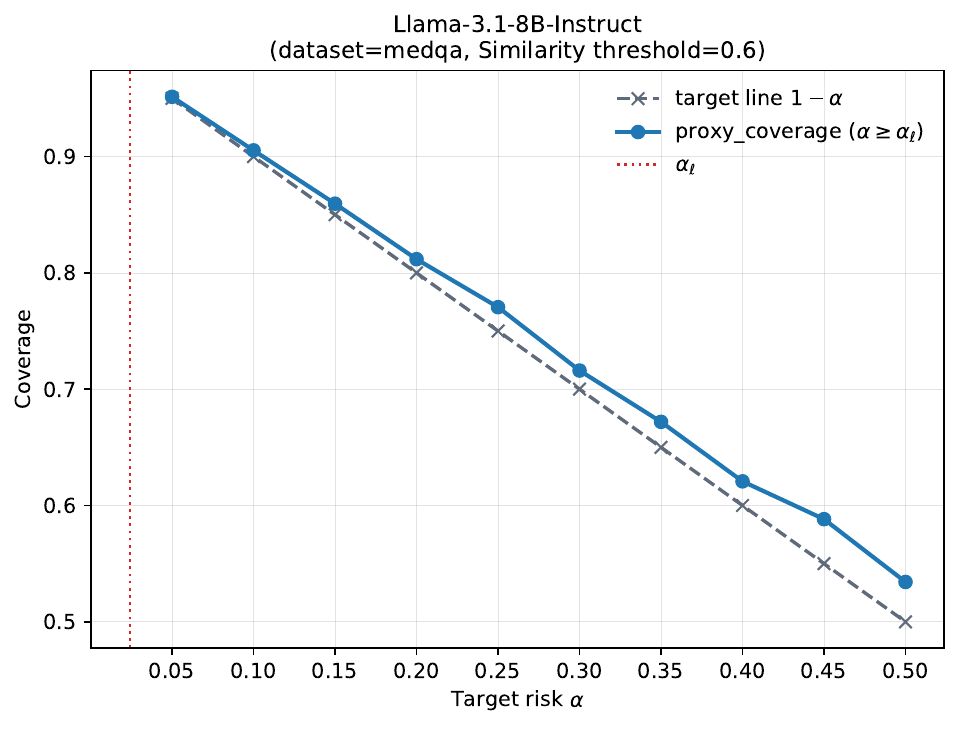}
    % \caption{LLaMA-3.1-8B}
  \end{subfigure}
  \hfill
  \begin{subfigure}[b]{0.195\textwidth}
    \centering
    \includegraphics[width=\textwidth]{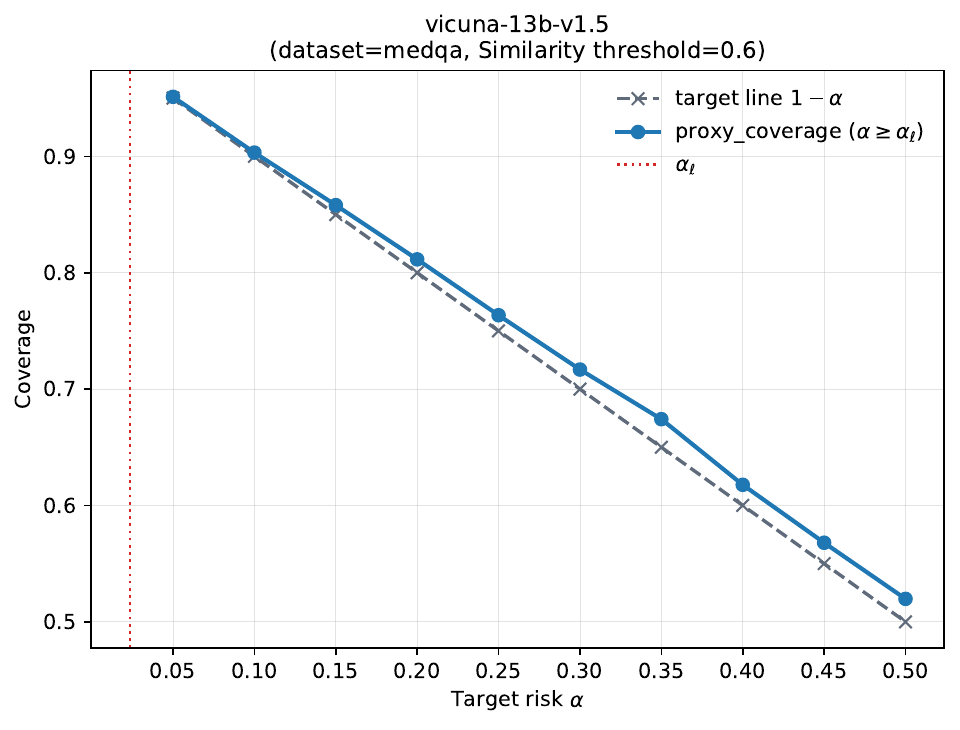}
    % \caption{Vicuna-13B}
  \end{subfigure}
  \hfill
  \begin{subfigure}[b]{0.195\textwidth}
    \centering
    \includegraphics[width=\textwidth]{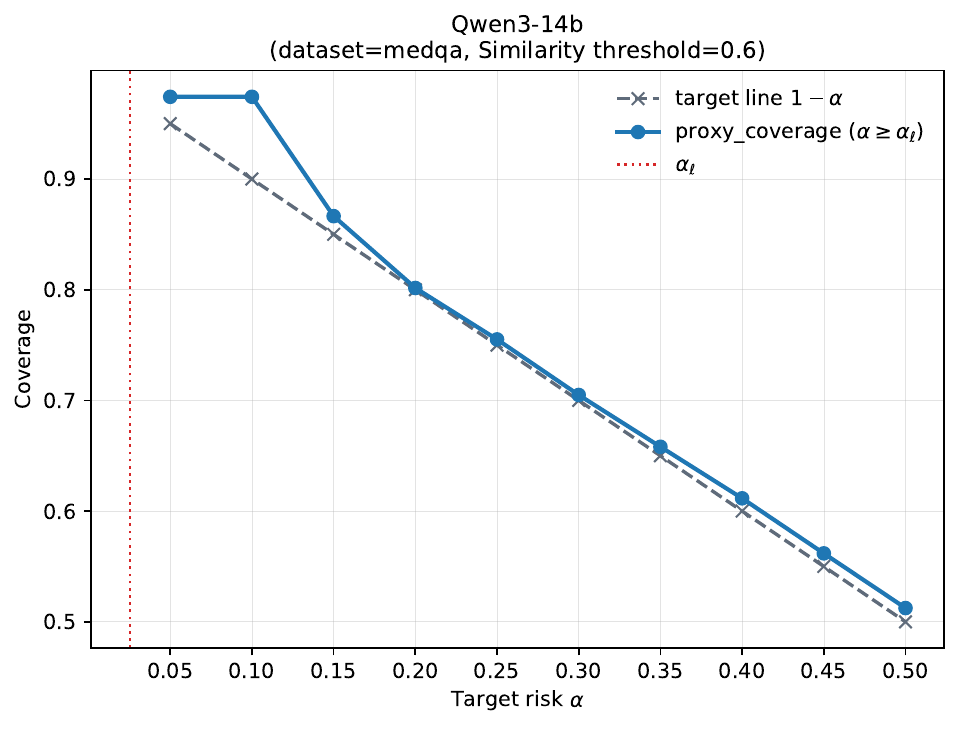}
    % \caption{Qwen3-14B}
  \end{subfigure}

  \begin{subfigure}[b]{0.195\textwidth}
    \centering
    \includegraphics[width=\textwidth]{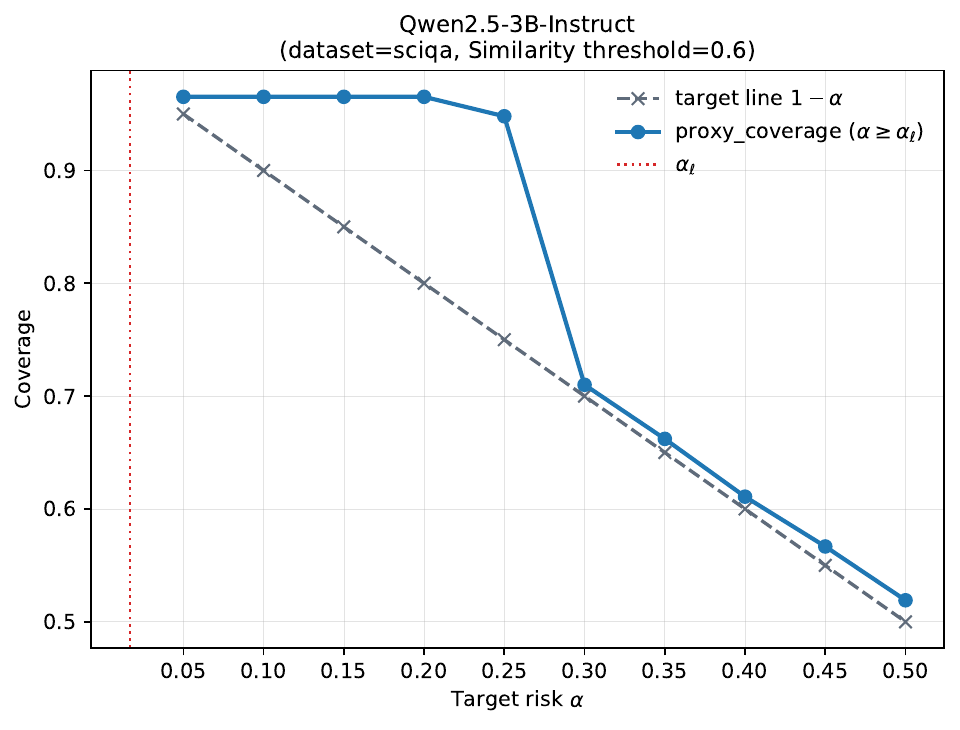}
    % \caption{Qwen2.5-3B}
  \end{subfigure}
  \hfill
  \begin{subfigure}[b]{0.195\textwidth}
    \centering
    \includegraphics[width=\textwidth]{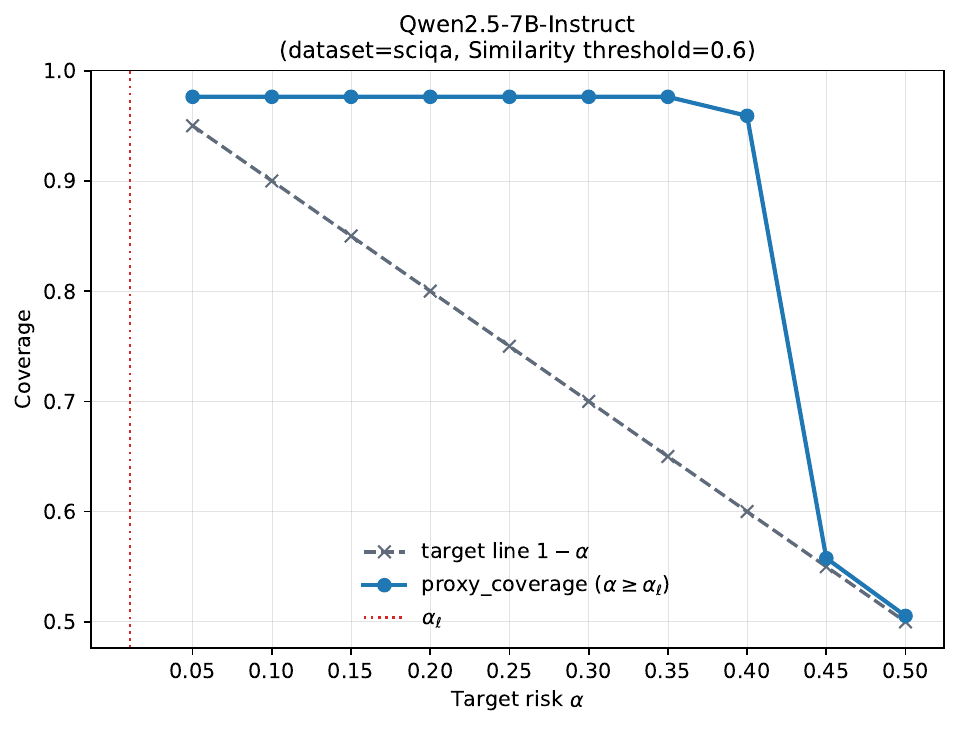}
    % \caption{Qwen2.5-7B}
  \end{subfigure}
  \hfill
  \begin{subfigure}[b]{0.195\textwidth}
    \centering
    \includegraphics[width=\textwidth]{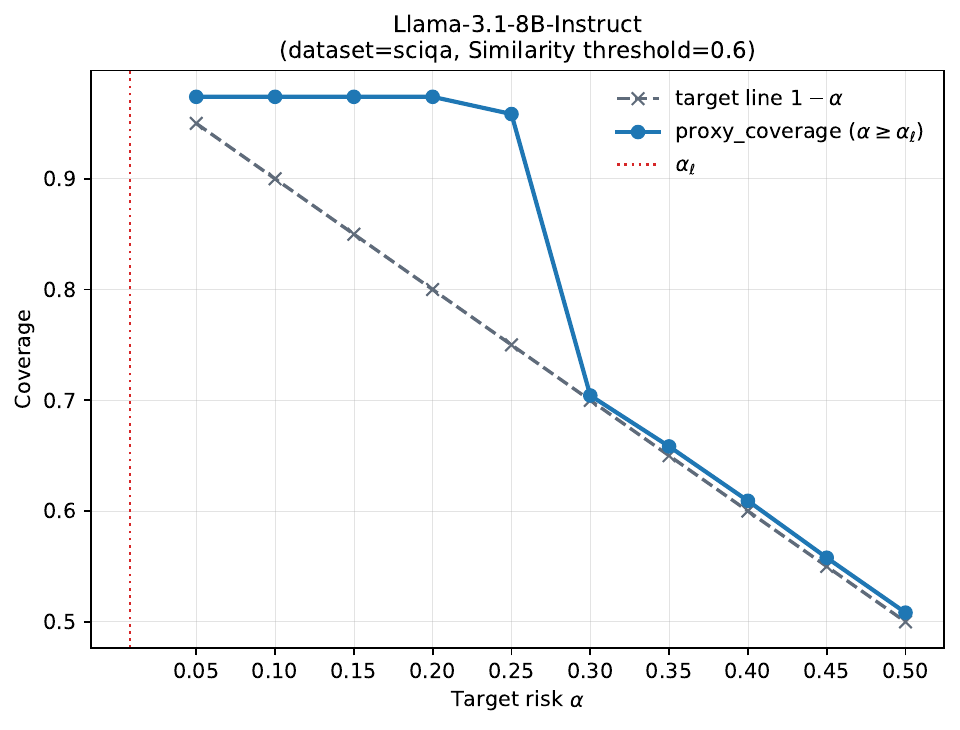}
    % \caption{LLaMA-3.1-8B}
  \end{subfigure}
  \hfill
  \begin{subfigure}[b]{0.195\textwidth}
    \centering
    \includegraphics[width=\textwidth]{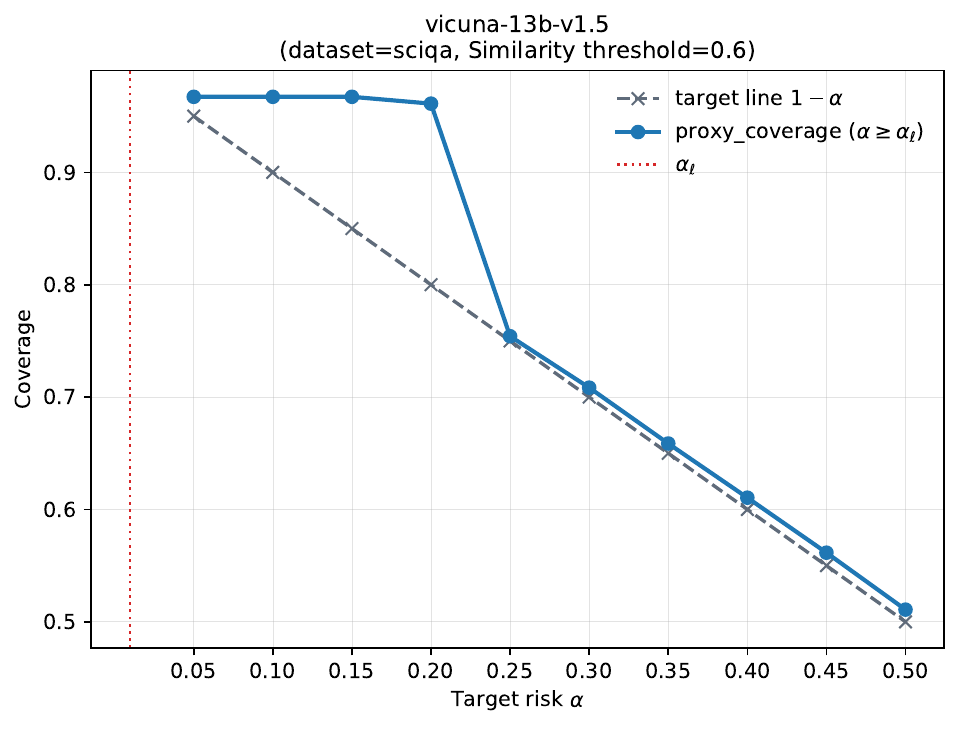}
    % \caption{Vicuna-13B}
  \end{subfigure}
  \hfill
  \begin{subfigure}[b]{0.195\textwidth}
    \centering
    \includegraphics[width=\textwidth]{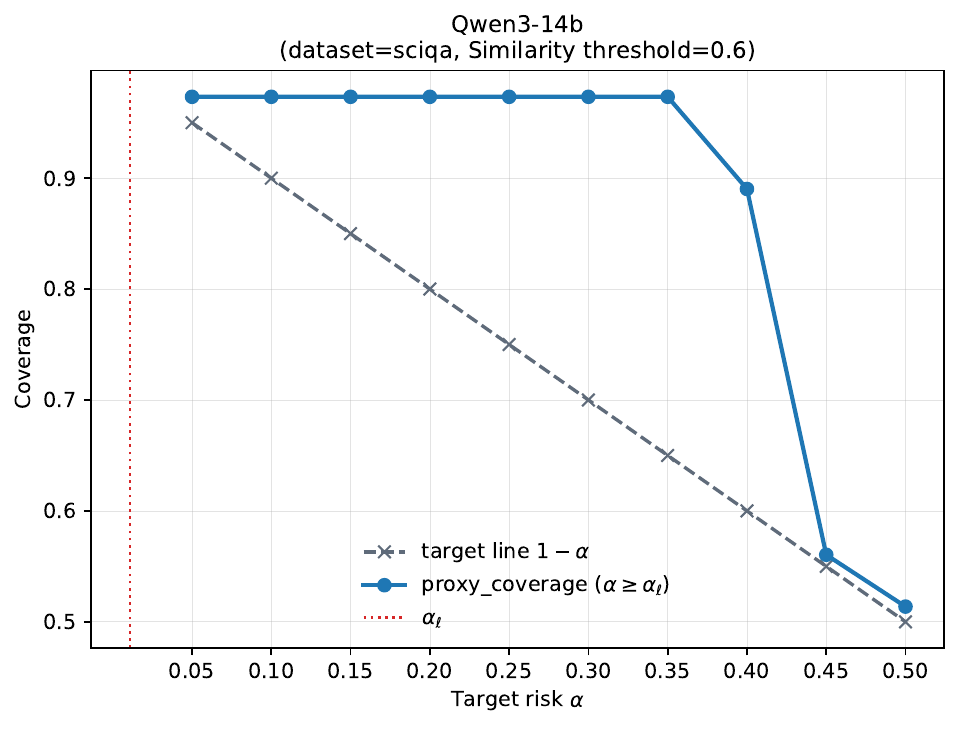}
    % \caption{Qwen3-14B}
  \end{subfigure}

  \begin{subfigure}[b]{0.195\textwidth}
    \centering
    \includegraphics[width=\textwidth]{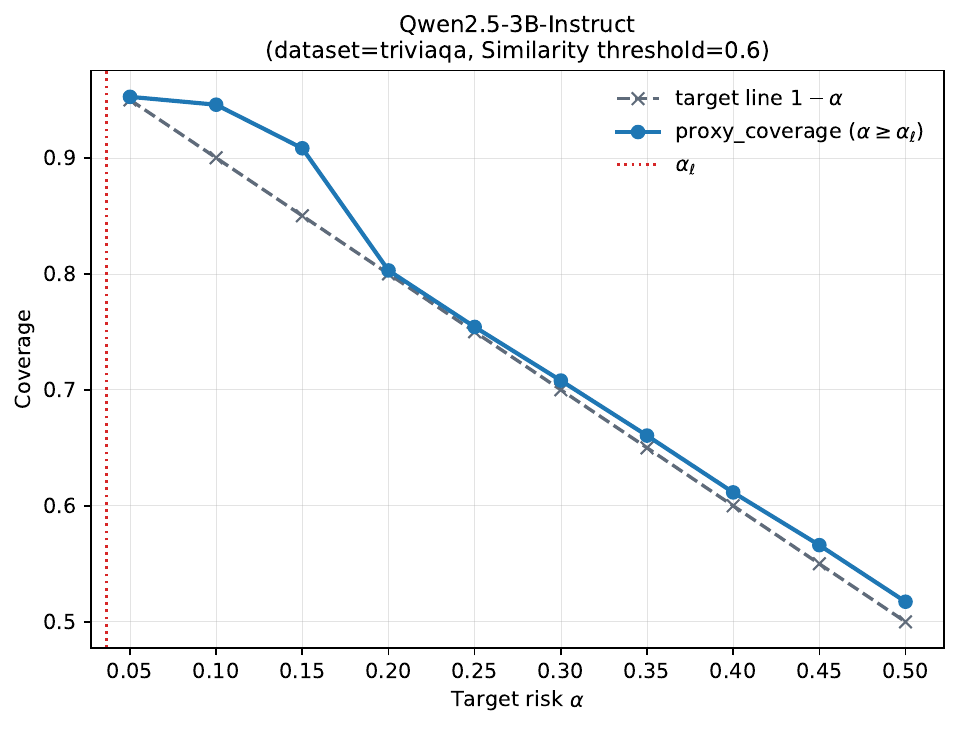}
    \caption{Qwen2.5-3B}
  \end{subfigure}
  \hfill
  \begin{subfigure}[b]{0.195\textwidth}
    \centering
    \includegraphics[width=\textwidth]{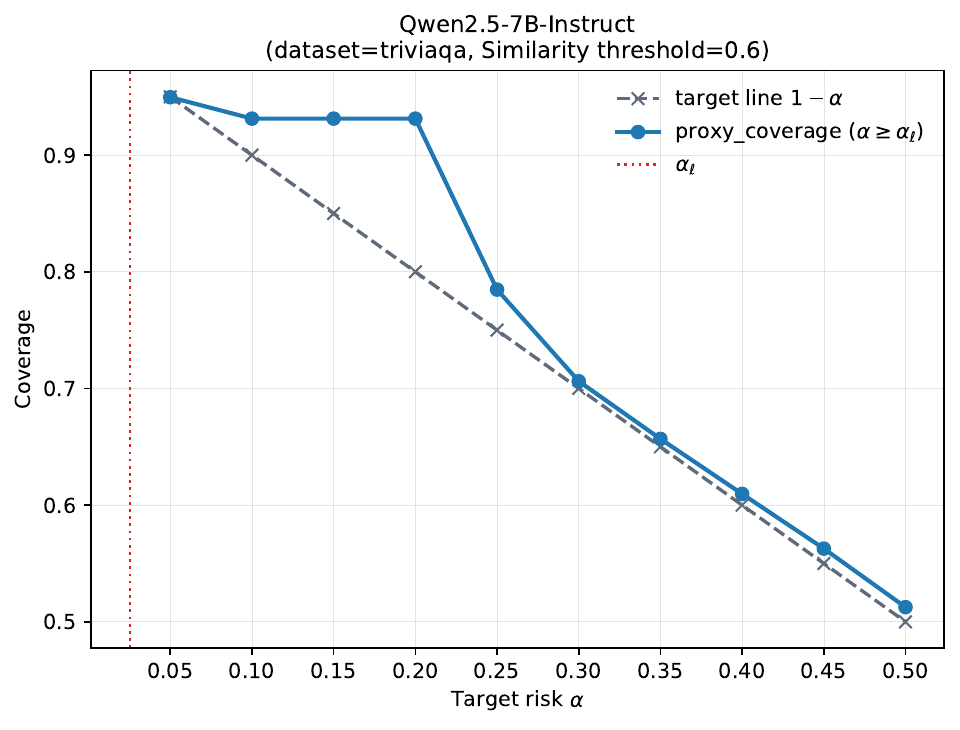}
    \caption{Qwen2.5-7B}
  \end{subfigure}
  \hfill
  \begin{subfigure}[b]{0.195\textwidth}
    \centering
    \includegraphics[width=\textwidth]{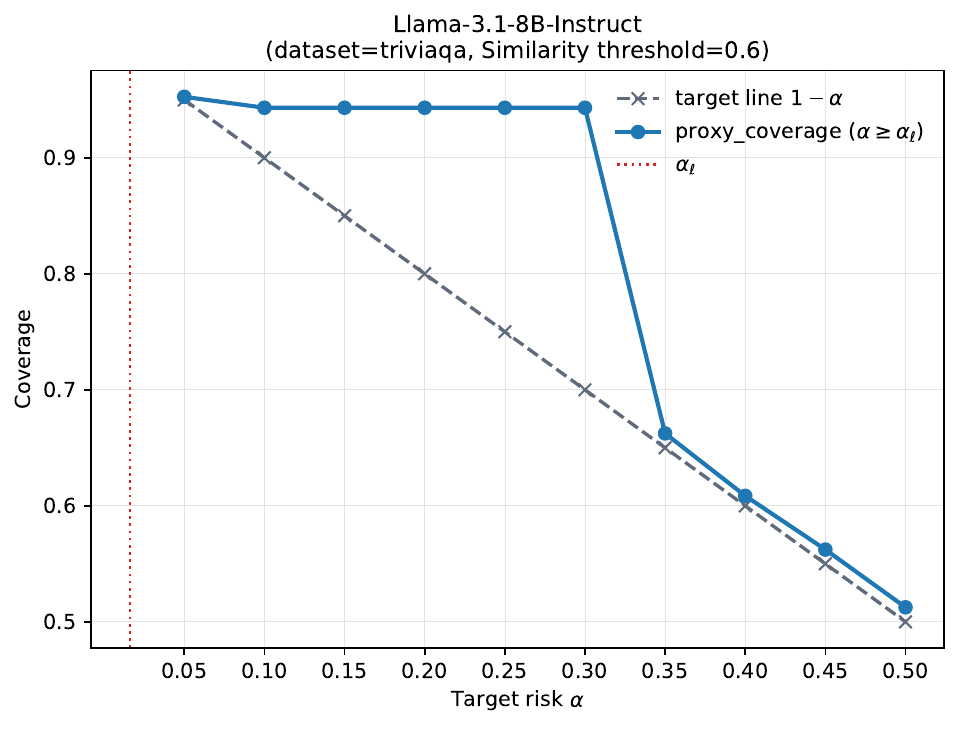}
    \caption{LLaMA-3.1-8B}
  \end{subfigure}
  \hfill
  \begin{subfigure}[b]{0.195\textwidth}
    \centering
    \includegraphics[width=\textwidth]{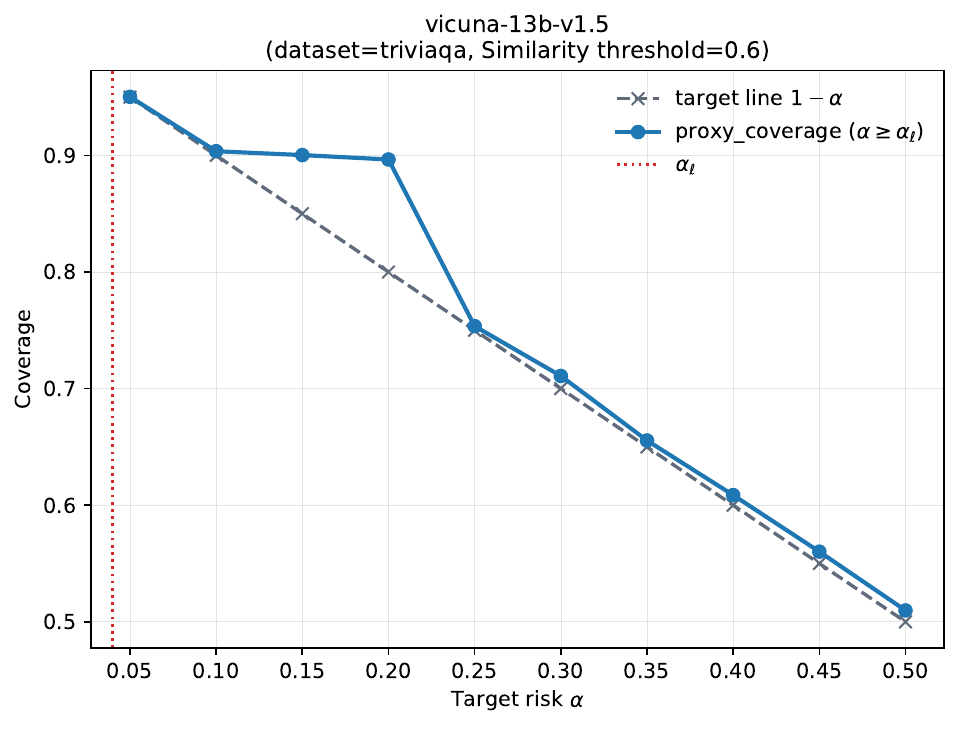}
    \caption{Vicuna-13B}
  \end{subfigure}
  \hfill
  \begin{subfigure}[b]{0.195\textwidth}
    \centering
    \includegraphics[width=\textwidth]{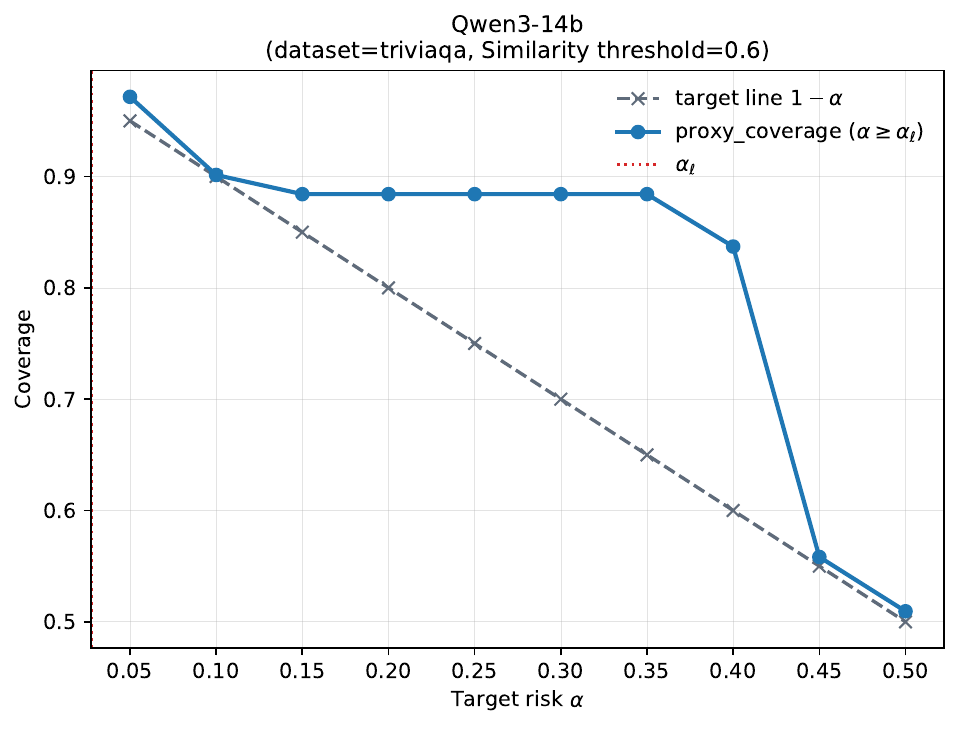}
    \caption{Qwen3-14B}
  \end{subfigure}

      \caption{Coverage Guarantees on six NLG benchmarks utilizing five LLMs. The threshold of sentence similarity is fixed at 0.6.}
  \label{fig: Coverage Guarantees (0.6)}
\end{figure*}

\paragraph{2) Coverage guarantees are realized only in the feasible regime.}
We next examine whether the proposed feasibility-aware guarantee is borne out empirically. Figure~\ref{fig: Coverage Guarantees (0.7)} reports empirical coverage across six datasets and five LLMs under the default semantic admission threshold. The result is highly consistent: when the target risk falls below $\alpha_l$, the desired coverage guarantee is typically violated; once the target risk enters the feasible regime, i.e., $\alpha \ge \alpha_l$, empirical coverage closely tracks or stays above the target line $1-\alpha$.

This transition is precisely the behavior predicted by our theory. In the infeasible regime, the calibrated predictor cannot overcome the finite-sampling failure mode, since some sampled candidate sets contain no admissible answer at all. As a result, no downstream thresholding rule can recover the nominal coverage. By contrast, once $\alpha$ exceeds $\alpha_l$, the learn-then-test calibration successfully turns the sampled candidate pool into a statistically valid prediction set. Therefore, Figure~3 confirms that $\alpha_l$ is not merely a conservative statistic, but an operational feasibility boundary that sharply separates unattainable and attainable coverage regimes.

\paragraph{3) The same feasibility-aware pattern persists across semantic admission thresholds.}
Figure~\ref{fig: Coverage Guarantees (0.5)} and Figure~\ref{fig: Coverage Guarantees (0.6)} repeat the same coverage analysis under alternative semantic admission thresholds. Although the absolute coverage levels and the location of $\alpha_l$ vary with the admission criterion, the qualitative conclusion remains unchanged: the guarantee is generally violated below $\alpha_l$ and consistently recovered once $\alpha \ge \alpha_l$. This shows that the feasibility-aware nature of the proposed framework is robust to the specific semantic strictness used to define admissibility.

\begin{figure*}[!t]
    \centering
    \includegraphics[width=0.8\linewidth]{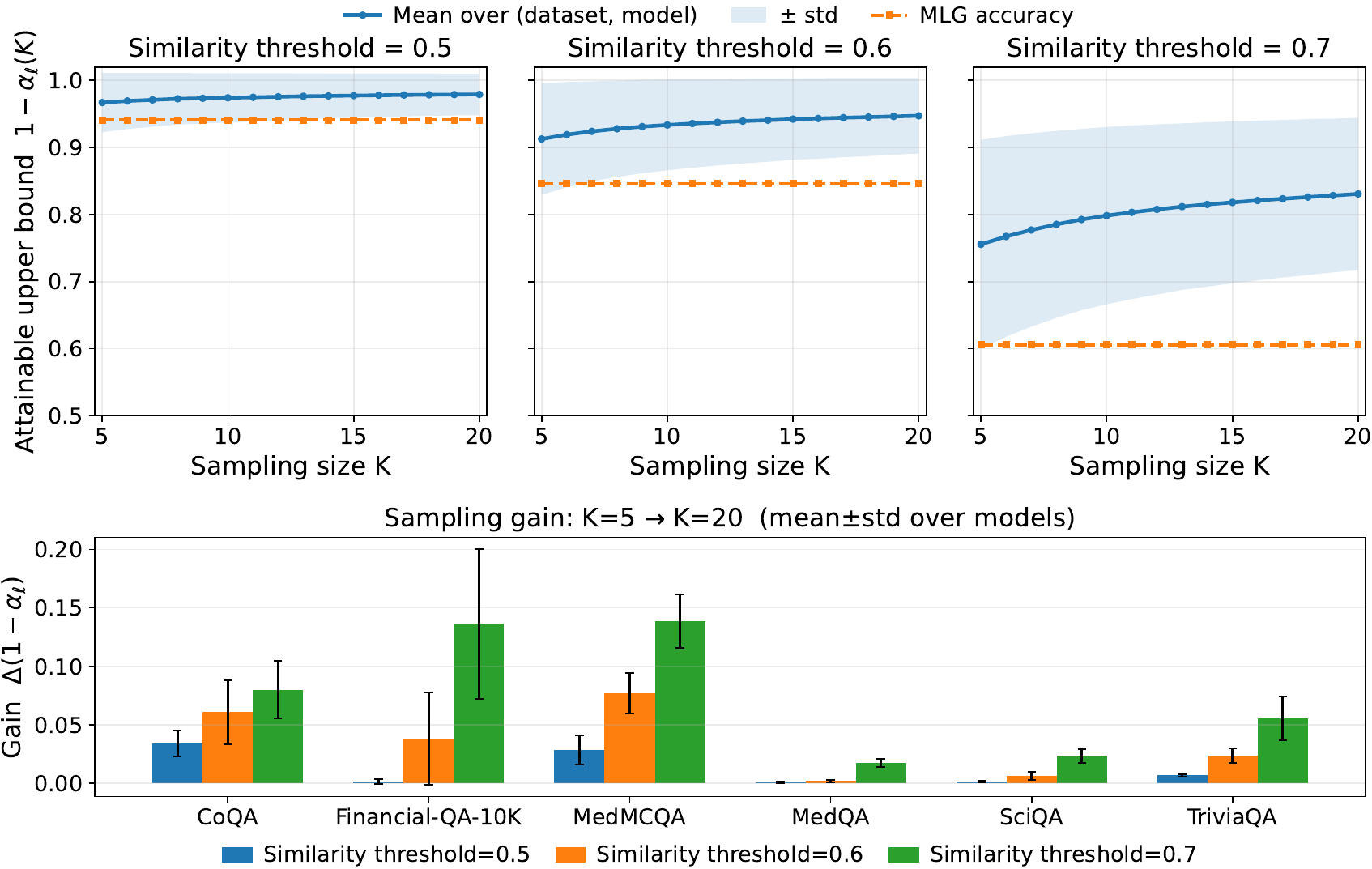}
    \caption{Sampling budget shifts the feasibility boundary. We plot the attainable upper bound $1-\alpha_l(K)$ as a function of the sampling budget $K$ under different semantic admission thresholds. A larger $K$ lowers the MRL $\alpha_l(K)$, thereby enlarging the feasible region for coverage guarantees. The improvement is consistent across settings but gradually saturates.}
    \label{fig: rq3}
\end{figure*}

\begin{figure*}[!t]
    \centering
    \includegraphics[width=0.8\linewidth]{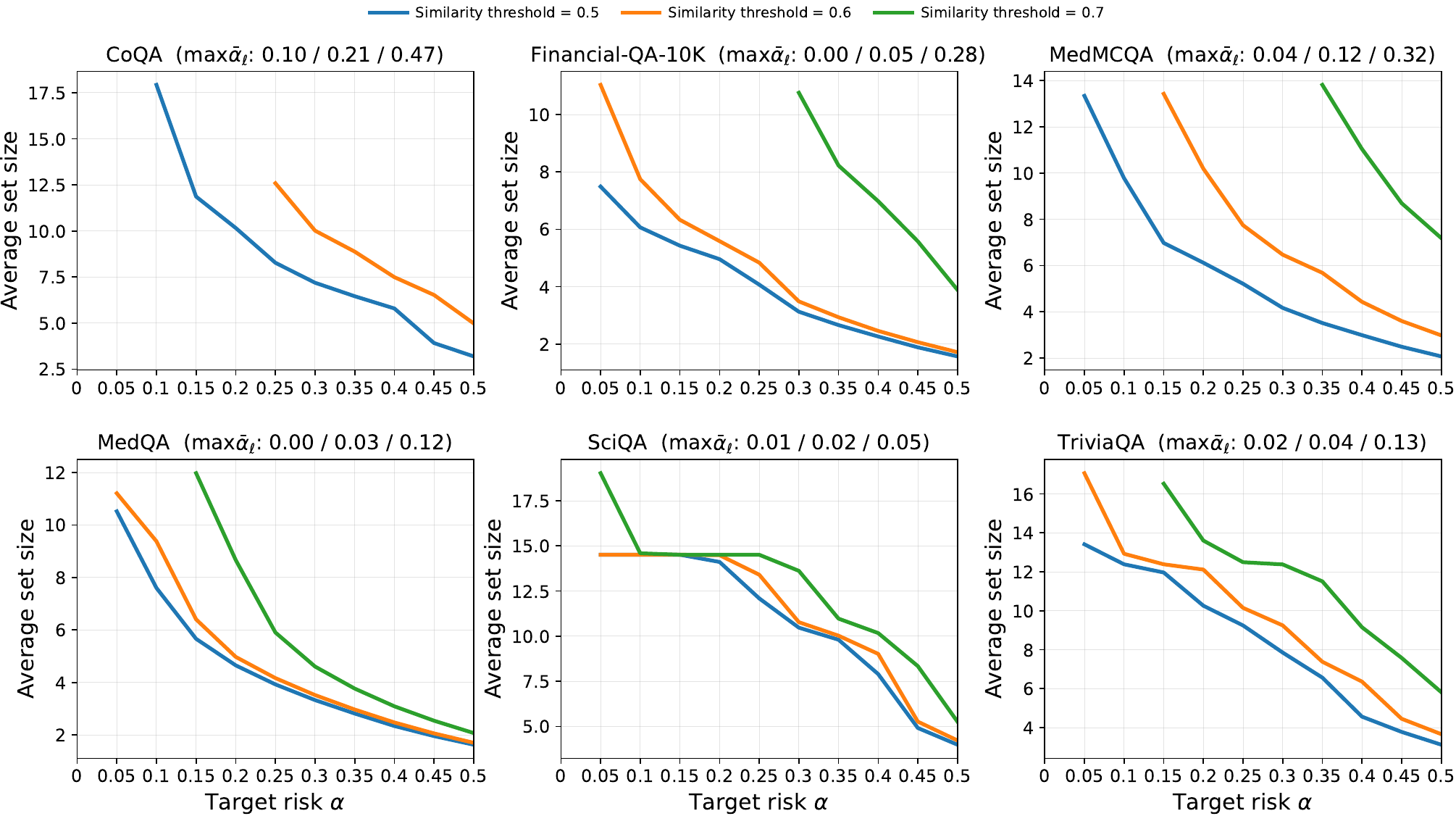}
    \caption{APSS vs. risk level. APSS decreases monotonically as $\alpha$ increases, reflecting the trade-off between prediction efficiency and coverage guarantee. Stricter admission thresholds generally require larger prediction sets and induce a higher MRL $\alpha_l$.}
    \label{fig: rq4}
\end{figure*}

\paragraph{4) The sampling budget directly shifts the feasibility boundary.}
We further investigate how the sampling budget affects the attainable risk floor. As the sampling budget increases, the model has more opportunities to generate at least one admissible answer, and the corresponding feasibility boundary shifts accordingly. The results in figure~\ref{fig: rq3} show that larger sampling budgets consistently lower $\alpha_l$, thereby enlarging the feasible region for coverage guarantees. This confirms that the sampling budget is a direct control knob for feasibility: stronger sampling alleviates the finite-sampling failure mode and makes more stringent risks attainable.

At the same time, the gain is not linear. The improvement in attainability becomes gradually smaller as the candidate pool grows, revealing clear diminishing returns. This suggests that increasing the sampling budget is effective but should be balanced against computational cost, since beyond moderate values, additional samples mainly provide marginal reductions in the risk floor.

\begin{figure*}[!t]
    \centering
    \includegraphics[width=0.8\linewidth]{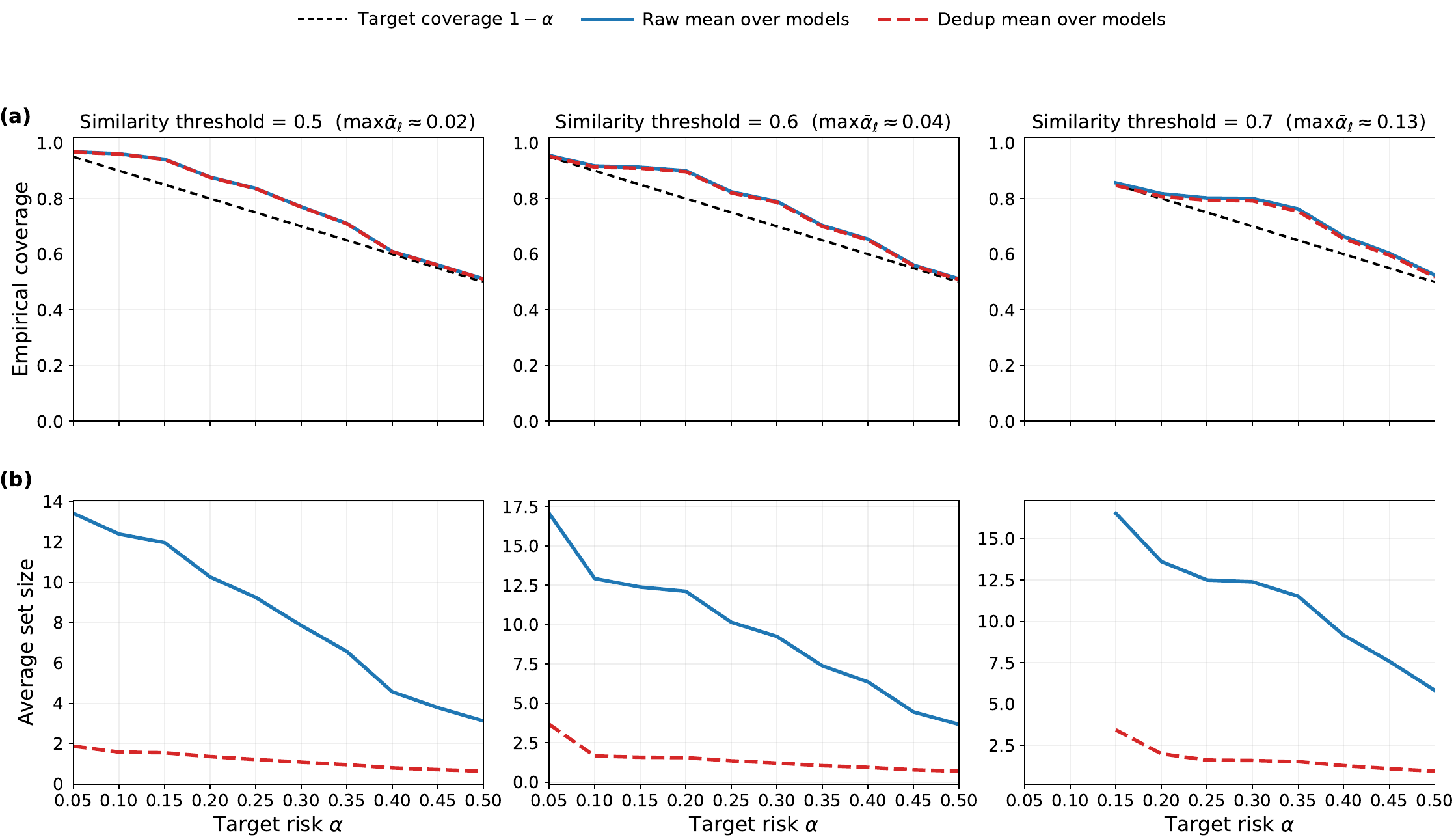}
    \caption{Effect of semantic deduplication on prediction efficiency. We merge responses with sentence similarity above $0.9$ within each prediction set.}
    \label{fig: rq5}
\end{figure*}

\begin{figure*}[!t]
    \centering
    \includegraphics[width=0.8\linewidth]{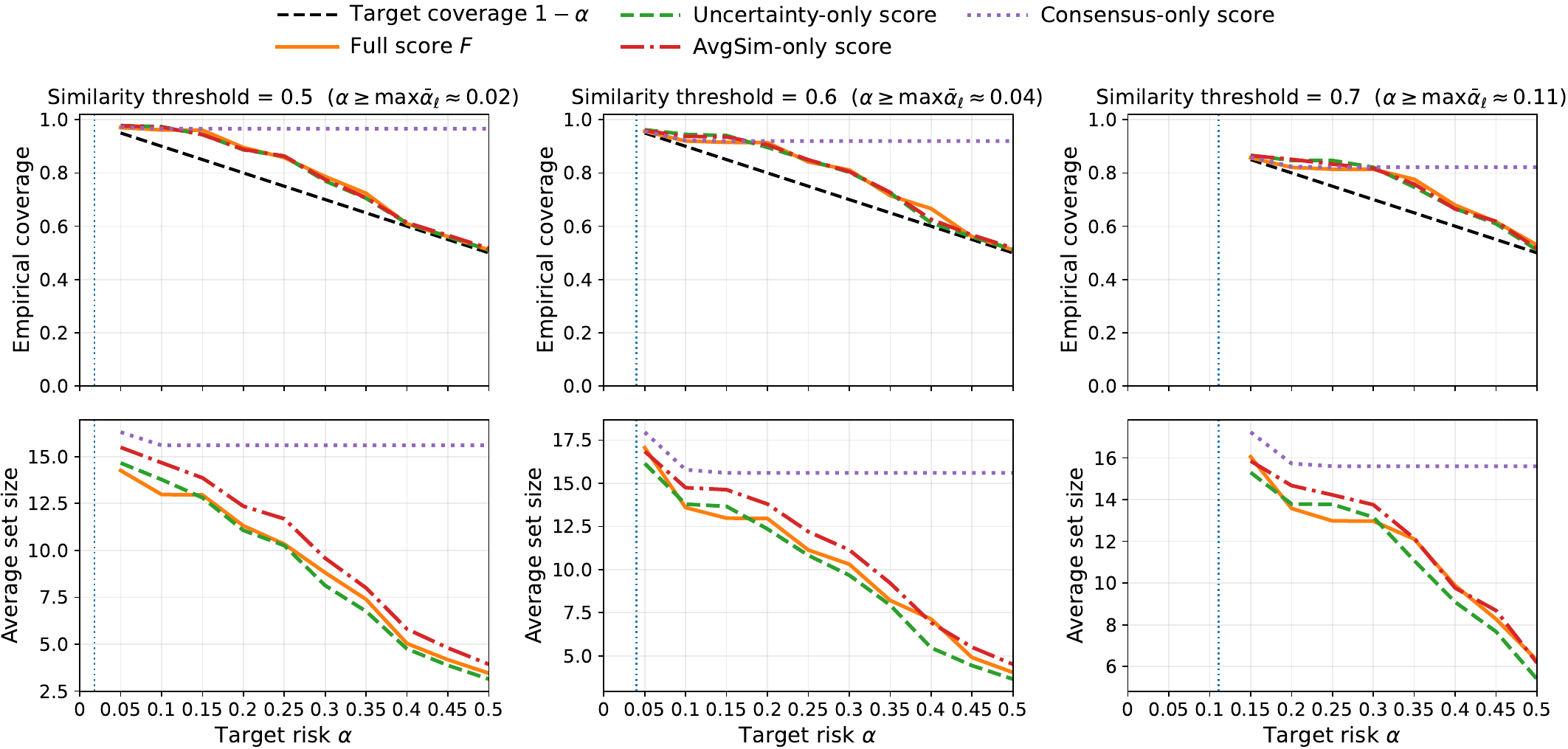}
    \caption{Ablation of reliability-score components in the confidence evaluation function $F$.}
    \label{fig: rq6}
\end{figure*}

\paragraph{5) APSS reveals the coverage--efficiency trade-off and serves as an uncertainty proxy.}
We next study how APSS varies with the target risk level. As demonstrated by figure~\ref{fig: rq4}, across datasets and semantic admission thresholds, APSS decreases monotonically as $\alpha$ increases. This reflects the expected coverage--efficiency trade-off: stricter guarantees require larger prediction sets, whereas looser guarantees allow more compact sets. Importantly, this behavior is observed only in the feasible regime, where the calibrated threshold is statistically meaningful.

Beyond efficiency, APSS is also informative about uncertainty. For more ambiguous inputs, stricter semantic criteria, or more open-ended tasks, the framework typically needs to retain more candidates to maintain the same target risk. In this sense, APSS is not merely a secondary metric, but a useful operational signal of model uncertainty under explicit risk control. Larger prediction sets indicate that the model requires more admissible alternatives to preserve the statistical validity of set-valued prediction.

\begin{figure*}[!t]
    \centering
    \includegraphics[width=0.8\linewidth]{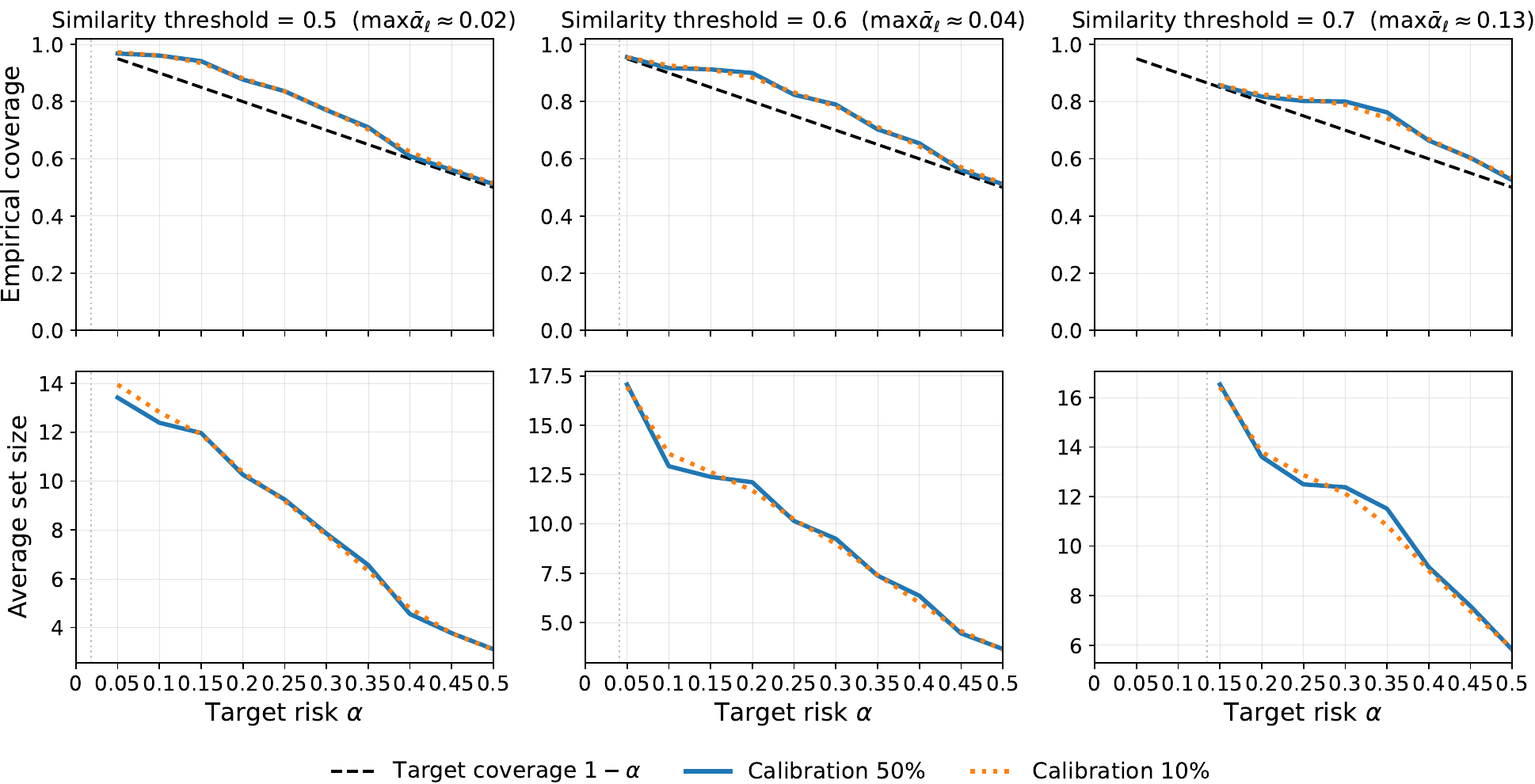}
    \caption{Sensitivity to the split ratio on TriviaQA. We compare the default $50\%$ calibration split with a reduced $10\%$ split. Evaluation is restricted to the common feasible regime determined by the larger MRL. The similar coverage and APSS curves indicate that the proposed framework remains robust and calibration-efficient even with substantially fewer calibration samples.}
    \label{fig: rq7}
\end{figure*}

\paragraph{6) Semantic deduplication improves practical efficiency without undermining utility.}
Although set-valued prediction is beneficial, sampled candidate pools often contain semantically redundant responses that differ only in surface wording. To address this, we perform semantic deduplication using a sentence similarity threshold of $0.9$, merging only highly similar responses within each prediction set. This choice makes the post-processing step conservative: it removes near-duplicates while preserving genuinely distinct admissible answers.

Figure~\ref{fig: rq5} shows that semantic deduplication consistently reduces prediction set size, confirming that a nontrivial fraction of retained candidates are semantically repetitive. This improvement is operationally meaningful rather than merely cosmetic. More compact prediction sets are easier for users or downstream systems to inspect, while preserving the main advantage of set-valued prediction, namely, the ability to provide multiple admissible candidates under a formal guarantee. Therefore, semantic deduplication offers a simple but effective way to improve usability without changing the underlying calibration mechanism.

\paragraph{7) Robustness to specific construction of the reliability score.}
We ablate the components of the reliability score to understand whether the validity of the framework depends heavily on the particular design of $F$. The results in figure~\ref{fig: rq6} show that the empirical coverage curves remain broadly stable across different scoring variants. In other words, once the threshold is calibrated on the held-out calibration set under exchangeability, valid risk control is largely preserved regardless of the exact functional form of the score.

The main difference appears in efficiency rather than validity. Different variants of $F$ produce noticeably different APSS values, indicating different abilities to concentrate admissible responses into compact sets. This suggests that the conformal calibration step is responsible for the statistical guarantee, while the choice of reliability-score components primarily determines how efficiently the admissible answer space is organized. Put differently, $F$ acts as a plug-and-play module for efficiency, not a prerequisite for validity.

\paragraph{8) Robustness under smaller split ratios.}
Finally, we test the sensitivity of the framework to the calibration-test split ratio. As illustrated by figure~\ref{fig: rq7}, even when the available calibration data are substantially reduced to $10\%$, the empirical coverage and APSS curves remain highly similar in the common feasible regime. This indicates that the method is calibration-efficient: it does not rely on an excessively large held-out calibration set to maintain reliable performance.

This robustness is practically important. In real applications, calibration samples are often limited, and methods that require a large held-out set can become difficult to deploy. The results suggest that our framework remains effective even under relatively data-constrained settings, further supporting its practicality for real-world open-ended generation.

\section{Conclusion}
We present a feasibility-aware framework for set-valued prediction in open-ended LLM generation. The key challenge in this setting is that finite sampling may fail to produce any admissible answer, making conventional conformal guarantees inapplicable. To address this, we introduce the minimum risk level to characterize the feasibility boundary induced by finite sampling, and develop a data-driven learn-then-test calibration procedure to construct set-valued predictors with finite-sample marginal coverage guarantees whenever the target risk level is feasible. 
Experiments on six NLG benchmarks with five ``off-the-shelf'' LLMs demonstrate that the proposed framework consistently outperforms MLG-based point prediction, attains valid coverage in the feasible regime, and yields practically useful prediction sets with strong efficiency and robustness. These findings suggest that set-valued prediction is not merely a looser alternative to point prediction, but a more faithful and reliable formulation for open-ended generation. In future work, we plan to explore stronger forms of coverage, more adaptive candidate generation strategies, and broader applications to interactive and agentic LLM systems.

% \appendix
% \section{Appendix}
% ... ...


\bibliographystyle{cas-model2-names}

\bibliography{cas-refs}


% Author biography without author photo.
% Author biography. Author biography. Author biography.
% Author biography. Author biography. Author biography.
% Author biography. Author biography. Author biography.
% Author biography. Author biography. Author biography.
% Author biography. Author biography. Author biography.
% Author biography. Author biography. Author biography.
% Author biography. Author biography. Author biography.
% Author biography. Author biography. Author biography.
% Author biography. Author biography. Author biography.

% \bio{figs/cas-pic1}
% Author biography with author photo.
% Author biography. Author biography. Author biography.
% Author biography. Author biography. Author biography.
% Author biography. Author biography. Author biography.
% Author biography. Author biography. Author biography.
% Author biography. Author biography. Author biography.
% Author biography. Author biography. Author biography.
% Author biography. Author biography. Author biography.
% Author biography. Author biography. Author biography.
% Author biography. Author biography. Author biography.

% \bio{figs/cas-pic1}
% Author biography with author photo.
% Author biography. Author biography. Author biography.
% Author biography. Author biography. Author biography.
% Author biography. Author biography. Author biography.
% Author biography. Author biography. Author biography.

\end{document}